\documentclass[twoside,11pt]{article}

\usepackage{jmlr2e}
\usepackage{amsmath}
\usepackage{amssymb}
\usepackage{bbold}
\usepackage{subcaption}
\usepackage[ruled]{algorithm2e}

\usepackage{tikz}

\jmlrheading{0}{0000}{0-00}{00/00}{00/00}{00000}{Jakob Knollm\"uller and Torsten A. En\ss lin}

\ShortHeadings{Metric Gaussian Variational Inference}{Knollm\"uller and En\ss lin}
\firstpageno{1}

\begin{document}

\title{Metric Gaussian Variational Inference}

\author{\name Jakob  Knollm\"uller \email jakob@mpa-garching.mpg.de
		\AND
		\name Torsten A. En\ss lin \email ensslin@mpa-garching.mpg.de\\
       \addr Max-Planck-Institut f\"ur Astrophysik, Karl-Schwarzschildstr.~1 \\
       85748 Garching, Germany\\
       Ludwig-Maximilians-Universit\"at M\"unchen, Geschwister-Scholl-Platz{\small{}~}1\\
		80539 Munich, Germany}

\editor{ }
\maketitle

\begin{abstract}
	Solving Bayesian inference problems approximately with variational approaches can provide fast and accurate results. Capturing correlation within the approximation requires an explicit parametrization. This intrinsically limits this approach to either moderately dimensional problems, or requiring the strongly simplifying mean-field approach. We propose Metric Gaussian Variational Inference (MGVI) as a method that goes beyond mean-field. Here correlations between all model parameters are taken into account, while still scaling linearly in computational time and memory. With this method we achieve higher accuracy and in many cases a significant speedup compared to traditional methods. MGVI is an iterative method that performs a series of Gaussian approximations to the posterior. We alternate between approximating the  covariance with the inverse Fisher information metric evaluated at an intermediate mean estimate and optimizing the KL-divergence for the given covariance with respect to the mean. This procedure is iterated until the uncertainty estimate is self-consistent with the mean parameter. We achieve linear scaling by avoiding to store the covariance explicitly at any time. Instead we draw samples from the approximating distribution relying on an implicit representation and numerical schemes to approximately solve linear equations. Those samples are used to approximate the KL-divergence and its gradient. The usage of natural gradient descent allows for rapid convergence. Formulating the Bayesian model in standardized coordinates makes MGVI applicable to any inference problem with continuous parameters. We demonstrate the high accuracy of MGVI by comparing it to HMC and its fast convergence relative to other established methods in a number of examples. We investigate real-data applications, as well as synthetic examples of varying size and complexity and up to a million model parameters.
\end{abstract}

\begin{keywords}
Variational Inference, Bayesian Inference, Fisher Information Metric, Gaussian Approximation, Standardization
\end{keywords}
\section{Introduction}
Performing Bayesian inference in large and complex models is challenging. Analytic posteriors are not available for non-conjugate models and only approximate solutions are possible. Depending on the requirements and resources, a large variety of approaches is available. MCMC sampling techniques recover the true posterior exactly in the limit of infinite samples, but are  computationally expensive. An efficient variant is Hamiltonian Monte Carlo (HMC) \citep{HMC}, which explores the posterior distribution following the Hamilton equations. The choice of the parameter coordinate system is also relevant, as it is a way to decouple the different quantities. To increase sampling efficiency, \citet{HMCHierarchy} proposes to choose a standardized coordinate system, in which the deep hierarchical structure of the problem is resolved and flattened down. Here the reparametrization trick \citep{AEVB} is applied to the model parameters directly.

A completely different approach to solve the inference problem is calculating the Maximum Posterior estimate (MAP). To obtain it, one only has to maximize the posterior probability, which is far easier than sampling the entire posterior density. This makes the MAP approach still applicable in extremely high parameter dimensions. The problem with it is that it does not provide any uncertainty quantification on its own. It is also sensitive to any multi-modal feature or degenerate direction in the posterior distribution. This results in over-fitting the data realization or delivering implausible parameter configurations. One way to fix the shortcoming of the missing uncertainty later on is the Laplace approximation (for details see \citet{Bishop}). Here the true posterior is approximated with a Gaussian distribution centered around the MAP estimate. The inverse Hessian of the potential landscape is adapted as covariance estimate. Sometimes also the Fisher information metric is used making it a Fisher-Laplace approximation \citep{kass1995bayes, hartmann2017laplace}. This requires, however, that MAP provides a reasonable result in the first place, which in complex models often is not the case.

It is therefore better to take the uncertainty already into account when approximating the posterior distribution. A way to do this is Variational Inference. For a comprehensive review on this topic see \citet{VariationalReview}. Here a family of parametric probability distributions is selected and the variational parameters are optimized by minimizing the Kullback-Leibler (KL) divergence \citep{KLdivergence}  between the approximate distribution and the true posterior distribution. The KL-divergence measures the average information discrepancy between the two distributions.
For large problems, the mean-field approximation is commonly used, which scales linearly with the problem size \citep{braun2010variational, knowles2011non}. The approximate distribution factorizes over all individual parameters, ignoring any posterior correlation. Often Gaussian distributions are chosen as the parametric family, which provide an uncertainty associated to the mean position, making it Gaussian Variational Inference \citep{GaussianRevisited, lazaro2011variational}. By explicitly parametrizing the covariance, it allows to express correlations between model parameters. Here the problem is the quadratic scaling of the variational parameters with the dimension of the posterior distribution, limiting full-covariance Gaussian variational inference only to moderately sized problems. 

In special cases an exact covariance can be parametrized  in terms of a quantity that only scales linearly with the model parameters \citep{GaussianRevisited}. The associated optimization problem is harder than in the explicit parametrization, but efficient solvers are investigated \citep{khan2013fast}. We want to approach problems with a more general structure, where the linear scaling is not necessarily available.

The choice of the coordinate systems of the parameters also matter when approximating the posterior distribution. Combining the previously mentioned standardization with Gaussian Variational Inference, one obtains Automatic Differentiation Variational Inference (ADVI) \citep{ADVI}. The standardization extends one common variational approach to any posterior over continuous model parameters, making it extremely flexible. For high dimensional posteriors, however, one is again restricted to the mean-field approach. To avoid the heavy computational load associated with a full-covariance approach, Linear Response (LR-)ADVI has been proposed \citep{LR-ADVI} to first perform mean-field ADVI, and then to construct an uncertainty estimate around the obtained mean utilizing the inverse Hessian of the KL-divergence as an uncertainty estimate instead of the obtained mean-field variance. This covariance estimate measures the sensitivity of the approximation with respect to small variations in the variational parameters, containing cross-correlation between all quantities. It follows the logic of the Laplace approximation by first obtaining a comparably inexpensive estimate, and then fixing certain shortcomings later on. Here again one relies on a simpler method to find a good-enough solution. The uncertainty is then not self-consistent with the mean estimate. A problem of this covariance estimate is again the scaling behavior. The sparsity of the matrix depends on the number of global parameters, which are collectively informed by multiple likelihoods. This is a problem for e.g. Gaussian process regression, where one data point informs all latent parameters in the standardized formulation.

Here we want to propose Metric Gaussian Variational Inference (MGVI) to perform approximate Bayesian inference to extremely high-dimensional and complex posterior distributions. Instead of trying to fix the correlations between all parameters in the end, we take them into account during the optimization to obtain self-consistent mean and uncertainty estimates. We make use of standardized model parameters, as they permit a uniform treatment of many problems and thereby effectively widen the applicability of the method. MGVI does not directly optimize the KL-divergence  for a parametric family, instead it performs a number of subsequent Gaussian approximations to the posterior distribution. It iterates between updating the covariance with a term based on the inverse Fisher information metric evaluated at the mean estimate and updating the mean estimate by minimizing the KL-divergence for this given covariance. This procedure is iterated until the mean estimate is consistent with the uncertainty estimate. The covariance estimate is equivalent to the one used for the Fisher-Laplace approximation, as the inverse Hessian of the posterior information is not a valid covariance at every location due to violated positive definiteness. In comparison to the Hessian of the KL-divergence used as covariance estimate in LR-ADVI, our covariance estimate will also be sparse in terms of global parameters, enabling for example large-scale Gaussian process regressions as part of the model. We achieve linear scaling with the posterior dimension by completely avoiding explicitly constructing the covariance at any time. Instead we draw samples from the approximate Gaussian distribution using implicit operators and numerical solutions to large sets of linear equations. All correlations are then stored implicitly within the sample realizations, which are then used to estimate the KL-divergence and its gradient. For minimizing the KL-divergence we rely on efficient Natural Gradient descent \citep{InformationGeometry, NaturalGradientReview}. In order to apply MGVI, a number of conditions have to be fulfilled by the underlying model. First, all parameters have to be continuous, and not discrete. Second, the Fisher information metric of the likelihood requires an accessible eigenbasis, which is e.g. the case for independently sampled data. Third, the true posterior has to be sufficiently Gaussian, and fourth, the standardizing transformation is locally well-approximated by a linear function and higher order terms can be neglected.

In the numerical experiments we apply MGVI to a wide range of different Bayesian inference problems. We validate the method by comparing results to HMC sampling in a synthetic Poisson log-normal Gaussian process regression and a hierarchical logistic regression problem with US presidential election polling data. We demonstrate the scaling of MGVI by approximating a posterior with more than a million parameters in a binary Gaussian process classification problem with simultaneous kernel learning. In this example we also explore the impact of meta-parameter choices for the method. We also apply MGVI to a non-negative matrix factorization problem with a Gamma-Poisson model on the Frey face data set.
Throughout the experiments, MGVI has the highest accuracy in most of the used metrics and is always closest to the HMC estimates. It behaves similarly to full-covariance ADVI, as it captures cross-correlation between all parameters, but is in many cases roughly one order of magnitude faster than even mean-field ADVI, as MGVI relies on natural gradient descent and has only half the number of variational parameters.

\section{Variational Inference}
\subsection{Bayesian Inference}
Bayesian inference in general describes how the knowledge on one quantity of a system affects the knowledge on some other quantity of interest, following Bayes theorem:
\begin{align}
\mathcal{P}(\theta \vert d) = \frac{\mathcal{P}(d\vert \theta) \mathcal{P}(\theta)}{\mathcal{P}(d)} \text{\quad .}
\end{align}
The posterior distribution $\mathcal{P}(\theta\vert d)$ of the unknown quantity $\theta$ given some known data $d$ is equal to the likelihood $\mathcal{P}(d\vert \theta)$ of observing the data given a certain configuration of $\theta$ multiplied by the prior distribution $\mathcal{P}(\theta)$. This whole expression is normalized by the evidence $\mathcal{P}(d)$.

Prior knowledge on the system is encoded in the prior distribution. The likelihood describes how the observed data is related to the parameters of the model. The main difficulty arises in the calculation of the evidence to obtain a properly normalized posterior distribution.

Often this normalization is analytically intractable, especially in non-conjugate models, which are more flexible to encode knowledge on the system. In such cases one has to approximate the true posterior distribution, for example via Maximum Posterior (MAP), variational inference, or MCMC based sampling techniques.

Instead of working with probability distributions, it is equivalent to discuss the problem in terms of information $\mathcal{H}$, defined as the negative logarithm of a probability distribution $\mathcal{P}$, i.e. $\mathcal{H}(\dots) \equiv -\mathrm{ln}\left( \mathcal{P}(\dots)\right)$. Bayes theorem in this perspective reads:

\begin{align}
\mathcal{H}(\theta \vert d) &\equiv - \mathrm{ln}\left(\mathcal{P}(\theta \vert d)\right) \\
 &= \mathcal{H}(d \vert \theta) + \mathcal{H}(\theta) - \mathcal{H}(d) \\
 &\: \widehat{=} \: \mathcal{H}(d \vert \theta) + \mathcal{H}(\theta) \text{\quad .}
\end{align}
In terms of information, the normalization is an additive constant, independent of the quantity of interest. Leaving these terms out is indicated here by the $\widehat{=}$ sign.
\subsection{Kullback-Leibler divergence}
Variational inference allows to approximate posterior distributions to complex problems within reasonable timescales \citep{VariationalReview}. One chooses a parametric family of distributions $\mathcal{Q}_{\eta}(\theta)$ with the variational parameters $\eta$ and minimizes the average information discrepancy between the true posterior and the approximation, measured by the Kullback-Leibler divergence \citep{KLdivergence}, with respect to these parameters. The KL-divergence is defined as:
\begin{align}
\mathcal{D}_\mathrm{KL}(\mathcal{Q}_{\eta}(\theta)\vert \vert \mathcal{P}(\theta\vert d)) &= \int d\theta \: \mathcal{Q}_{\eta}(\theta) \: \mathrm{ln} \: \frac{\mathcal{Q}_{\eta}(\theta)}{ \mathcal{P}(\theta\vert d)} \\
& \:\equiv \langle \mathcal{H}(\theta \vert d) \rangle_{\mathcal{Q}_{\eta}(\theta)} -  \langle \mathcal{H}_{\eta}(\theta) \rangle_{\mathcal{Q}_{\eta}(\theta)} \\
& \: \widehat{=} \: \langle \mathcal{H}(d, \theta) \rangle_{\mathcal{Q}_{\eta}(\theta)} -  \langle \mathcal{H}_{\eta}(\theta ) \rangle_{\mathcal{Q}_{\eta}(\theta)} \text{\quad.} \label{eq:KL_div} 
\end{align}
The first term is the cross-entropy between the distributions and the second is the Shannon-entropy of the approximation, where $\mathcal{H}_{\eta}(\theta)$ is the negative logarithm of the approximating distribution.
Expectation values are expressed by $\langle \dots \rangle_{\mathcal{P}(\dots)}$, noting the respective distribution as index. 
In order to minimize the KL-divergence, the normalization of the posterior is irrelevant, as it does not depend on the variational parameters and can be dropped. The expression in the last line is  equivalent to the negative Evidence Lower Bound (ELBO) \citep{Bishop}.
The parameter solution of minimal KL-divergence provides the variational approximation of the original problem. 

For complex models or approximations we cannot calculate the expectation values analytically, but the KL-divergence can be estimated via samples from the approximation. Together with the reparametrization trick  \citep{AEVB}, the gradients on the variational parameters can be estimated as well. This way we can minimize the KL-divergence in a stochastic optimization procedure even in high dimensions and analytically intractable expectation values. 

When approximating the true posterior with another distribution, certain aspects will be lost. Whether a variational approximation is useful or not depends on the problem-specific requirements and available resources. We want to approach problems with an enormous amount of model parameters and reasonable complexity, in which more accurate methods are unfeasible and variational inference can still provides answers.

\section{Gaussian Variational Inference}
Gaussian Variational Inference \citep{GaussianRevisited} describes variational inference with parametrized Gaussians as the approximating family. The Gaussian distribution exhibits a number of convenient properties, while still providing uncertainty and correlation between parameters.  In this case the approximate distribution is
\begin{align}
\label{eq:GaussianApproximation}
\mathcal{Q}_{\eta}(\theta) &= \mathcal{G}(\theta \vert \bar{\theta}, \Theta) \\
&= \frac{1}{\vert 2 \pi \Theta \vert^{\frac{1}{2}}} e^{-\frac{1}{2}(\theta - \bar{\theta})^\dagger \Theta^{-1} (\theta - \bar{\theta})} \text{\quad ,}
\end{align}
with variational parameters $\eta = (\bar{\theta},\Theta)$ and corresponding KL-divergence
\begin{align}
\label{eq:GaussKL}
\mathcal{D}_\mathrm{KL}\left(\mathcal{G}(\theta \vert \bar{\theta}, \Theta) \vert \vert \mathcal{P}(\theta \vert d)\right)\: \widehat{=}\: \Big\langle\mathcal{H}(d,\theta) \Big\rangle_{\mathcal{G}(\theta \vert \bar{\theta}, \Theta)} - \left\langle\mathcal{H}_{\bar{\theta},\Theta}(\theta) \right\rangle_{\mathcal{G}(\theta \vert \bar{\theta}, \Theta)} \text{\quad .}
\end{align}
In order to perform the variational inference of the parameters, the expression above is minimized with respect to the variational mean $\bar{\theta}$ and covariance $\Theta$ parameters. 
The second term in this equation is the Shannon entropy of the approximate Gaussian with the analytic form
\begin{align}
\label{eq:Shannon}
 \left\langle\mathcal{H}_{\bar{\theta},\Theta}(\theta) \right\rangle_{\mathcal{G}(\theta \vert \bar{\theta}, \Theta)} \widehat{=} \frac{1}{2}\mathrm{ln} \left\vert 2 \pi e \Theta \right\vert \text{\quad .}
\end{align}
Here $\vert \dots \vert$ expresses a determinant and $e$ is Eulers' number. Note that this expression is independent of the variational mean parameter $\bar{\theta}$.
To efficiently optimize the KL-divergence we require gradient information with respect to the variational parameters. Derivatives with respect to the mean and covariance are simply the expected gradient and curvature over the Gaussian distribution, respectively \citep{GaussianRevisited}.
\begin{align}
\label{eq:KLGradient}
\frac{\partial}{\partial \bar{\theta}}\mathcal{D}_\mathrm{KL}&= \left \langle \frac{\partial}{\partial \theta} \mathcal{H}(d,\theta) \right\rangle_{\mathcal{G}(\theta \vert \bar{\theta}, \Theta)} \text{\quad \quad \quad \quad, and} \\
\label{eq:KLGradientCurvature}
\frac{\partial}{\partial \Theta}\mathcal{D}_\mathrm{KL} &= \frac{1}{2} \left\langle\frac{\partial^2 }{\partial \theta\partial \theta^\dagger}  \mathcal{H}(d,\theta)\right\rangle_{\mathcal{G}(\theta \vert \bar{\theta}, \Theta)} - \frac{1}{2}\Theta^{-1} \text{\quad .}
\end{align}
For the mean parameter only the cross-entropy term is relevant and if we were to optimize only with respect to this parameter, we avoid the necessity of calculating determinants of possibly large matrices. Setting the derivative with respect to the covariance to zero, we obtain the following implicit relation:
 \begin{align}
 \Theta^{-1} =&  \left\langle\frac{\partial^2 }{\partial \theta\partial \theta^\dagger}  \mathcal{H}(d,\theta)\right\rangle_{\mathcal{G}(\theta \vert \bar{\theta}, \Theta)} \\ 
 =& \left\langle\frac{\partial  \mathcal{H}(d,\theta)}{\partial \theta} \frac{\partial \mathcal{H}(d,\theta) }{\partial \theta^\dagger}  \right\rangle_{\mathcal{G}(\theta \vert \bar{\theta}, \Theta)} - \left\langle \frac{1}{\mathcal{P}(d,\theta)} \frac{\partial^2 \mathcal{P}(d,\theta) }{\partial \theta\partial \theta^\dagger}\right\rangle_{\mathcal{G}(\theta \vert \bar{\theta}, \Theta)} \text{\quad .}
 \label{eq:actual_covariance}
 \end{align}
This relation serves as starting point for Metric Gaussian Variational Inference. We will set up an iterative fixed-point scheme where we start with some initial mean value $\bar{\theta}$, and adapt an implicit solution for the covariance, similarly to the expression above. For this Gaussian distribution we can then optimize the KL-divergence only with respect to the mean parameter, keeping the covariance fixed. Once it is optimized, we update the covariance to the implicit solution for the new mean parameter. This procedure is then iterated until convergence.
Unfortunately the right side of the above equation is not necessarily compatible with a covariance, as in general it is not strictly positive definite. The first term, containing the outer product of first derivatives certainly is. Problematic is the second term, which involves second derivatives of the probability distribution. It might contain negative eigenvalues, harming the overall positive definiteness of the covariance of the Gaussian in this approximation. For this reasons we cannot use this expression. It is also a dense matrix for global parameters, which are collectively informed by common likelihoods. We will instead use a similar expression as covariance based on the inverse Fisher information metric as approximation, which overcomes these limitations. 

Often the covariance is parametrized explicitly in terms of another matrix $A$ via $\Theta = A A^\dagger$ to ensure positive definiteness. 
The problem with an explicit parametrization of the variational covariance is the quadratic scaling in the model parameters. It allows only for moderately sized problems. To overcome this limitation, usually a diagonal covariance is assumed, which is a mean-field approach. A diagonal covariance approximation cannot capture correlations between posterior parameters, severely limiting the expressiveness of the result.

We cannot calculate the KL-divergence for arbitrary problems analytically, but it is always possible to approximate the expectation value through sample averages. Therefore, we optimize a stochastic estimate of the KL-divergence with the corresponding stochastic gradient.
\begin{align}
 \langle  \mathcal{H}(d,\theta) \rangle_{\mathcal{G}(\theta \vert \bar{\theta}, \Theta)} &\approx \frac{1}{N}\sum_{i=1}^N \mathcal{H}(d,\theta_*^i) =  \frac{1}{N}\sum_{i=1}^N \mathcal{H}(d,\bar{\theta} +\Delta\theta_*^i) \\
 \theta_*^i &\sim \mathcal{G}(\theta \vert \bar{\theta}, \Theta)) \text{\quad or \quad} \Delta\theta_*^i\sim \mathcal{G}(\theta \vert 0, \Theta) \text{\quad .}
\end{align}
We indicate sample realizations with the lower $*$-index, and note $\Delta$ for zero-centered Gaussian samples.
Splitting the sample in a mean contribution and Gaussian residual $\theta_*^i = \bar{\theta} + \Delta \theta_*^i$ allows us to adapt the samples to an updated mean, which is the reparametrization trick in its simplest form \citep{AEVB}. In the end we will be following an implicit optimization scheme, as briefly discussed above. For this it is therefore sufficient to obtain residual samples $\Delta\theta_*^i$ to learn only the mean $\bar{\theta}$ of the approximate Gaussian for a given covariance.
\section{Standardization}
Deep hierarchical Bayesian models are used to describe sophisticated models and complex dependencies and they strongly vary throughout different applications. To remove large parts of the problem-specific complexity from the variational inference, we prefer to work in standardized parameter coordinates, following Automatic Differentiation Variational Inference (ADVI) \citep{ADVI}. In hierarchical models, certain parameters might be restricted to only a certain parameter range. Performing the variational approximation with a Gaussian in these original coordinates might not be possible due to the infinite support of the Gaussian distribution. In the standard coordinates all parameters follow a priori a standard Gaussian distribution, removing this complication. This transformation opens the door to apply the here proposed algorithm to any problem with continuous parameters. It might not be necessary to standardize problems with infinite support on all parameters, and there the method should also work in the original coordinates. We do not want to treat this special case separately and choose the more unified standard parametrization. In the hierarchical formulation the interdependence between the different quantities might be strong, resulting in a numerically stiff problem. The hierarchical structure is resolved by applying the  reparametrization trick  \citep{AEVB} to the model parameters, leading to a flat model. In the context of HMC sampling, these standard coordinates are also used to explore the posterior more efficiently \citep{HMCHierarchy}. These  numerical and conceptual advantages also apply to variational inference, especially if the true distribution is well approximated with a Gaussian \citep{WhitePriors}.

Conceptually one takes a likelihood $\mathcal{P}(d\vert \theta) $ together with a hierarchical prior $\mathcal{P}(\theta)=\mathcal{P}(\theta_1\vert \theta_2 \dots \theta_N) \dots \mathcal{P}(\theta_{N-1}\vert \theta_N) \mathcal{P}(\theta_N)$ and performs coordinate transformation to uniform parameters using the multivariate distributional transform $\mathcal{F}^{-1}_{\mathcal{P(\theta)}}( \dots)$ \citep{DistributionalTransform}. This uses the inverse conditional cumulative density functions, following the logic of inverse transform sampling \citep{InverseTransform}. 
\begin{align}
u &\sim \mathcal{U}(u)\\
\theta & = \mathcal{F}_{\mathcal{P}(\theta)}^{-1}(u)\\
\Rightarrow \theta &\sim \mathcal{P}(\theta)\text{\quad .}
\end{align}
 We draw samples from the prior distribution by drawing samples $u$ from the uniform distribution $\mathcal{U}(u)$, and processing them through $\mathcal{F}^{-1}_{\mathcal{P(\theta)}}( \dots)$. The sample $u$ has finite support on the unit interval and performing a Gaussian approximation in these coordinates is not sensible. A second transformation to standard Gaussian coordinates enables this. The transformation is given by the cumulative density function of the Gaussian $\mathcal{F}_{\mathcal{G}(\xi \vert 0, \mathbb{1})}$.
\begin{align}
\xi &\sim \mathcal{G}(\xi \vert 0, \mathbb{1})\\
u & =\mathcal{F}_{\mathcal{G}(\xi \vert 0, \mathbb{1})}(\xi)\\
\Rightarrow u &\sim \mathcal{U}(u)\text{\quad .}
\end{align}
The resulting $\xi$ parameters are a priori independent and the entire complexity is encoded in the composition of the two transformations  $\theta = \mathcal{F}^{-1}_{\mathcal{P(\theta)}}\circ \mathcal{F}_{\mathcal{G}(\xi ,\mathbb{1})}(\xi) \equiv f(\xi)$.
The  probability distribution and its information in these coordinates are
\begin{align}
\mathcal{P}(d,\xi) &= \mathcal{P}\left(d\vert f(\xi)\right) \mathcal{G}(\xi \vert 0,\mathbb{1}) \\
\mathcal{H}(d,\xi) \:&\widehat{=} \:\mathcal{H}\left(d\vert f(\xi)\right) + \frac{1}{2} \xi^\dagger \mathbb{1} \xi \text{\quad .}
\label{eq:StandardHamiltonian}
\end{align}
For the rest of the paper we will indicate standardized parameters with $\xi$, whereas general parameters are $\theta$. The Gaussian approximation in standard coordinates is denoted as $\mathcal{G}(\xi\vert\bar{\xi},\Xi)$. This standardization allows us to obtain an uncertainty estimate of a certain structure, which enables us to draw samples from the approximate distribution.
\subsection{Gaussian Variational Inference in standard coordinates}
It is often stated that in the case of Gaussian prior distributions for all $N$ parameters, Gaussian variational inference only requires $N + M$ variational parameters to express the full mean and covariance \citep{GaussianRevisited}, with $M$ being the number of independent likelihood contributions. With standardization we can express any continuous probability distribution in terms of a standard Gaussian prior and a corresponding transformation. We want to emphasize that this statement does not hold for arbitrary transformations. To be precise, it only holds for a linear mixture of latent variables, followed by a point-wise non-linear function. Consider $M$ independent likelihoods with data $d_i$, parameters $\theta_i$ and their relation to the latent Gaussian parameters $\theta_i = f_i(\xi)$. According to Eq.~\ref{eq:actual_covariance}, the covariance must satisfy the following relation:

\begin{align}
\Xi^{-1} =&  \mathbb{1} + \left \langle \sum_{i=1}^M \frac{\partial \mathcal{H}(d_i\vert \theta_i)}{\partial\xi} \frac{\partial \mathcal{H}(d_i\vert \theta_i)}{\partial \xi^\dagger}\right \rangle _{\mathcal{G}(\xi \vert \bar{\xi}, \Theta)}- \left \langle  \sum_{i=1}^M \frac{1}{\mathcal{P}(d_i\vert \theta_i)} \frac{\partial^2 \mathcal{H}(d_i \vert \theta_i)}{\partial \xi \partial \xi^\dagger}\right \rangle_{\mathcal{G}(\xi \vert \bar{\xi}, \Theta)} \\
=& \mathbb{1} +  \left \langle \sum_{i=1}^M \frac{\partial f_i(\xi)}{\partial \xi}\frac{\partial \mathcal{H}(d_i\vert \theta_i)}{\partial\theta_i} \frac{\partial \mathcal{H}(d_i\vert \theta_i)}{\partial \theta_i^\dagger}\frac{\partial f_i(\xi)^\dagger}{\partial \xi^\dagger}\right \rangle_{\mathcal{G}(\xi \vert \bar{\xi}, \Theta)} \nonumber \\
 &-  \left \langle  \sum_{i=1}^M \frac{1}{\mathcal{P}(d_i\vert \theta_i)} \frac{\partial \mathcal{H}(d_i \vert \theta_i)}{\partial \theta_i}\frac{\partial^2 f_i(\xi)}{\partial \xi \partial \xi^\dagger}\right \rangle_{\mathcal{G}(\xi \vert \bar{\xi}, \Theta)} \nonumber \\
 &-  \left \langle  \sum_{i=1}^M \frac{1}{\mathcal{P}(d_i\vert \theta_i)}\frac{\partial f_i(\xi)}{\partial \xi } \frac{\partial^2 \mathcal{H}(d_i \vert \theta_i)}{\partial \theta_i \partial \theta_i^\dagger }\frac{\partial f_i(\xi)}{\partial \xi^\dagger }\right \rangle_{\mathcal{G}(\xi \vert \bar{\xi}, \Theta)}\text{\quad .}
 \label{eq:GVA_covariance}
\end{align}
It is proposed to parametrize this covariance in the following form:
\begin{align}
\Xi^{-1} = \mathbb{1} + R^\dagger \Lambda R\text{\quad .}
\label{eq:exact_covariance_para}
\end{align}
with $\Lambda$ being a diagonal matrix of dimension $M$, containing the variational parameters for the covariance. This, however, is only be exact if the standardization has the following form:
\begin{align}
\theta = f(\xi) =g(R \xi)\text{\quad .}
\end{align}
Here $R$ is an arbitrary, matrix and $g$ an arbitrary, point-wise, non-linear function. The first and second derivatives of this function with respect to the parameters reads:
\begin{align}
 \frac{\partial f}{\partial \xi} &=\frac{\partial g(R\xi)}{\partial \xi} = g'(R \xi) R \\
 \frac{\partial^2 f}{\partial\xi \partial(\xi)^\dagger}&=\frac{\partial^2 g(R\xi)}{\partial \xi\partial\xi^\dagger} = R^\dagger g''(R \xi) R \text{\quad .}
 \end{align}
 The parameter-dependent parts $g'(R\xi)$  and $g''(R\xi)$ are diagonal matrices of dimension $M$, and the matrix $R$ maps from the $N$-dimensional parameter space to the $M$-dimensional space. 

We insert these derivatives into the expectation values in Eq.~\ref{eq:GVA_covariance} and pull out the linear $R$ terms out of the integrals, resulting in an expression of the form
\begin{align}
\Xi^{-1} = \mathbb{1} + R^\dagger \left \langle X_1(\xi) - X_2(\xi)- X_2(\xi)\right \rangle_{\mathcal{G}(\xi \vert \bar{\xi}, \Xi)} R\text{\quad .}
\end{align}
Such a term can be exactly approximated by a parametrization of the form  Eq.~\ref{eq:exact_covariance_para}, as $X_1$, $X_2$ and $X_3$ are diagonal matrices depending on the parameter. 

For more general standardization functions $f(\xi)$, containing a number of consecutive linear and and point-wise non-linear transformations, this is not possible, as only the outermost matrix can be pulled out of the expectation value.
 So in the general, non-linear case, the number of required variational parameters to express the covariance fully does scale quadratically with the number of model parameters.
\section{Approximating the covariance}

We want to explore the properties of extremely high dimensional posterior distributions through an efficient approximation. The associated volume in such high dimensional spaces is enormous and in it the posterior might exhibit a rich structure. Capturing the posterior structure within the approximation requires a global perspective on it, involving large numbers of parameters to be learned. Already capturing correlations between all model parameters explicitly requires a memory that scales quadratically with the posterior dimension. 

In order to avoid such unfavorable scaling, we have to explore the posterior only from a more local perspective, where we only rely on quantities scaling linearly with dimensions. One example for such an approach is the MAP approach. It, however, is susceptible to implausible results, and to getting stuck in local minima and at improbable parameter configurations of elongated valleys along degenerate directions in complex models. The reason for this is that MAP can be regarded as an approximation to the posterior with a delta distribution, which is highly sensitive to local structures in the information landscape.

To avoid this, we have to account for uncertainty all along the way. We want to do this by using a Gaussian distribution to approximate the posterior, which, in addition to a location, also has a scale. This scale is extremely helpful in maneuvering through the landscape outlined by the posterior, as the Gaussian simply cannot fit into all the small local features and degenerate directions a delta distribution is sensitive to. Only structures of the posterior comparable to its own size or larger couple to the Gaussian.

For this, we have to extract an estimate of the posterior uncertainty from a local perspective. The first thing that comes to mind is the Laplace approximation, which uses the inverse Hessian at the location of the MAP solution as a covariance. It explores locally the curvature of the negative log-posterior and associates strongly curved directions with low uncertainty and vice versa. This Laplace approximation is widely used to extract uncertainties from point estimates, but it fundamentally requires the MAP approach to provide reasonable results in the first place. 

For our purpose, we cannot use the inverse Hessian as it is not necessarily a valid covariance outside a mode. A covariance matrix exhibits strictly positive eigenvalues, but the Hessian measures curvature, which has vanishing or negative eigenvalues in plateaus and concave directions, respectively, which both are often encountered in high dimensional and complex models. This is the same reason we cannot use the expression given in Eq.~\ref{eq:actual_covariance}, the implicit solution to the covariance in Gaussian variational inference. Here one could drop the problematic term, which for approximately Gaussian posteriors will be small anyway and use
\begin{align}
\Theta^{-1} \approx \left\langle\frac{\partial  \mathcal{H}(d,\theta)}{\partial \theta} \frac{\partial \mathcal{H}(d,\theta) }{\partial \theta^\dagger}  \right\rangle_{\mathcal{G}(\theta \vert \bar{\theta}, \Theta)} \text{\quad .}
\end{align}
This is precisely the term used in LR-ADVI \citep{LR-ADVI} to approximate the covariance around the mean-field ADVI  mean estimate. 
 It is certainly positive definite and somewhat close to the true covariance, but we cannot efficiently represent it in high dimensions without severe limitations on the used models. It is a dense matrix for global parameters, which are collectively informed by the same data points. One example where the overall problem does not factorize into independent sub-problems is Gaussian process regression in the standardized coordinates.  We have therefore no access to its eigenbasis without storing and decomposing the dense (sub-)matrices explicitly, something we cannot afford in large problems. We need access to the eigenbasis to generate samples from the approximation, used for estimating the KL-divergence and its gradient. This term itself is an Gaussian expectation value and can be approximated via samples. However, such a sub-sampled matrix is only invertible if the samples constitute a full basis, requiring at least as many samples as parameter dimensions. This, again, is equivalent to storing an entire matrix directly, and therefore not practical. A covariance of this form, however, will serve as the inspiration for the approximation we will be using.

\subsection{Fisher information metric as covariance}
To approach truly large inference problems we require three fundamental properties from the covariance approximation. First, it has to be strictly positive definite, a defining feature of any covariance. Second, it has to resemble the true covariance as closely as possible, at least in limiting cases. Third, the structure of the approximation allows to draw samples from the approximate Gaussian, without the necessity of ever constructing the explicit covariance. All these properties are fulfilled by the covariance proposed in this section based on the inverse Fisher information metric $I^{-1}$. 
Inside the mode it is considered to be inferior to the Laplace approximation \citep{kass1995bayes}, but it is a valid covariance outside. Nevertheless, sometimes it is used to describe the uncertainty around the MAP location \citep{hartmann2017laplace}. It measures the sensitivity of the posterior with respect to small parameter variations and it consists out of two parts $I = I_d + I_\theta$. First, the Fisher information metric of the likelihood:

 \begin{align}
 I_d(\theta) &\equiv \left\langle \frac{\partial\mathcal{H}(d \vert \theta)}{\partial\theta} \frac{\partial\mathcal{H}(d \vert \theta)}{\partial \theta^{\dagger}} \right\rangle_{\mathcal{P}(d \vert \theta)} \text{\quad .}
 \end{align} 
 In a frequentist setting, the inverse of this metric gives the  Cram{\'e}r-Rao bound \citep{cramer,rao}, a lower bound to the uncertainty of an estimator $\widehat{\theta}$:
 \begin{align}
I_d(\theta)^{-1} \leq \left\langle\left( \theta-\widehat{\theta}\right)\left(\theta-\widehat{\theta}\right)^\dagger\right\rangle_{\mathcal{P}(d\vert\theta)}\text{\quad .}
 \end{align}
 The $\leq$ indicates that the right minus the left side of the equation exhibits a positive semi-definite matrix.
 
 The second part is the information metric of the prior distribution, given by:
 
 \begin{align}
 I_\theta &= \left\langle\frac{\partial \mathcal{H}(\theta)}{\partial \theta}\frac{\partial \mathcal{H}(\theta)}{\partial \theta^\dagger}\right\rangle_{\mathcal{P}(\theta)} \text{\quad .}
 \end{align}
 This quantity is a lower bound to the prior variance of an estimator (see  \citet{schutzenberger1957generalization} for vanishing likelihood), i.e.: 
  \begin{align}
  I_\theta^{-1} \leq \left\langle\left( \theta-\widehat{\theta}\right)\left(\theta-\widehat{\theta}\right)^\dagger\right\rangle_{\mathcal{P}(\theta)} \text{\quad .}
 \end{align}
 We now have two bounds on the variance of the estimator, originating from information provided by prior and likelihood. To get to the posterior, we have to add up those two information sources. In the spirit of Gaussian error propagation, we constrain the posterior uncertainty by adding up the corresponding Fisher metrics. So the posterior covariance, compared to the prior one, will be at least reduced by the inverse Fisher metric of the likelihood. We cannot evaluate the resulting term at the location of the ground truth $\theta$, as it is not available. We instead assume the estimator $\widehat{\theta}$ to be sufficiently close to provide a good approximation, which assumes sufficient local Gaussianity in the posterior, an assumption we will rely on later anyway. In this case, the inverse sum of the two metrics, evaluated at the estimator, should tend to be a lower bound to the true posterior variance. This is not a precise inequality and how it behaves in certain conditions will have to be explored in the future or case by case. We expect it to hold for not too extreme models, and otherwise at least to be sufficiently close. Thus, we state
 
 \begin{align}
 I(\widehat{\theta}) &\equiv I_d(\widehat{\theta}) +I_\theta \text{\qquad  and} \\
 I(\widehat{\theta})^{-1}&\lessapprox \left\langle\left( \theta-\widehat{\theta}\right)\left(\theta-\widehat{\theta}\right)^\dagger\right\rangle_{\mathcal{P}(\theta \vert d)}  \text{.}
 \end{align}
 By construction, the inverse Fisher information metric has only positive eigenvalues and we do not necessarily have to be in a mode, making it valid to use as covariance at every location, compared to the inverse Hessian with its potentially negative eigenvalues. 
 
From now on, we identify the estimator with the estimate $\widehat{\theta}$, and interpret the variance of an estimator as uncertainty around the estimate. The inverse Fisher metric is then a lower bound to this uncertainty. Those two quantities constitute a Gaussian distribution $\mathcal{G}(\theta  \vert \bar{\theta}, I(\widehat{\theta})^{-1})$ with mean $\bar{\theta} \leftarrow \widehat{\theta}$. 

Here it is important to distinguish between the estimate $\widehat{\theta}$ and the mean of the Gaussian $\bar{\theta}$, which only initially coincide. In an iterative scheme we will use the location of the estimate $\widehat{\theta}$ to estimate the local uncertainty. While keeping this quantity fixed, we optimize for the mean parameter $\bar{\theta}$ via variational inference, such that the resulting Gaussian better matches the true posterior distribution. At this location we update the parameter estimate $\widehat{\theta} \leftarrow \bar{\theta}$. This way, we resolve the explicit dependence of the uncertainty estimate on the mean parameter and alleviate the necessity of calculating the Shannon entropy terms in the KL-divergence, containing the determinant of the possibly large covariance. 

 The surrounding landscape of this new estimate will have changed, compared to the previous location, and so will the inverse of the local metric. We set this as new covariance and repeat the procedure. Once the location and uncertainty are self-consistent with the posterior, we have converged to our final approximation. Instead of minimizing the KL-divergence within the family of a parametric distribution, we iteratively solve the locally Gaussian approximation problem, to narrow in towards the posterior mode. This bares a similarity to second order optimization, where always the locally quadratic problem is solved to iteratively find optima.

\subsection{Standardized Metric}

The information metric as an abstract mathematical object is invariant under coordinate transformation. In the previously discussed standard coordinates, the metric has an especially simple structure. A priori we deal with independent, standard Gaussian parameters $\xi \sim \mathcal{G}(\xi\vert 0, \mathbb{1})$, without any hierarchical structure. Here the prior information metric is simply the covariance of the Gaussian, the identity operator:
\begin{align}
I_{\xi} = \mathbb{1}\text{\quad .}
\end{align}
The standard parameters are related to the original parametrization of the system via the possibly complex nonlinear transformation $\theta = f(\xi)$. 
The likelihood metric in the standard  coordinates therefore is simply the push-forward from the likelihood metric in the original parametrization. 
\begin{align}
I_{d}(\xi) = &
\left\langle \frac{\partial \mathcal{H}(d \vert \xi)}{\partial\xi }  \frac{\partial \mathcal{H}(d  \vert \xi)}{\partial\xi ^\dagger}\right\rangle_{\mathcal{P}(d\vert \xi)} \\=& \:\left( \frac{\partial f(\xi)}{\partial \xi} \right)^\dagger\left\langle \frac{\partial \mathcal{H}(d \vert \theta)}{\partial\theta }  \frac{\partial \mathcal{H}(d \vert \theta)}{\partial\theta^\dagger}\right\rangle_{\mathcal{P}(d\vert \theta)}\frac{\partial f(\xi)}{\partial \xi} \\ 
=& \: J(\xi)^\dagger I_d(f(\xi)) J(\xi)\text{\quad .}
\label{eq:LLmetric}
\end{align}
Here $ J(\xi) = \frac{\partial f(\xi)}{\partial \xi}$ is the Jacobian of the transformation with respect to the new coordinates.

The  Cram{\'e}r-Rao bound in the standardized coordinates acquires additional curvature terms $X$ \citep{barrau2013note}:
 \begin{align}
\left\langle\left( \xi-\widehat{\xi}\right)\left(\xi-\widehat{\xi}\right)^\dagger\right\rangle_{\mathcal{P}(d\vert\xi)} \geq I_{d}(\xi)^{-1} + X\text{\quad .}
\end{align}
We will neglect those additional $X$ terms, restricting ourselves to only parameter models with a sufficiently linear standardization transformation, at least locally. Therefore, we do not expect the method to perform well in the case of models with extreme $X$ terms, which should be hard in general. Extensions of MGVI that treat this term better are left for future research.

For the uncertainty approximation we evaluate this expression at the current parameter estimate. The overall metric in standardized coordinates will therefore always have the following structure:

\begin{align}
\label{eq:UsedMetric}
I(\widehat{\xi}) \: &\equiv I_d(\widehat{\xi}) + I_\xi \\
 &= J(\widehat{\xi})^\dagger I_d(f(\widehat{\xi})) J(\widehat{\xi}) + \mathbb{1}  \text{\quad .}\\
\end{align}
It only consists of three parts. First, the prior metric, which is the identity operator in the space of standard parameters. Second, the Fisher information metric of the likelihood, which is available for a large number of commonly used likelihoods. And third, the Jacobian of the standardization transformation. This transformation has to be implemented anyway, as it is equivalent to the model implementation.
Its Jacobian can then be obtained by auto-differentiation, or consistently applying the chain rule. As long as the likelihood metric in the original coordinates does not require it, none of these quantities have to be stored in the form of a dense matrix. For the common case of independent data points, the likelihood factorizes and thus allowing for an implicit metric. We will elaborate on the concept of implicit operators in the dedicated Section \ref{sec:implicit}. Nevertheless, using the inverse of the metric as approximate covariance  is a non-diagonal approximation that captures correlations between all involved parameters.

\subsection{Validity of the covariance approximation}
The validity of the covariance approximation will depend on the properties of the system at hand. Here we will discuss three limiting cases in which the inverse Fisher metric is an accurate representation of the true uncertainty. 
The first scenario is asymptotic normality of the posterior distribution under the Bayesian Central Limit Theorem \citep{BCLT}. For a large amount of independently drawn data, the prior information will become irrelevant, according to the  Bernstein-von Mises Theorem \citep{van2000asymptotic}. The posterior will approach the Gaussian distribution:
\begin{align}
\mathcal{P}(\xi \vert d) \approx \mathcal{G}\left(\xi\vert \widehat{\xi}, I_d(\widehat{\xi})^{-1}\right)\text{\quad .}
\end{align}
Here $\widehat{\xi}$ is a Bayesian estimator of $\xi$. The resulting covariance is equivalent to the term given in Eq.~\ref{eq:LLmetric}. Our covariance approximation contains additional to this term also the prior metric. In this highly informed setup, the likelihood will be by far the most dominant term, and the additive $\mathbb{1}$  prior metric becomes irrelevant, obeying the Bernstein-von Mises Theorem. So, in the Bayesian central limit, our approximate covariance coincides with the true posterior uncertainty. In this scenario, however, a MAP estimate, which essentially is the Maximum Likelihood estimate, will also provide reasonable results. Nevertheless, this behavior in the regime of large amounts of data is reassuring. 

The opposite case is a vanishing likelihood. If data is scarce, we do not gain much information compared to the prior distribution.  In truly large inference problems we easily encounter situations where we have to constrain millions of parameters with only thousands of data points. If we want to approach such problems, it is vital to be accurate in this limit, and the key to this is the standardized parametrization. In the trivial case of no likelihood at all, the posterior is equivalent to the standardized Gaussian prior and the likelihood contribution to the metric vanishes. Here our approximate covariance will again be equivalent to the true uncertainty. 

The inverse Fisher information metric is therefore a good approximation for large, as well as small amounts of data. The remaining question is, how well our approximation interpolates between those two limiting scenarios.
So, the prior uncertainty is exact and we combine it with a lower bound of the uncertainty originating from the likelihood by adding the Fisher information metrics.
In the limiting case of a Gaussian likelihood and linear standardization we actually obtain the true posterior covariance and our approximation will be also exact. In general cases the fidelity of the uncertainty estimate will depend on how well the inverse likelihood metric describes the uncertainty originating from the data. This is essentially a statement on how tight the Cram{\'e}r-Rao bound is to the true uncertainty. In the worst case scenario, the inverse likelihood metric vanishes, and our approximation approaches a delta distribution, ultimately resulting in a MAP estimate. If it does not vanish, our approximation will be a better representation of the posterior.

The Cram{\'e}r-Rao bound can be attained if, and only if, the likelihood is a member of the exponential family \citep{wijsman1973attainment}, which includes a large number of commonly used likelihoods. In such cases, we expect the inverse Fisher metric to well represent the uncertainty and our approximation to be valid. Also in cases where the likelihood is close to a member of the exponential family, the covariance should be reasonable. 

In the context of high-dimensional and complex problems, several of these scenarios might be realized within the same model. Certain data might constrain a number of parameters extremely well, whereas other are only weakly informed. The former parameters might be in the regime of the central limit, the later could still be prior-dominated, and others will fall in between. As long as the model is not too extreme, our proposed covariance can capture the uncertainty and correlations in all these regimes simultaneously.

\section{Implicit Operators}
\label{sec:implicit}
The information metric as a matrix has a dimension of the number of parameters squared. Storing it explicitly on a computer is already unfeasible for relatively small problems. In imaging for example, millions of pixel parameters are not uncommon and we will demonstrate MGVI for such an example at the end. The metric is built out of a collection of linear transformations, projections, and diagonal operators that all can be expressed efficiently by sparse matrices represented by computer routines. The metric itself is therefore expressible as an implicit operator, described by the composition of these simple operators. By construction, the metric is linear and positive definite, and therefore invertible. The inverse of the metric correlates all parameters with each other, usually resulting in a dense matrix expression, which will serve as approximate posterior covariance. This object is of interest during the inference, as well as for posterior analysis. As mentioned before, we cannot afford to store the posterior covariance at any moment explicitly. We have to extract all relevant information on correlations from the implicit metric only. This requires to apply the metric, as well as its inverse to vectors.

The implicit representation allows us to apply the metric $I=\Theta^{-1}$ to some vector $x$ efficiently. 
\begin{align}
\label{eq:QuadraticProblem}
b &= \Theta^{-1} x \text{\quad .}
\end{align}
More problematic is the application of the covariance $\Theta$, the inverse metric, to some vector $b$.
\begin{align}
\label{eq:SolvedProblem}
x &= \Theta b \text{\quad .}
\end{align}
This matrix inversion can be done by solving Eq.~\ref{eq:QuadraticProblem} numerically for $x$, equivalent to solving a set of linear equations. The metric is certainly positive definite and hermitian, allowing the use of the Conjugate Gradient algorithm \citep{AgonizingPain} for this inversion. This algorithm makes extensive use of the positive definiteness of the problem, leading to rapid convergence, compared to more general solvers. The resulting vector $x$ then approximately satisfies Eq.~\ref{eq:SolvedProblem}. 
\subsection{Drawing samples from the approximation}
\label{sec:sampling}
The numerical inversion is the key to drawing samples from a Gaussian distribution with only an accessible precision matrix. We need those samples to estimate the KL-divergence and in the end they can be used to propagate uncertainty to any quantity of interest, based on the posterior. Those samples can be drawn by following the scheme outlined in \citet{papandreou2010gaussian}, as our approximate covariance conveniently follows the structure of a conditional Gaussian distribution. This procedure scales linearly in time and memory with the dimensionality of the posterior. The main idea is to draw a sample from a Gaussian distribution with the inverse covariance, and then obtain a sample from the actual Gaussian by applying the covariance via numerical inversion.

 In general, we can draw samples from a zero-centered Gaussian by drawing independent, white noise $\eta_*$ in the eigenbasis of the covariance $\Theta = Q \Lambda Q^\dagger$ with eigenvectors $Q$ and eigenvalues on the diagonal of $\Lambda$, weighting it with the square-root of the eigenvalues, and transforming it into the original space:
 \begin{align}
 \Delta\theta_* &= Q \sqrt{\Lambda} \eta_* \text{\quad , therefore}\\
 \Delta\theta_* &\sim \mathcal{G}(\theta \vert 0, \Theta)\text{\quad .}
 \end{align}
 Unfortunately,  we do not have direct access to $\Theta$, as it is only implicitly given trough its inverse $\Theta^{-1}$, but we can draw samples according to $\Delta\phi_* \sim \mathcal{G}(\phi \vert 0, \Theta^{-1})$, via $\Delta\phi_* = Q \sqrt{\Lambda^{-1}} \eta$. Numerically, we can approximately apply $\Theta$ to this sample from the Gaussian with inverse covariance, which yields
\begin{align}
\label{eq:inverse_samples}
 \Delta\theta_* &\equiv \Theta \Delta\phi_*  \\
 &= Q \Lambda Q^\dagger Q \sqrt{\Lambda^{-1} } \eta_* \\
 &= Q \sqrt{\Lambda} \eta_* \text{\quad .}
\end{align}
Therefore, $\Delta\theta_*$ is then a sample from $\mathcal{G}(\theta \vert 0, \Theta)$. Note that $Q$ is unitary, therefore its adjoint is the inverse, $Q^\dagger Q = \mathbb{1}$. A set of  such samples $\{\Delta\theta_*\}_N$ serves now as a representation of the intractable, dense covariance. 
\begin{align}
\Theta = \langle \theta \theta^\dagger \rangle_{\mathcal{G}(\theta\vert 0,\Theta)} \approx  \frac{1}{N}\sum_{i=1}^N \Delta\theta_*^i \Delta\theta_*^{i\dagger}\text{\quad .}
\end{align}

In the standardized parametrization, the approximate covariance always has the identical structure:
\begin{align}
\label{eq:approximate_covariance}
\Xi(\widehat{\xi}) = \left(J(\widehat{\xi})^\dagger I_d(f(\widehat{\xi})) J(\widehat{\xi}) + \mathbb{1}\right)^{-1}\text{\quad .}
\end{align}
To draw samples according to the covariance $\Xi$, we start by drawing from the constituents of $\Xi^{-1}$:
\begin{align}
\label{eq:lh_sampling}
n_* &\sim \mathcal{G}\left(n \vert 0,I_d\left(f(\widehat{\xi})\right)\right) \\
\label{eq:pr_sampling}
\eta_* &\sim \mathcal{G}(\eta\vert 0, \mathbb{1})\\
\label{eq:po_sampling}
\Delta \phi_*&= J(\widehat{\xi})^\dagger n_* + \eta_*\text{\quad .}
\end{align}
This requires the likelihood Fisher metric to be accessible in the eigenbasis, which is the case for example with independently sampled data points. Now $\Delta \phi_* \sim \mathcal{G}(\phi\vert 0, \Xi(\widehat{\xi})^{-1})$ is distributed according to the inverse covariance. Using Eq.~\ref{eq:inverse_samples}, we numerically apply the covariance itself to this sample via conjugate gradient, following Eq.~\ref{eq:QuadraticProblem}. 
\begin{align}
\label{eq:sampling}
\Delta \xi_* &= \Xi(\widehat{\xi})\Delta \phi_*\text{\quad , and therefore}\\
\Delta\xi_* &\sim \mathcal{G}(\xi \vert 0, \Xi(\widehat{\xi})) \text{\quad .}
\end{align}
These samples are drawn from a zero-mean Gaussian with the correct covariance. Our overall approximation will not be zero-centered, but this corresponds only to a shift  by the mean vector $\bar{\xi}$.
\begin{align}
\bar{\xi} +\Delta\xi_*  \sim \mathcal{G}(\xi\vert \bar{\xi}, \Xi(\widehat{\xi}))\text{\quad .}
\end{align}
This is essentially the reparametrization trick \citep{AEVB}, which allows us to stochastically approximate the KL-divergence, while still providing gradients to the variational parameters, in our case only $\bar{\xi}$.

Using this procedure, we can draw a set of independent samples from the approximate posterior distribution, which allows us to statistically estimate the Kullback-Leibler divergence. Drawing these samples can be relatively costly, as every sample requires the numerical inversion of the inverse covariance, but drawing several samples is completely independent from each other and it can be done in parallel. Overall we might want to use as little samples as possible to reduce the numerical effort.

Another important point is how accurately the numerical inversion is performed. Of course, a higher accuracy results in better samples, but also requires more computations. The effect of un-converged samples depends mainly of the starting position of the conjugate gradient. Roughly speaking, the conjugate gradient method updates first the most informative directions. These correspond to the smallest eigenvalues of the covariance. 

Starting at a sample from the standard Gaussian prior, after $n$ iterations of the conjugate gradient at least the $n$ most informative directions are updated towards the posterior uncertainty, whereas the remaining directions still have the prior variance. Overall, un-converged samples will have the correct variance for the best informed directions and the remaining directions over-estimate the actual variance, encoded in the approximate covariance. This behavior safeguards us from a number of pitfalls that can be observed in MAP estimators by underestimating, or ignoring uncertainty variance. We will explore the impact of this accuracy on the method in one of the numerical examples in the end.

\subsection{Antithetic sampling}
We will perform a stochastic estimate the KL-divergence and its gradient. This estimate is subject to sampling noise, which is reduced by increasing the number of samples. This increase will significantly impact the performance of the method. An additional way to reduce the variance of the estimates is antithetic sampling  \citep{kroese2013handbook}. Here anti-correlated samples are used to obtain better estimates. Because we use a Gaussian approximation to the posterior, generating an additional, totally anti-correlated sample is trivial, as $ \bar{\xi} - \Delta\xi_*^{i}$ is an equally valid sample as $ \bar{\xi} + \Delta\xi_*^{i} $. 
Consider some monotonic function $g(\xi)$. The antithetic estimator $\widehat{g}^{(a)}$ of the function value is the average over the anti-correlated samples. 
\begin{align}
\widehat{g}^{(a)} &=\frac{1}{N} \sum_{i=1}^{N/2} \left( g_-^i + g_+^i \right) \text{\quad , and}\\
\mathrm{Var}\left(\widehat{g}^{(a)}\right) &=\frac{\mathrm{Var}\left(g(\xi) \right)}{N}\left(1+ \varrho_{g_+,g_-}\right)\text{\quad .}
\end{align}
Here we indicate  $g(\bar{\xi} \pm \Delta\xi_*^{i} ) = g_\pm ^i$ and $\varrho_{g_+,g_-}$ is the correlation between the antithetic pairs. The smaller this correlation is, the better the estimate will be. For the parameter mean, i.e. $g(\xi) = \xi$, this variance will completely vanish. For non-linear functions and transformations, the anti-correlation in the samples could be reduced. In the worst case, the pairs are fully correlated, and we fall back to the $N/2$ independent samples in terms of the resulting variance, only wasting computations.  However, only artificially constructed systems seem to be capable of showing such behavior.

 Empirically we found that adding antithetic samples is extremely helpful in stabilizing the algorithm by counterbalancing extreme fluctuations in certain parameters. We will show in the numerical examples that even as little as one single pair of antithetic samples can be sufficient to obtain reasonable results, at least in the early stages of the procedure. This speeds up the overall convergence of the method, reducing the time to draw samples, as well as reducing the overall number of required samples due to lower variance of the estimates.

\section{Metric Gaussian Variational Inference}
At this point we want to summarize the key concepts of Metric Gaussian Variational Inference. 
MGVI performs a series of approximations of a complex posterior with Gaussian distributions. The covariance of the approximating Gaussian is extracted from the local properties of the true posterior, describing the vicinity around the current mean estimate and it consists of the inverse Fisher information metric of likelihood and prior. Given this covariance, the approximate distribution is shifted to better represent the true posterior by minimizing the KL-divergence between truth and approximation with respect to the mean parameter. Given this covariance, the posterior is now optimally approximated by the Gaussian. However, at this new location, the vicinity around the mean might have changed, and we possibly represent the uncertainty better by again adapting the local properties of the true posterior. We iterate this procedure until the mean estimate is self-consistent with the uncertainty estimate.

MGVI cannot capture multi-modal structure in the posterior, as a Gaussian distribution is used to describe it. It also breaks down for severely non-linear models where second order terms of the transformation cannot be neglected. In the limits of small amounts of data, Gaussian posteriors, and the Bayesian central limit, MGVI will provide excellent results. In large-scale problems, certain parameters might be constrained extremely well, whereas others are almost uninformed by the data. Here MGVI can capture both limits simultaneously.

The standardization procedure of the hierarchical model might be optional for models with parameters of infinite support, but more complex models often contain parameters restricted to certain ranges. In this case the posterior cannot be approximated with a Gaussian. Standardization allows to approach these problems as well, as outlined for ADVI \citep{ADVI}. Here we will not treat the special case where MGVI is used in hierarchical coordinates, and we will only discuss the more unified, and structurally simpler case in standard coordinates. This results in non-Gaussian solutions for the original parametrization, as the approximation transforms according to the standardization transformation to the original parameters.

In this parametrization the information of the joint distribution of data and standardized parameters $\xi$ with standardization transformation $f$ always reads
\begin{align}
\mathcal{H}(d,\xi) = \mathcal{H}(d\vert f(\xi)) + \frac{1}{2}\xi^\dagger \mathbb{1} \xi \text{\quad ,}
\end{align}
as outlined in Eq.~\ref{eq:StandardHamiltonian}. We want to variationally approximate the posterior corresponding to this model with a Gaussian distribution of the form (Eq.~\ref{eq:GaussianApproximation}) 
\begin{align}
\widetilde{\mathcal{P}}(\xi \vert \bar{\xi}, \Xi) = \mathcal{G}(\xi\vert\bar{\xi},\Xi) \text{\quad .}
\end{align}
For an initial parameter estimate $\widehat{\xi}$, we construct the initial mean value $\bar{\xi} = \widehat{\xi}$ and  the uncertainty estimate from the local Fisher information metric:
\begin{align}
\Xi = \Xi(\widehat{\xi})= \left(J(\widehat{\xi})^\dagger I_d(f(\widehat{\xi})) J(\widehat{\xi}) + \mathbb{1} \right)^{-1} \text{\quad .}
\end{align}

Here  $ I_d(f(\widehat{\xi}))$ is the Fisher metric of the likelihood and  $J(\widehat{\xi})$ the Jacobian of the standardizing transformation evaluated at the latent parameter estimate $\widehat{\xi}$ and $\mathbb{1}$ the prior metric, the identity operator in standard coordinates. This is a non-diagonal full-rank, positive definite matrix that correlates all parameters with another. We cannot store it explicitly at any time, but its inverse, the precision matrix can be well represented by a collection of sparse, implicit operations. In order to work with the covariance, we do have to rely on numerical operator inversion, as outlined in Sec.~\ref{sec:implicit}. 

Given this covariance, we want to match the Gaussian distribution as closely as possible to the true posterior distribution by minimizing the KL-divergence with respect to $\bar{\xi}$, while keeping the covariance fixed:

\begin{align}
\mathcal{D}_\mathrm{KL}\left(\mathcal{G}(\xi\vert\bar{\xi},\Xi(\widehat{\xi}))\vert\vert \mathcal{P}(\xi\vert d)\right) & \;\widehat{=} \left\langle \mathcal{H}(d,\xi) \right\rangle_{ \mathcal{G}(\xi\vert \bar{\xi}, \Xi(\widehat{\xi}))} \\
& \approx \frac{1}{N} \sum_{i=1}^N \mathcal{H}(d, \bar{\xi}+\Delta\xi_*^i)  \text{, with  } \label{eq:MGVIKL}\\
& \text{\quad} \xi_* \sim \mathcal{G}(\xi\vert 0,\Xi(\widehat{\xi})) \text{ \quad .}
\end{align}
When minimizing only with respect to the mean of a Gaussian, the Shannon entropy term is irrelevant for the KL-divergence, which therefore simplifies to the cross-entropy. We approximate the expectation value with a set of samples drawn from our approximation following the implicit sampling scheme described in Sec.~\ref{sec:sampling}. To minimize the stochastic estimate of the KL-divergence with respect to $\bar{\xi}$, we calculate the gradient, as well using these samples:
\begin{align}
\frac{\partial \mathcal{D}_\mathrm{KL}}{\partial \bar{\xi}} &= \left\langle \frac{\partial \mathcal{H}(d,\xi)}{\partial \xi} \right\rangle_{ \mathcal{G}(\xi \vert \bar{\xi}, \Xi(\widehat{\xi}))}\\
& \approx \frac{1}{N} \sum_{i=1}^N\frac{\partial \mathcal{H}}{\partial \xi}\left(d, \bar{\xi}+\Delta\xi_*^i\right)\text{\quad .}
\label{eq:MGVIgrad}
\end{align}

To efficiently optimize the stochastic estimate of the KL-divergence, we rely on a (relaxed) natural gradient descent \citep{amari1997neural, NaturalGradientReview}. We do have the Fisher information metric of the problem available anyway, so we use it to weight the gradient with the local inverse metric, followed by a line-search along this direction to account for non-quadratic features in the landscape. We repeat this procedure until the KL-divergence is minimized. The Fisher information metric of this stochastic estimate of the loss function is the average of the individual metrics evaluated at the sample location. As the samples collectively move through the landscapes, coupled by the mean, we re-evaluate the averaged metric at intermediate steps towards the minimum. 
\begin{align}
\left\langle \Xi^{-1} \right\rangle(\bar{\xi})\equiv \frac{1}{N}\sum_{i=1}^N  \Xi^{-1}(\bar{\xi}+\Delta\xi_*^i)\text{\quad .}
\label{eq:MGVImet}
\end{align}
The sum of implicit operators is still an implicit operator, and we can approximately apply the inverse of it to the gradient. The result is roughly the natural gradient, which we use as descent direction:
 \begin{align}
 \Delta_{\bar{\xi}}= \left\langle\Xi^{-1}\right\rangle^{-1} \frac{\partial \mathcal{D}_\mathrm{KL}}{\partial \bar{\xi}} \text{\quad .}
 \end{align}
Now that we optimized the KL-divergence for the fixed covariance, we obtained a new parameter estimate in form of the mean of the variational Gaussian. We continue to repeat this procedure until the mean is self-consistent with the uncertainty estimate and it no longer changes.
In Al.~\ref{al:MGVI} we present a sketch of the  MGVI algorithm.\\

\begin{algorithm}
		\caption{Metric Gaussian Variational Inference}
	\KwIn{Data $d$, Likelihood $\mathcal{P}(d\vert \theta)$, Fisher metric $I_d(\theta)$, Standardization $\theta = f(\xi)$}
	Initialize global iteration counter $i = 0$\\
	Initialize $\widehat{\xi}^{(0)} = 0$ or small perturbation \\
	\While{ $\widehat{\xi}$ not converged}{
	Construct covariance approximation $\Xi(\widehat{\xi}^{(i)})$  (Eq.~\ref{eq:approximate_covariance})\\
	\For{$N$ samples}{
		Draw sample $n_* \sim \mathcal{G}(n \vert 0,I_d(f(\widehat{\xi}^{(i)})))$ (Eq.~\ref{eq:lh_sampling}) \\
		Draw sample $\eta_* \sim \mathcal{G}(\eta \vert 0,\mathbb{1})$ (Eq.~\ref{eq:pr_sampling})\\
		Calculate $\Delta\phi_* = J(\widehat{\xi}^{(i)})^\dagger n_* + \eta_*$ (Eq.~\ref{eq:po_sampling})\\
		Solve $\Delta\xi_* = \Xi(\widehat{\xi}^{(i)}) \Delta\phi_*$ implicitly via numerical inversion (Eq.~\ref{eq:sampling})\\
		Store $\Delta\xi_*$ (and  $-\Delta\xi_*$) in the set of samples $\{\Delta\xi_*\}^{(i)}_N$ 
}
	Set $\bar{\xi}^{(0)} \leftarrow \widehat{\xi}^{(i)}$ \\
	Initialize local iteration counter $j = 0$\\
	\While{$\mathcal{D}_{\mathrm{KL}}$ not minimized}{
		Estimate $\frac{\partial \mathcal{D}^{(i)}_\mathrm{KL}}{\partial \bar{\xi}}(\bar{\xi}^{(j)})$ with samples  $\{\Delta\xi_*\}^{(i)}_N$ (Eq.~\ref{eq:MGVIgrad})\\
		Construct Fisher information metric $\left\langle \Xi^{(i)-1} \right\rangle(\bar{\xi}^{(j)})$ (Eq.~\ref{eq:MGVImet})\\
		Solve for natural gradient $\Delta^{(j)}_{\bar{\xi}} =\left\langle\Xi^{(i)-1}\right\rangle^{-1}\frac{\partial \mathcal{D}^{(i)}_\mathrm{KL}}{\partial \bar{\xi}}$ implicitly (Eq.~\ref{eq:SolvedProblem})\\
		Find step-length $\eta$ via line search of $\mathcal{D}^{(i)}_\mathrm{KL}\left(\bar{\xi}^{(j)} - \eta \Delta^{(j)}_{\bar{\xi}}\right)$ (Eq.~\ref{eq:MGVIKL})\\
		Update $\bar{\xi}^{(j+1)} \leftarrow \bar{\xi}^{(j)} - \eta \Delta^{(j)}_{\bar{\xi}}$ \\
		Increment local iteration counter $j$
}
	Update $\widehat{\xi}^{(i+1)} \leftarrow \bar{\xi}^{(j)}$ \\
	Increment global iteration counter $i$

}
	\KwRet $\widehat{\xi} \leftarrow \widehat{\xi}^{(i)}$ \\
	\KwRet $\{ \Delta\xi_*\}_N\leftarrow\{\Delta\xi_*\}^{(i-1)}_N$ 
	\label{al:MGVI}
\end{algorithm}
Initializing the parameter estimate with zero can be problematic due to vanishing gradients and numerical artifacts, which is resolved by using Gaussian noise with small variance instead. The convergence of $\widehat{\xi}$ can be determined by observing the changes between iterations. Because we use samples to determine all relevant quantities for the optimization, we are always subject to sampling errors. The more samples we use, the more accurate our solutions will be, so we can only convergence within the intrinsic sampling noise, given a number of samples. For more samples, we can achieve deeper convergence, and in practice we will increase the number samples throughout the algorithm. 

Other meta-parameters of the algorithm are the accuracy of the numerical inversion to draw samples, i.e. the number of performed conjugate gradient steps, and how well we optimize the KL-divergence for a given parameter estimate. We will illustrate and discuss the impact of certain choices in the second numerical example. 
To use antithetical sampling for better stochastic estimates, one simply also includes $-\Delta\xi_*$ to the set of samples $\{\Delta\xi_*\}^{(i)}_N$.

In Al.~\ref{al:MGVI} we use an approximate Relaxed Newton scheme to optimize the KL-divergence in the inner while-loop, but in principle any optimization scheme could be used. Especially the Newton-CG algorithm also performs well. In any case, we recommend to make use of the Fisher information for the optimization, as we have all ingredients available anyway and it can provide enormous speed-ups in high-dimensional problems.

In the end, MGVI provides a parameter estimate $\widehat{\xi}$ and a set of samples $\{ \Delta\xi_*\}_N$, which together are samples from the approximate Gaussian distribution. This parameter estimate is self-consistent with the uncertainty estimate provided by the used approximation. The samples can then be used to propagate the uncertainty to any quantity of interest. 

\section{Numerical examples}
We will demonstrate MGVI in several examples, showcasing a diverse spectrum of applications, of both, synthetic- and real-data applications. We compare our approach to MAP estimates, HMC, mean-, and full-covariance ADVI. 

In the first example we discuss the problem of inferring the rate of a Poisson distribution described as a log-Gaussian process. This process exhibits a squared exponential kernel of known amplitude and width. 

The second example demonstrates the well behaved scaling of MGVI with the problem size, as well as its viability in the context of complex models with conceptually distinct parameters.
Here we discuss the problem of binary Gaussian process classification in two dimensions with non-parametric kernel estimation. The data consists of binary values with associated location. The likelihood is the Bernoulli distribution and its rate is linked through a sigmoid function to a Gaussian process with unknown kernel.
The size of the posterior exceeds one million model parameters. The computation and storage of a dense covariance as used by ADVI with  a full covariance is computationally unfeasible as it would require to maintain $10^{12}$ entries.
This problem size and complexity prohibits validation with the other methods and we compare the result of MGVI only to mean-field ADVI, as well as the underlying truth. Additionally, we showcase and discuss the impact of several important meta-parameter choices on a smaller scale version of this problem.

The third example solves a non-negative matrix factorization problem on the Frey Face data-set assuming a Gamma-Poisson model.

In the last example we explore a hierarchical logistic regression problem involving polling data of the 1988 US presidential election with several regressors. We use a simplified model to again validate MGVI against HMC, as well as all the other methods, and a more complex model to discuss the convergence behavior of MGVI.

For an even larger numerical example with real data we refer to \citet{leike2019charting}, where a three dimensional dust map in our galactic vicinity is reconstructed in a resolution of $256^3$ voxels from dust absorption measures and star locations obtained by the Gaia satellite. This problem involved a truncated Gaussian likelihood with log-normal prior and unknown kernel, analogous to the model used in the second example. The reconstruction was conducted using the here described MGVI procedure.

MGVI is further used by \citet{Radiocalibration} to jointly calibrate a radio interferometer data set and perform its imaging. This allows to use the stationarity of the science target to obtain better calibration solutions, which in turn lead to better image reconstructions. 

Another application of MGVI with multiple components and data fusion in spherical geometries is outlined in \citet{NewFaraday}, where the Galactic Faraday depth sky is reconstructed from a rotation measure catalogue and free-free emission data.

Finally, \citet{CausalFields} formulate locality and causality priors to learn the dynamics of a field from noisy and incomplete observation. Again, the inference of this field together with its dynamics is done via MGVI. 
\paragraph{Performance metrics:}
\label{sec:performance_metrics}
Comparing different methods against each other is not straight-forward. The preference of one method over another depends on many different factors, e.g. required accuracy, uncertainty quantification, or  available resources. Any performance metric will only tell something about a certain aspect of the methods and we do not have a universal scale available to strictly determine the superiority of one method over another. 

Ideally we want to validate against the true posterior distribution, but usually we do not have it available. MCMC methods allow to draw samples from the true posterior distribution, requiring large computational resources. Where it is feasible, we will use HMC as a reference, comparing the other methods against, but this restricts us to relatively small inference problems.

To explore the high-dimensional settings, for which MGVI was developed, we have to use a different approach. In real-world applications we do not know the true parameters underlying the data, but in a simulation we do. Performing the inference on such simulated data sets, we can always use the ground truth as reference scale and explore how well a method performs.

A simple metric in such a setting is  the root mean squared error (RMS) of the reconstruction. For some scalar estimator $\widehat{x}$, for example the model evaluated at the mean,  it reads:
\begin{align}
\mathrm{RMS} = \sqrt{\frac{1}{N}\sum_i \left(x_{\mathrm{true}}^{(i)} - \widehat{x}^{(i)}\right)^2}\text{\quad .}
\end{align}
In many applications the goal is to get as closely as possible to some underlying truth, and therefore a lower RMS error should correspond to a better result.

Another quality criterion of a method is the capability to accurately estimate its uncertainty associated with the prediction. Large deviations to the ground truth are acceptable in cases a large variance is expected. For this we weight the absolute residual with the predicted standard deviation, corresponding to the average significance of the residual in terms of standard deviations:
\begin{align}
\mathrm{AS}  = \frac{1}{N} \sum_i  \frac{1}{\widehat{\sigma}^{(i)}} \vert x_\mathrm{true}^{(i)} - \widehat{x}^{(i)} \vert \text{\quad .}
\end{align}
In the case of a Gaussian posterior this quantity should be close to $1$, expressing how significant on average the ground truth is, given in units of standard deviations. The posterior distributions we investigate will not be Gaussian, especially due to the non-linear transformations involved, but it should still provide an insight into the behavior of methods relative to each other, as long as the posterior resembles remotely a Gaussian and the non-linearity is not too extreme. 

In large real-data applications sampling is unfeasible and the ground truth is unknown. In such cases we split the total data in a small sub-set $d'$ for cross-validation of the approximation by evaluating how likely these reference data appear and use only the remaining data $d$ for the inference. For this, we can calculate the predictive likelihood:
\begin{align}
\mathcal{P}(d' \vert d) = \int d\theta \: \mathcal{P}(d' \vert \theta) \mathcal{Q}_{\eta}(\theta) \approx \frac{1}{N} \sum_{\theta_i} \mathcal{P}(d'\vert \theta_i)\text{\quad .}
\end{align}
Here $\theta_i$ are samples drawn from the approximate distribution $ \mathcal{Q}_{\eta}(\theta)$ fitted to the posterior given the remaining data $d$. It measures how predictive the obtained distribution is for the reference data. Generally a large value tells us that we can well extrapolate towards unobserved regions, which usually is desired. Nevertheless, this performance metric punishes uncertainty in a prediction. To see this, consider the maximum likelihood solution on the reference data. It is a point estimate and maximizes, by definition, the predictive likelihood. Now consider an uncertainty around this point, for example in form of a Laplace distribution. Every sample drawn from this presumably better approximation will have a lower predictive likelihood, and will therefore appear worse in this metric. We will encounter such a scenario in our examples and to make the comparison more fair, we will also state the predictive likelihood evaluated only at the latent mean parameter, corresponding to a best guess.

\subsection{Poisson log-normal}
\subsubsection{Setup}
\begin{figure}
 	\centering
 	\includegraphics[scale=0.7]{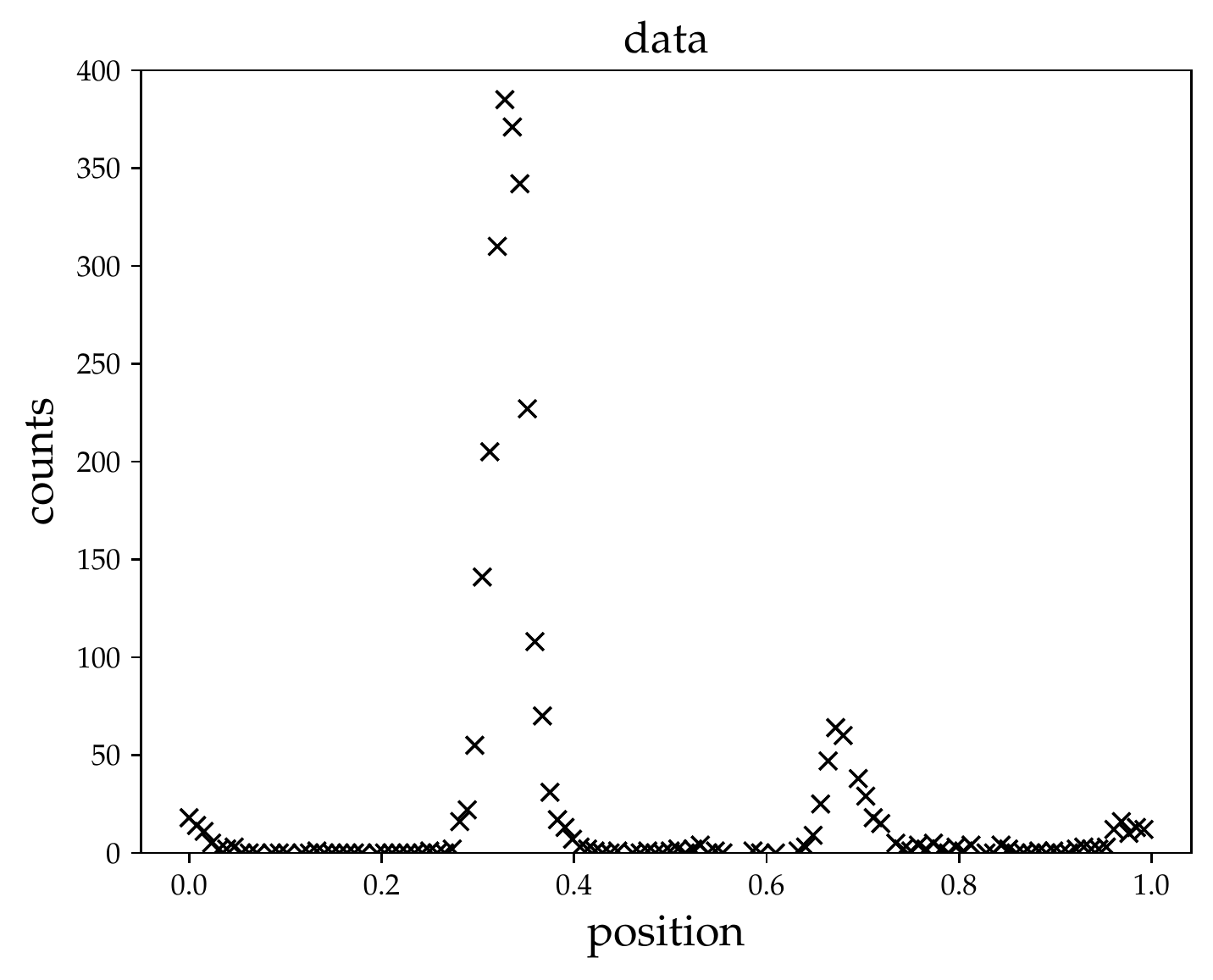}
 	\caption{A Poisson realization drawn according to a log-normal process with squared exponential kernel on linear scale.}
 	\label{fig:data_1}
\end{figure}

In this example we discuss the inference of the rate $\lambda$ of a Poisson likelihood providing count data $d$, where the logarithmic rate is modeled as a Gaussian process with squared exponential kernel of known amplitude and width. 
The count data for our experiment is displayed in Fig.~\ref{fig:data_1}.
The Poisson likelihood reads:

\begin{align}
\mathcal{P}(d\vert \lambda) = \prod_i \mathcal{P}(d_i \vert \lambda_i) \text{\quad , with} \\
\mathcal{P}(d_i \vert \lambda_i) = \frac{\lambda_i^{d_i} e^{-\lambda_i}}{d_i!}\text{\quad .}
\end{align}
Its Fisher information metric with respect to this rate parameter is:
\begin{align}
\label{poisson-metric}
I_d(\lambda) = \widetilde{\lambda}^{-1}\text{\quad .}
\end{align}

This is a diagonal matrix, indicated by the tilde, in the data space with the inverse of the rate $\lambda$ on its diagonal. A tilde over a vector raises it to diagonal matrix, i.e. $ \widetilde{a}_{ij} = \delta_{ij}a_i$. 
The rate is expressed in terms of the exponential of a Gaussian process $\lambda = R e^s$ with prior distribution $\mathcal{P}(s) = \mathcal{G}(s\vert 0,S)$ and some linear response operator $R$. 
Assuming a stationary, or homogeneous  and isotropic kernel, the kernel can be expressed in terms of a spectral density in the harmonic domain, i.e. $S = \mathbb{F}^{-1}  \widetilde{\mathbb{P}{p}}\, \mathbb{F} $ where $\mathbb{F}$ indicates the Fourier transformation, $\mathbb{P}^{\dagger}$ is the projection of the  one-dimensional spectral density onto the Fourier space of the signal coordinates, also one-dimensional in this example, but in general multi-dimensional.  Here $p(k) = \sqrt{2\pi}\sigma^2 l \:e^{-2\pi l^2k^2} $ represents  the squared exponential, or Gaussian, correlation kernel in Fourier space (in one dimension). 
The parameter $l$ is a characteristic length-scale, $\sigma^2$ a variance parameter  and $k$ is the harmonic coordinate. This defines the mathematical setup of this first example. 

The next step is to standardize.
As the prior is already Gaussian, we simply have to identify $S = A A^\dagger$ with $A = \mathbb{F}^{-1}  \widetilde{\mathbb{P}p^{\frac{1}{2}}}$ and rewrite $s = A \xi$. 
With this reparametrization we express the information of the problem for a given spectrum as
\begin{align}
\label{eq:example_1}
\mathcal{H}(d,\xi) \:\widehat{=}\:-d^\dagger \mathrm{ln} R e^{A\xi} + 1^\dagger R e^{A\xi} + \frac{1}{2} \xi^\dagger \mathbb{1} \xi\text{\quad .}
\end{align}
The $1^\dagger$ indicates a scalar product with the one vector, corresponding to the integration over the space. The overall standardization reads
\begin{align}
 \lambda&=f(\xi) \\
 &=R e^{A\xi} \text{\quad .}
\end{align}
This function allows us to build the local approximation to the covariance. Here the parameter dependence is still relatively simple and the Jacobian can be calculated by hand.
 \begin{align}
 \Xi^{-1} &= J(\widehat{\xi})^\dagger  \widetilde{f(\widehat{\xi})^{-1}} J(\widehat{\xi}) + \mathbb{1} \\
& = A^\dagger  \widetilde{e^{A\widehat{\xi}}}^\dagger R^\dagger  \widetilde{\frac{1}{Re^{A\widehat{\xi}}}}R  \widetilde{e^{A\widehat{\xi}}} A +\mathbb{1}  \text{\quad .}
 \end{align}
We see that the metric is composed out  of a collection of operators that can be simply implemented and combined.
 
Now, we approximate the posterior probability implied by the model as described by Eq.~\ref{eq:example_1} using MGVI.
 We start the optimization with one single pair of antithetic samples. Initially we perform three natural gradient steps and use 25 conjugate gradient iterations to draw the sample. After twenty global iterations we start to increase the number of samples and natural gradient steps by one until the thirtieth iteration and steadily increase the sampling accuracy by a total factor of four. Initially we do not want to waste computations for unnecessary accuracy and this purely heuristic scheme is derived from the meta-parameter discussion of the next example. 
 
The problem, as well as MGVI, mean-field (mf-) and full-covariance (fc-) ADVI \citep{ADVI}, HMC \citep{HMC}, and the Laplace approximation are implemented within Python using the NIFTy5\footnote{NIFTy documentation: \url{http://ift.pages.mpcdf.de/NIFTy/}\\ NIFTy code: \url{https://gitlab.mpcdf.mpg.de/ift/NIFTy}} package \citep{Nifty, Nifty3,Nifty5}. 

The posterior samples obtained from HMC will serve as the reference in the validation of our approach. We run five HMC chains in the standard coordinates and obtain a minimal effective sample size of $\mathrm{ESS}_{\mathrm{min}}=108$ and average $\mathrm{ESS}_{\mathrm{mean}}=220$. The average Gelman-Rubin test statistic for the five chains is $\widehat{R}_{\mathrm{mean}}=1.016$ with maximum $\widehat{R}_{\mathrm{max}}=1.052$.

For ADVI we perform both, a fully parametrized covariance, as well as a mean-field approximation, estimating only a diagonal covariance. Using the full covariance limits the possible problem size and we will stick to $128$ parameters to describe the Gaussian process, as well as $128$ equidistant data points. For the optimization procedure we follow the stochastic gradient descent scheme proposed by \citet{ADVI}. 

The data is is drawn according to the model and the realization is shown in Fig.~\ref{fig:data_1}. We monitor the performance and convergence by withholding $10\%$ of the data points and calculating the predictive likelihood of the intermediate result.
 
\begin{figure}
 	\centering
 	\includegraphics[width=\textwidth]{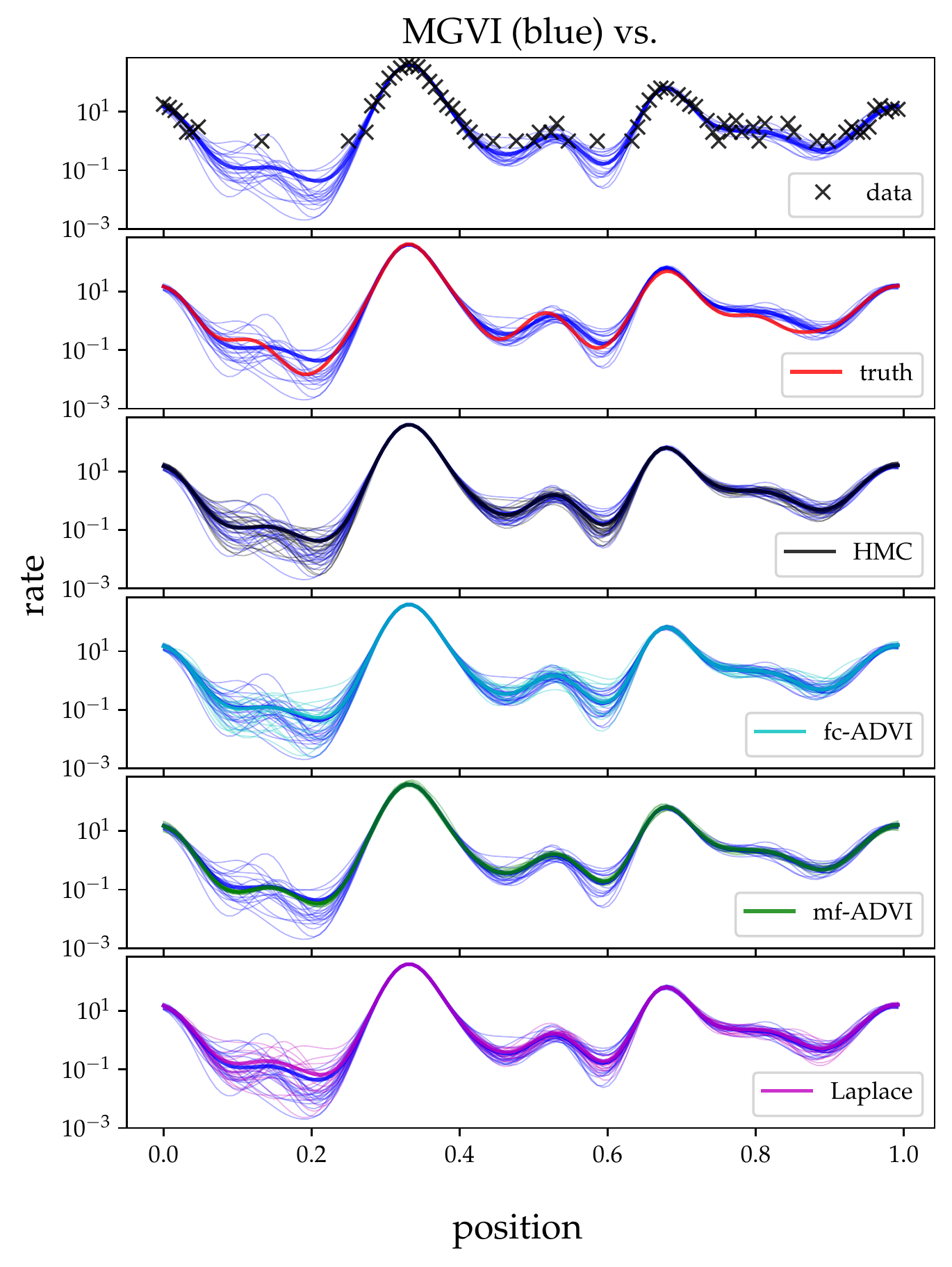}
 	\caption{Reconstructed rates and posterior samples provided by MGVI in comparison to those from various other methods.}
 	\label{fig:mean_results}
\end{figure}

\subsubsection{Results}
All methods recover the underlying rate quite well. 
The obtained rates $\lambda = Re^{A\xi}$ are shown in Fig.~\ref{fig:mean_results} for MGVI and all the other methods. 
The uncertainty of the different estimates are indicated by a set of posterior samples drawn around their corresponding mean rates for all methods. Visually, all methods, except mean-field ADVI, provide similar results. The later severely underestimates the true posterior variance in areas of large uncertainty and overestimates it in regions well-determined by the data.

We also note that overall the relative uncertainty is higher in regions of low counts and smaller in regions of high counts. This is expected from a Poisson likelihood, as its variance $\sigma^2_d$ is equal to its rate $\lambda$ and therefore the relative uncertainty increases with decreasing rate, $\sigma_d/\lambda = 1/\sqrt{\lambda}$.

\begin{figure}
 	\centering
 	\begin{subfigure}[b]{0.49\textwidth}
 		\includegraphics[width=\textwidth]{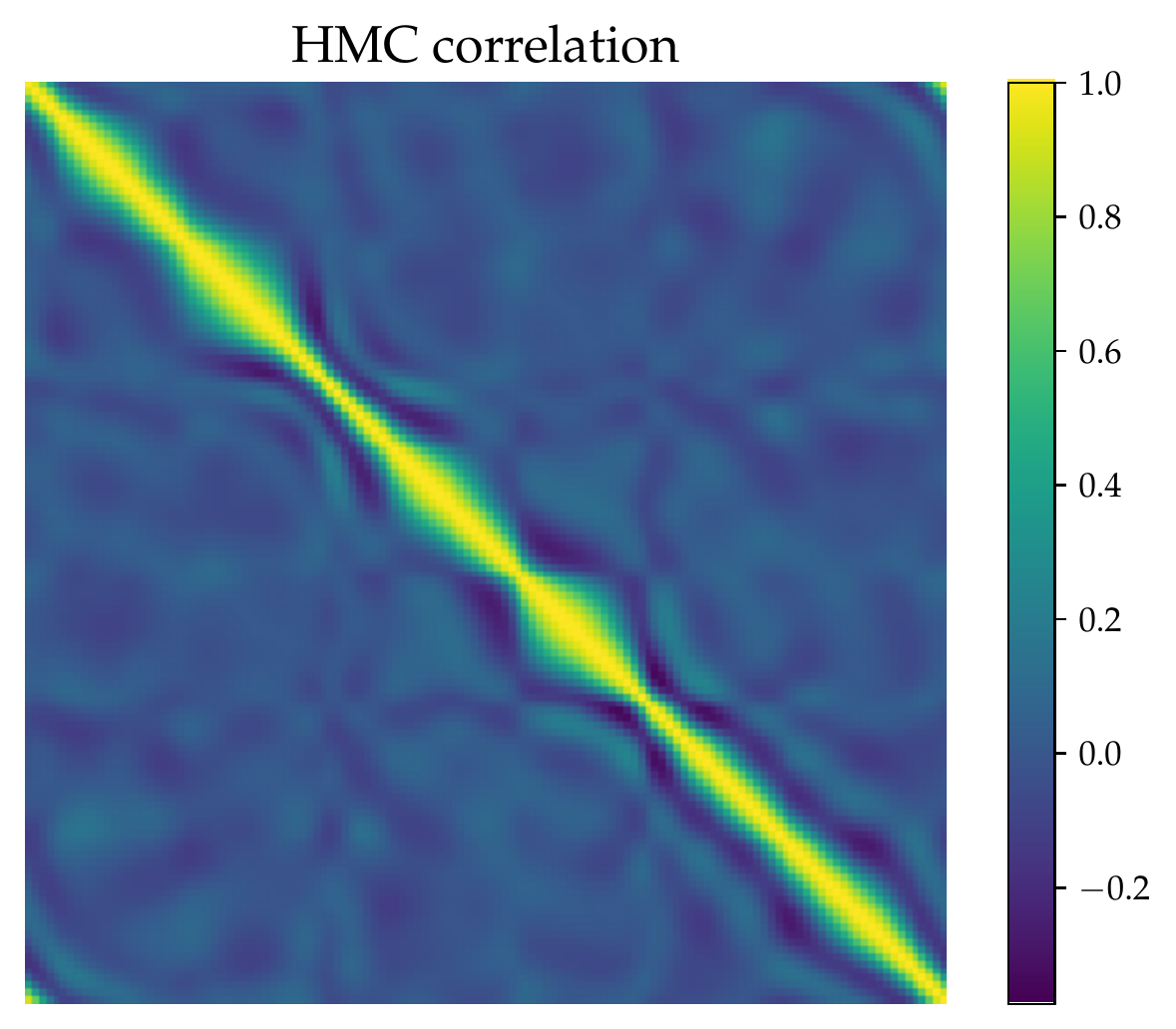}
 	\end{subfigure}
 	\begin{subfigure}[b]{0.49\textwidth}
 		\includegraphics[width=\textwidth]{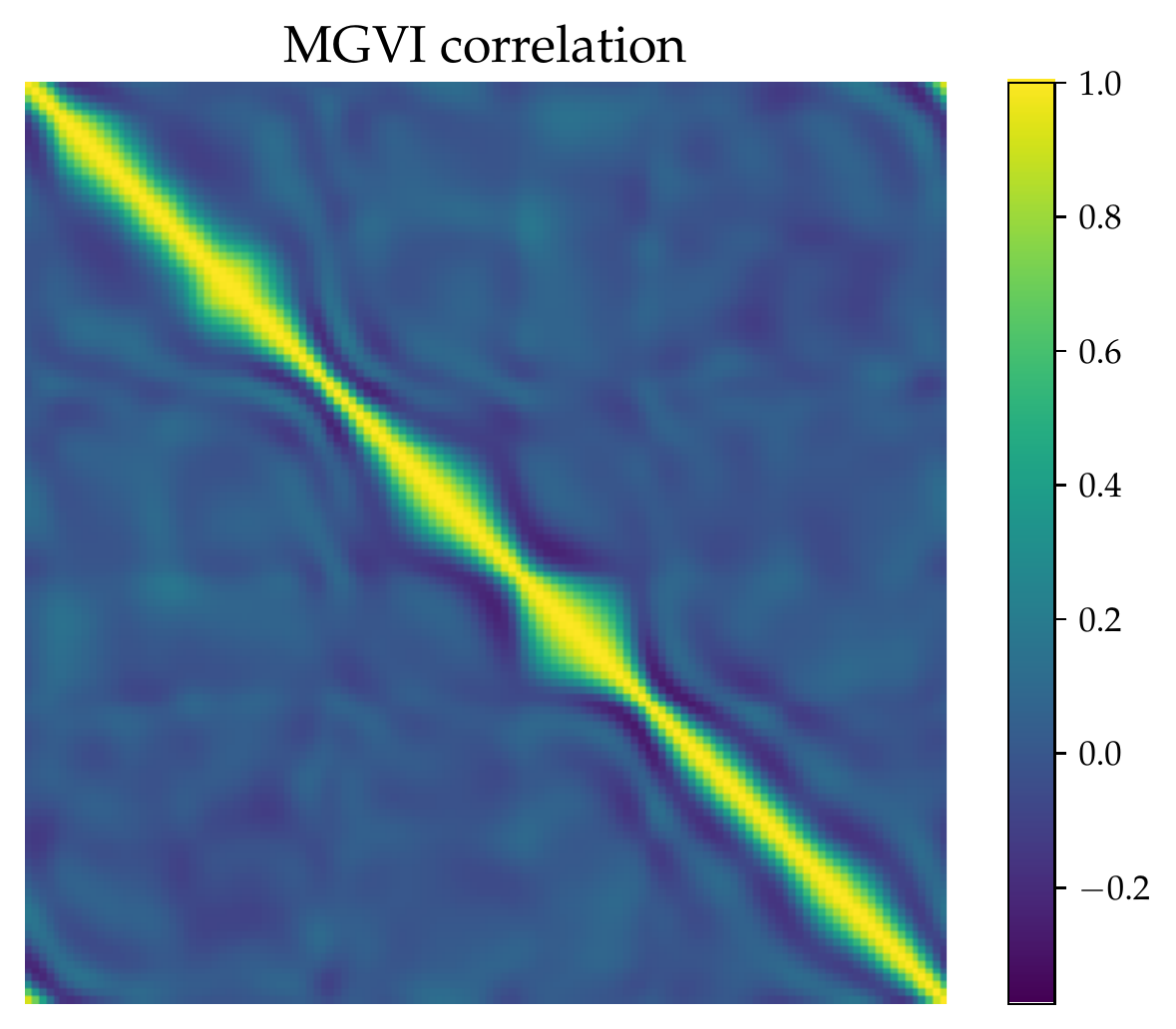}
 	\end{subfigure}
 	\begin{subfigure}[b]{0.32\textwidth}
 		\includegraphics[width=\textwidth]{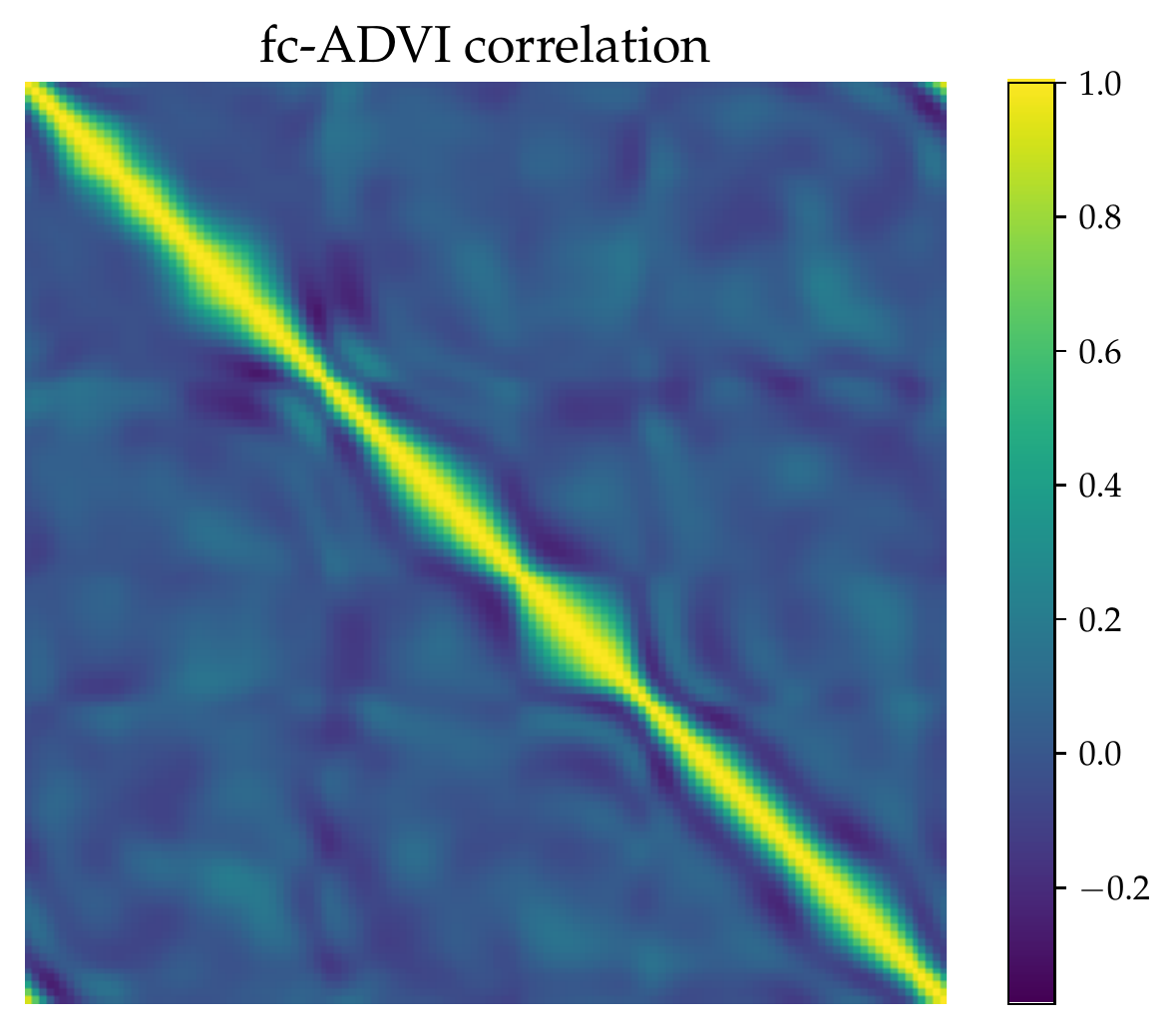}
 	\end{subfigure}
 	\begin{subfigure}[b]{0.32\textwidth}
 		\includegraphics[width=\textwidth]{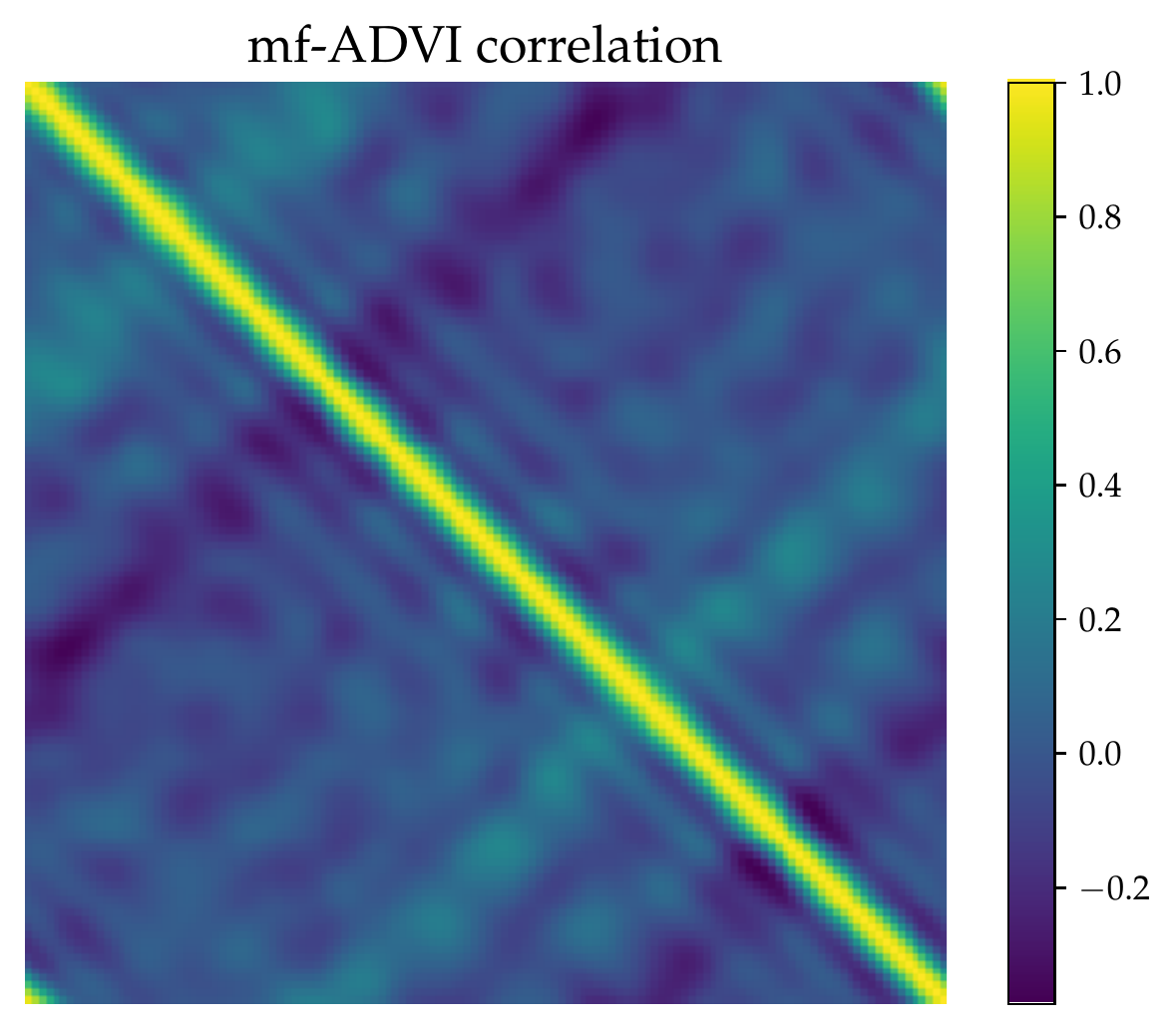}
 	\end{subfigure}
 	\begin{subfigure}[b]{0.32\textwidth}
 		\includegraphics[width=\textwidth]{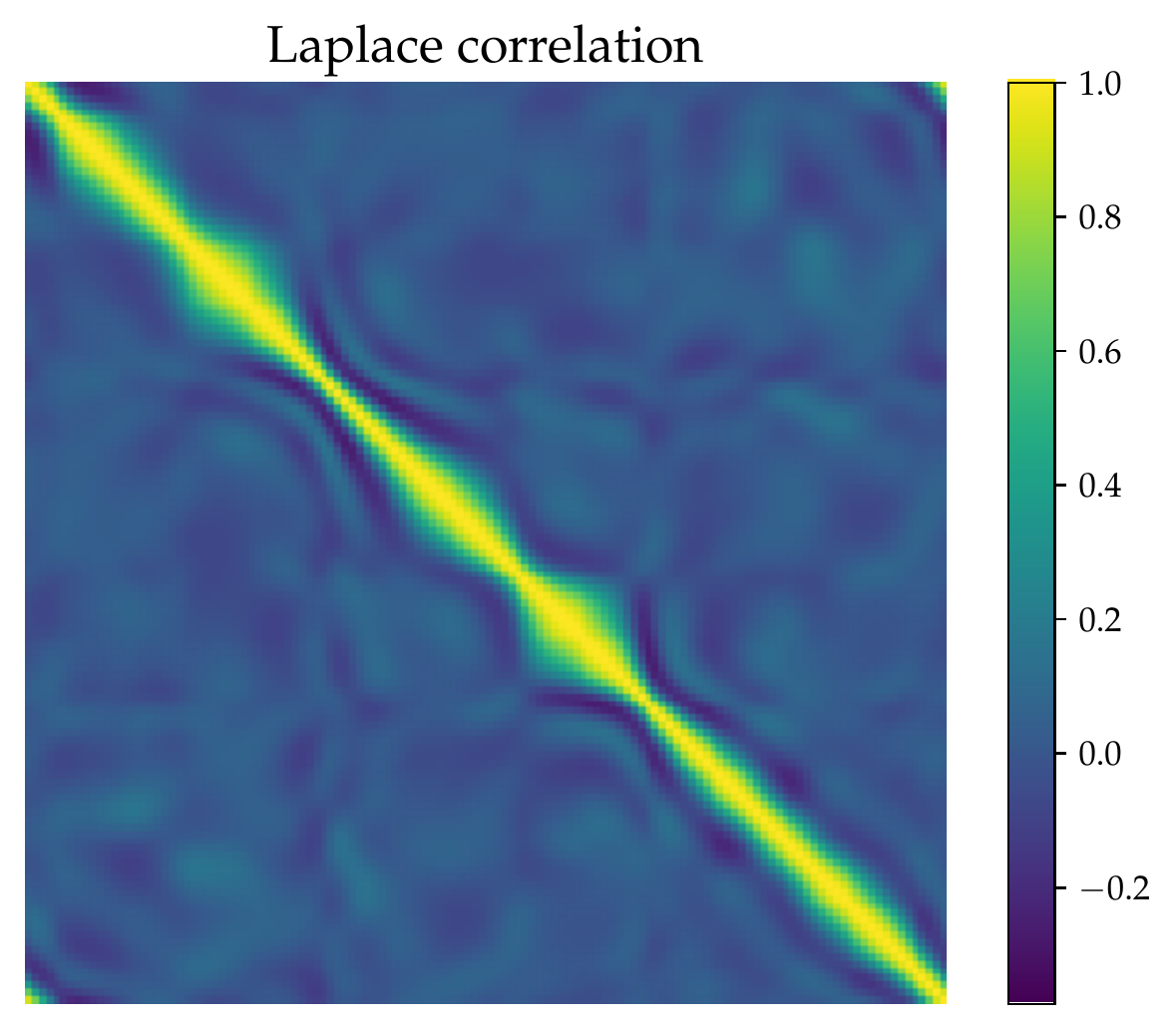}
 	\end{subfigure}
 	\caption{The sampled correlation structures for the different methods. }
 	\label{fig:corr}
 \end{figure}

The two-point correlation matrix of the rate constructed from samples is shown in Fig \ref{fig:corr}. Here again, all correlations do look similar, except mean-field ADVI. The correlation is diagonal dominant and spatially structured. Strong short-range correlations are wrapped by a band of anti-correlation, decaying towards zero for large distances of the points. The periodic boundary condition of this setup is showing up in the top right and bottom left corners. This pattern originates from the squared exponential kernel and is modified by the data. High-signal regions appear here to be more narrow, and the correlation is farther extending in low-rate regions. Here the mf-ADVI correlation structure is agnostic to spatial structure, compromising between high-data and uninformed regions due to the limited expressibility of a mean-field approximation.
 
We provide a snapshot of all methods against HMC in Fig.~\ref{fig:corr}, scattering the values for two locations against each other in three distinct scenarios. This provides a visual impression on how well correlations, as well as the marginal probabilities, are captured by the approximations. MGVI, fc-ADVI and the Laplace approximation match closely to the HMC samples, but mf-ADVI strongly underestimates the variance and, as observed in the correlation matrix, does not express much of the posterior covariance structure.

\begin{figure}

\centering
\begin{subfigure}[b]{0.49\textwidth}
	\includegraphics[width=\textwidth]{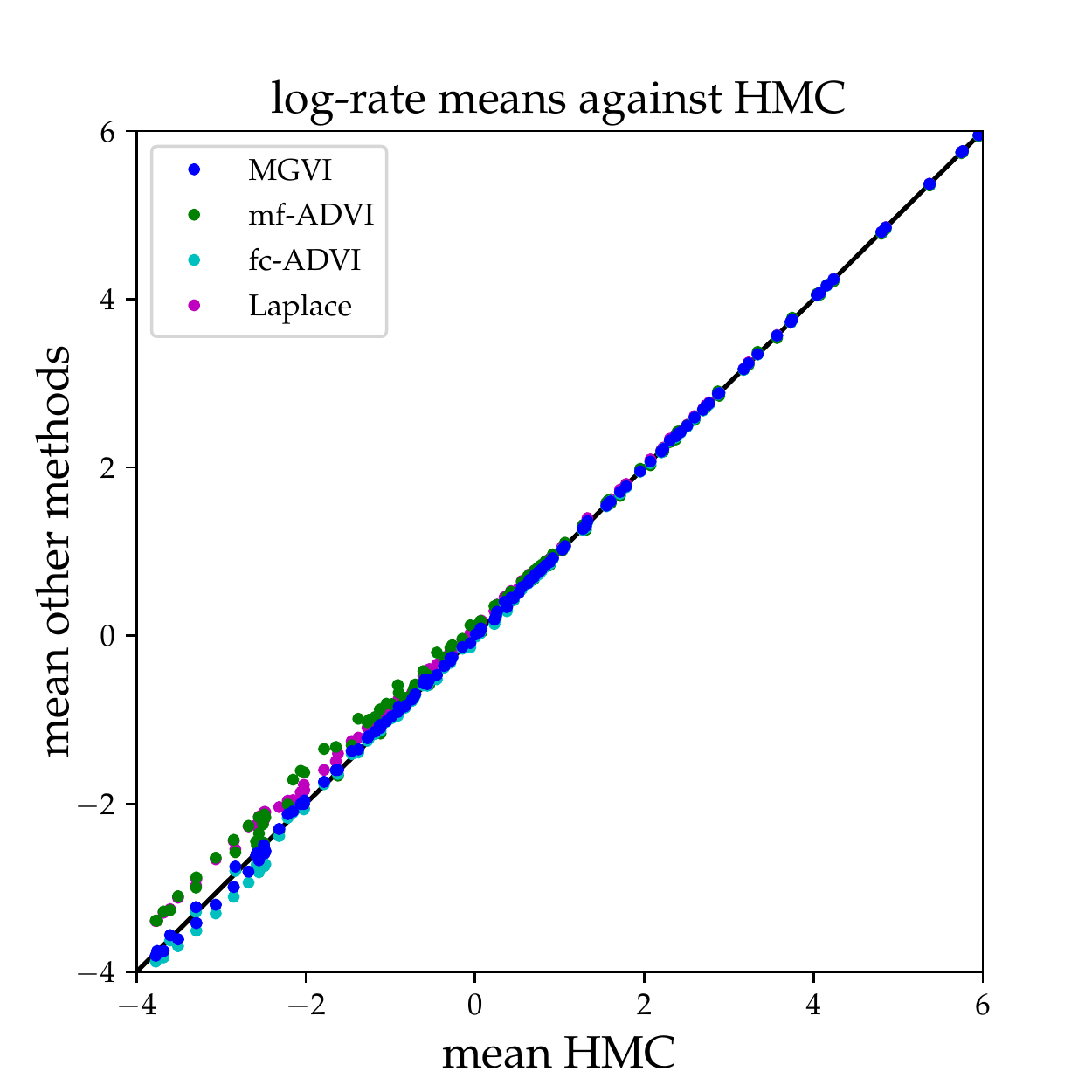}
\end{subfigure}
\begin{subfigure}[b]{0.49\textwidth}
	\includegraphics[width=\textwidth]{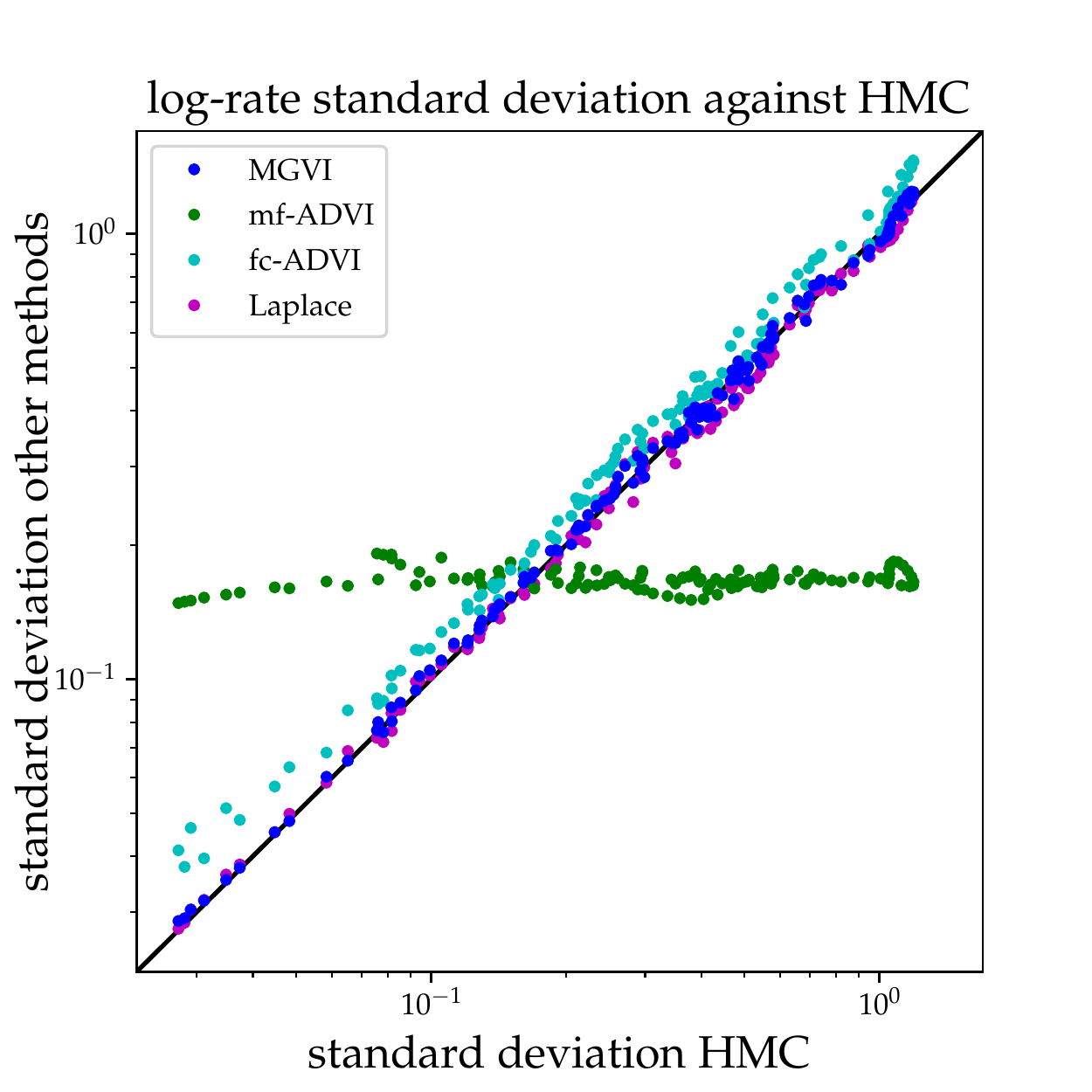}
\end{subfigure}
\caption{The parameter means and standard deviations of the Poisson log-normal problem from the different methods plotted against the HMC results.}
\label{fig:PLN_against_HMC}
\end{figure}

To validate this impression we plot the mean and standard deviation of the log-rate at every location obtained by HMC against the four other methods, as shown in Fig.~\ref{fig:PLN_against_HMC}. Towards large mean rates, all methods agree quite well, as they are well determined by the data. For small rates, the methods differ slightly, but systematically. Here the Laplace approximation, as well as mf-ADVI tends to overestimate the rate, whereas fc-ADVI and MGVI underestimate it, but in that agree well. Thus, it seems that MGVI behaves similarly to fc-ADVI here. 

The standard deviations are structurally more interesting. MGVI and the Laplace approximation agree well with the HMC results and fc-ADVI appears slightly shifted towards overestimated variances, which might be a remnant of insufficient convergence within the assigned computational budget. Finally, mf-ADVI is agnostic to parameter-specific uncertainty and only on average correct, over- and underestimating the standard deviation roughly the same amount of time, as already seen in the correlation matrix and the scatter plots. 

Collapsing these plots down to the RMS error relative to HMC gives Tab.~\ref{tab:PLN_RMS}. Here MGVI is the best method in terms of reproducing the mean, as well as the standard deviation obtained via HMC. Regarding the mean, fc-ADVI gives a similar, but slightly worse result, but mf-ADVI and the Laplace approximation exhibit significantly larger errors. Surprisingly, the latter exhibits one of the best standard deviations, better than fc-ADVI, which still might improve for even longer optimization. The deviations for mf-ADVI are, as expected, just off, dramatically. 

Overall, MGVI seems to be slightly better, but on par in terms of accuracy with the other methods in this example. Its true strength only becomes evident by considering the required computational time to obtain these results, as well as the in principle linear scaling behavior in terms of memory and computations. In the following we will discuss the temporal evolution of several quantities during the optimization for MGVI and both ADVI variants.

 \begin{table}
	\centering
	\caption{The RMS errors of mean and standard deviation of the pixel-wise log-rate with respect to HMC}
	\label{tab:PLN_RMS}
	\begin{tabular}{|c|c|c|c|c|} 
		\hline
		RMS HMC against&MGVI &fc-ADVI&mf-ADVI&Laplace\\
		\hline
		mean & $0.041$ &$0.076$ &$0.16$ &$ 0.17$ \\
		standard deviation & $0.023$ &$0.080$ &$0.42$ &$ 0.032$\\
		\hline
	\end{tabular}
\end{table}

\begin{figure}

	\centering
	\begin{subfigure}[b]{0.32\textwidth}
		\includegraphics[width=\textwidth]{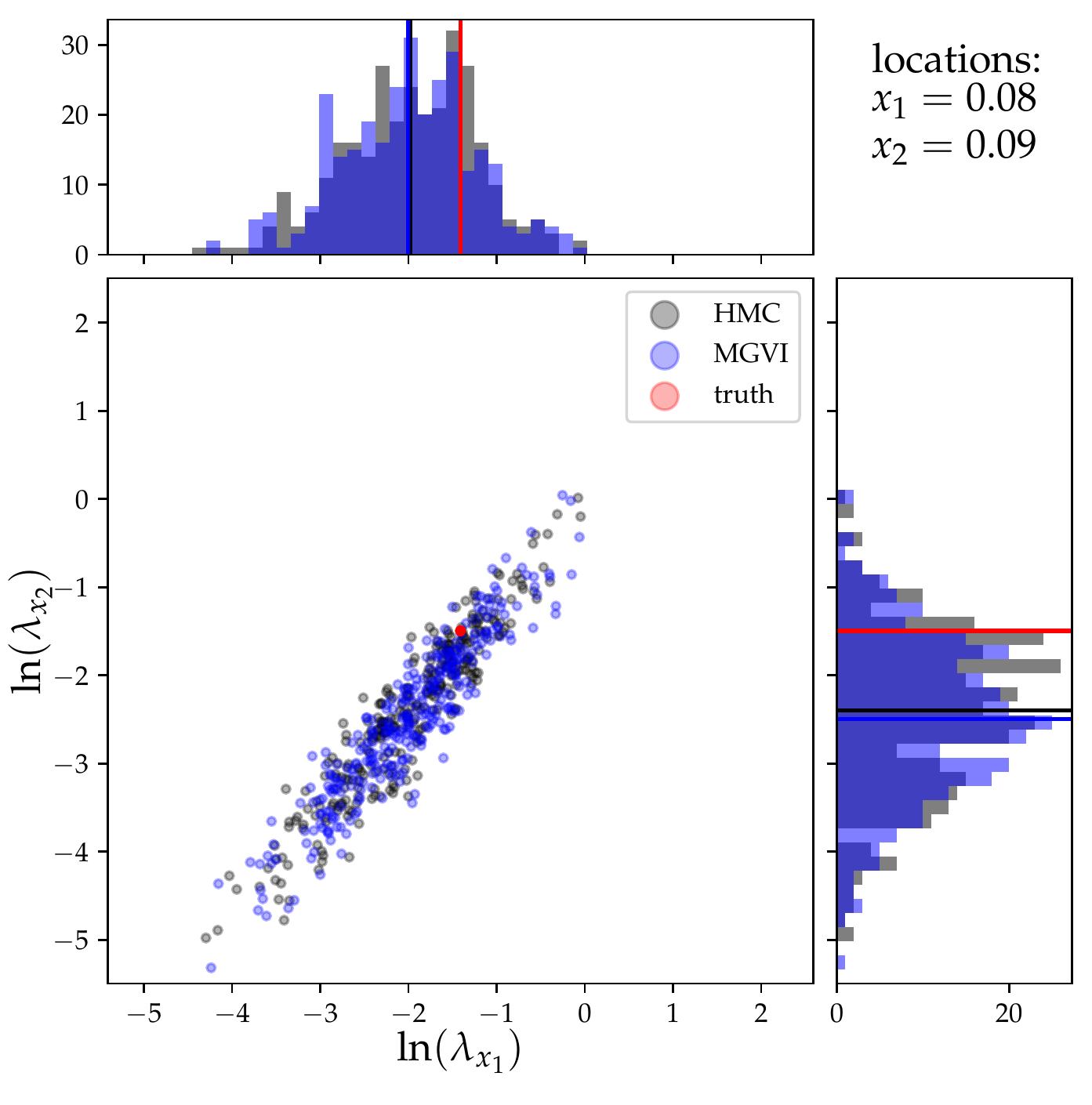}
	\end{subfigure}
	\begin{subfigure}[b]{0.32\textwidth}
	\includegraphics[width=\textwidth]{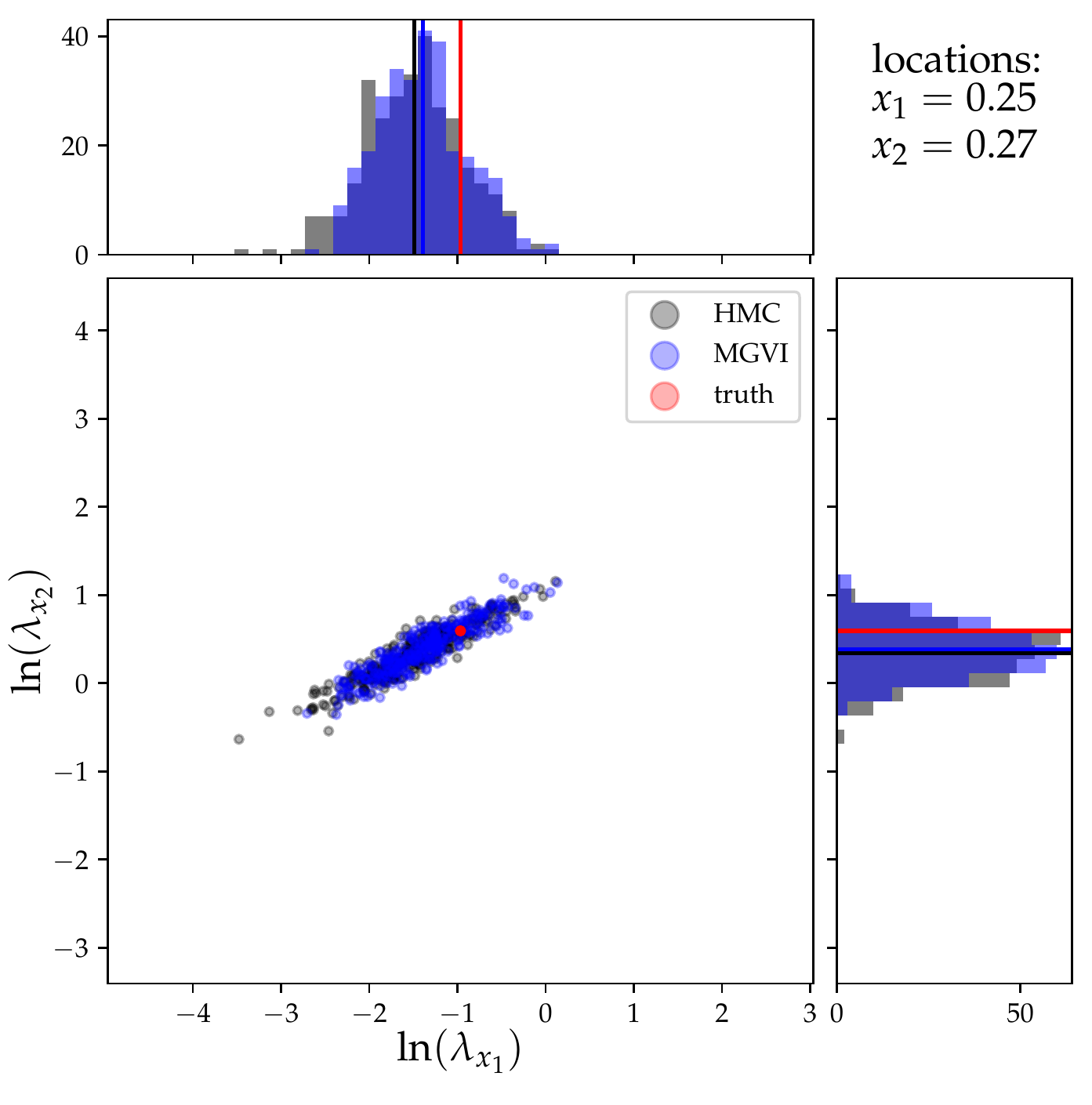}
\end{subfigure}
	\begin{subfigure}[b]{0.32\textwidth}
	\includegraphics[width=\textwidth]{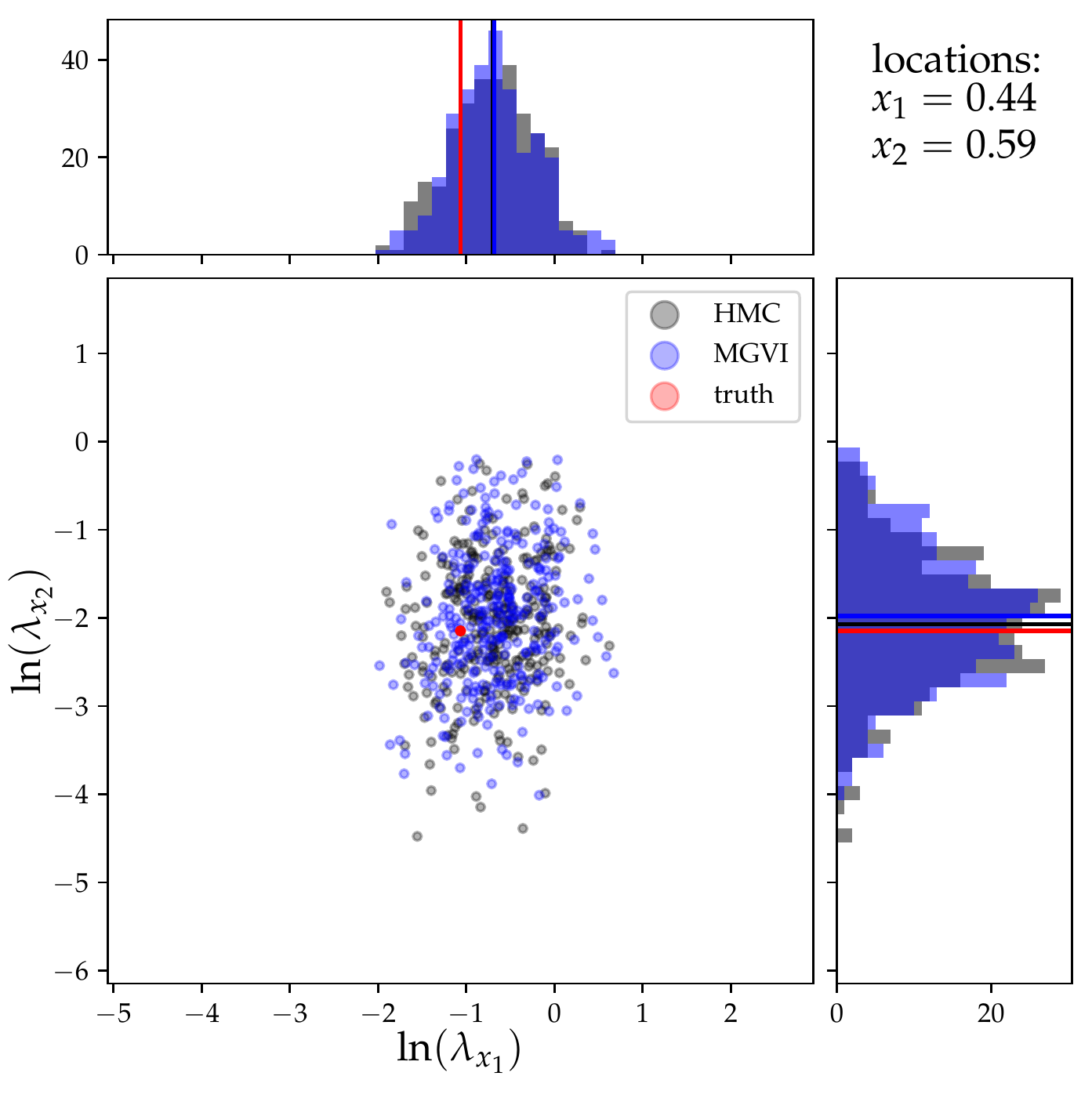}
\end{subfigure}
	\begin{subfigure}[b]{0.32\textwidth}
	\includegraphics[width=\textwidth]{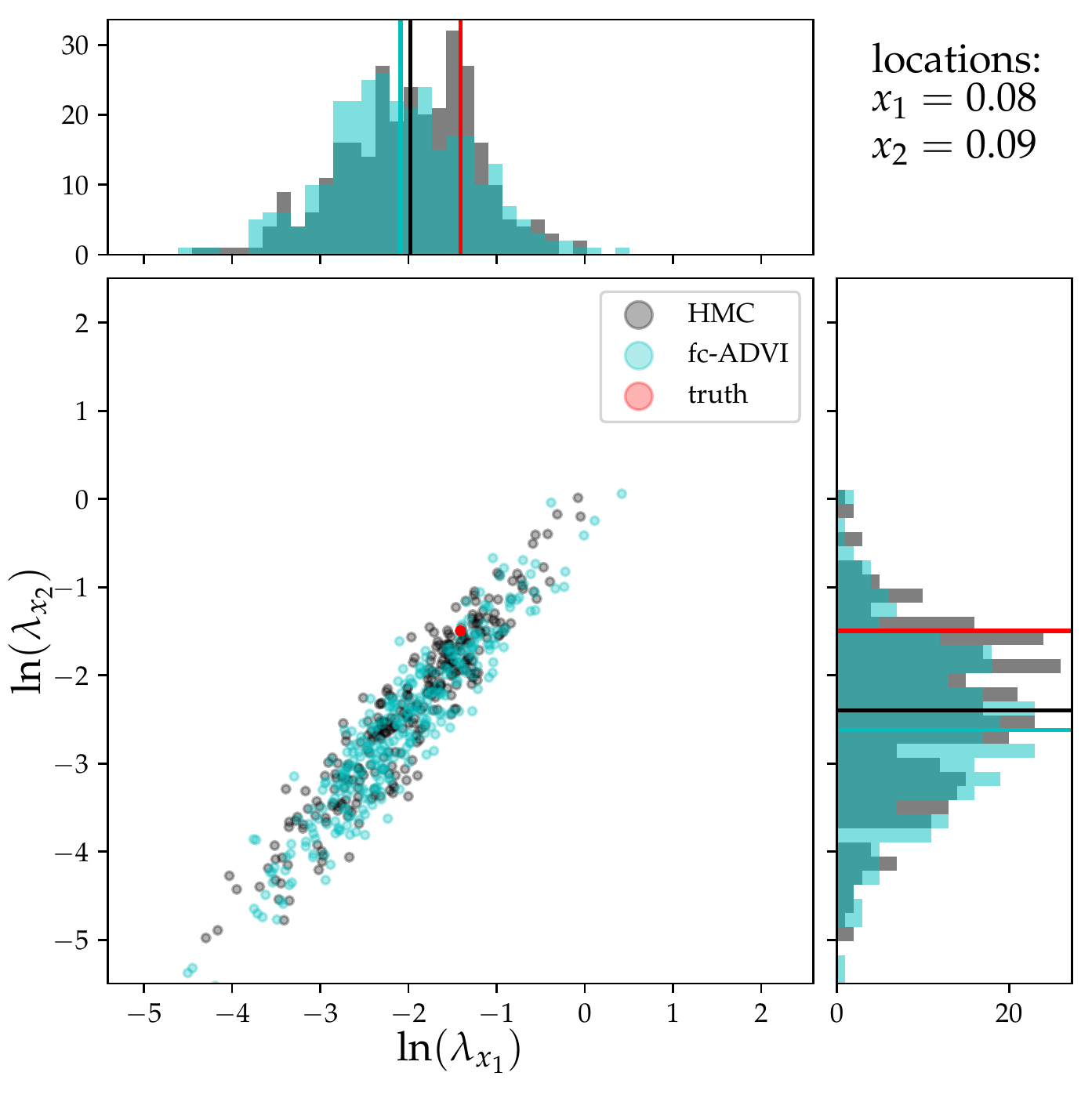}
\end{subfigure}
\begin{subfigure}[b]{0.32\textwidth}
	\includegraphics[width=\textwidth]{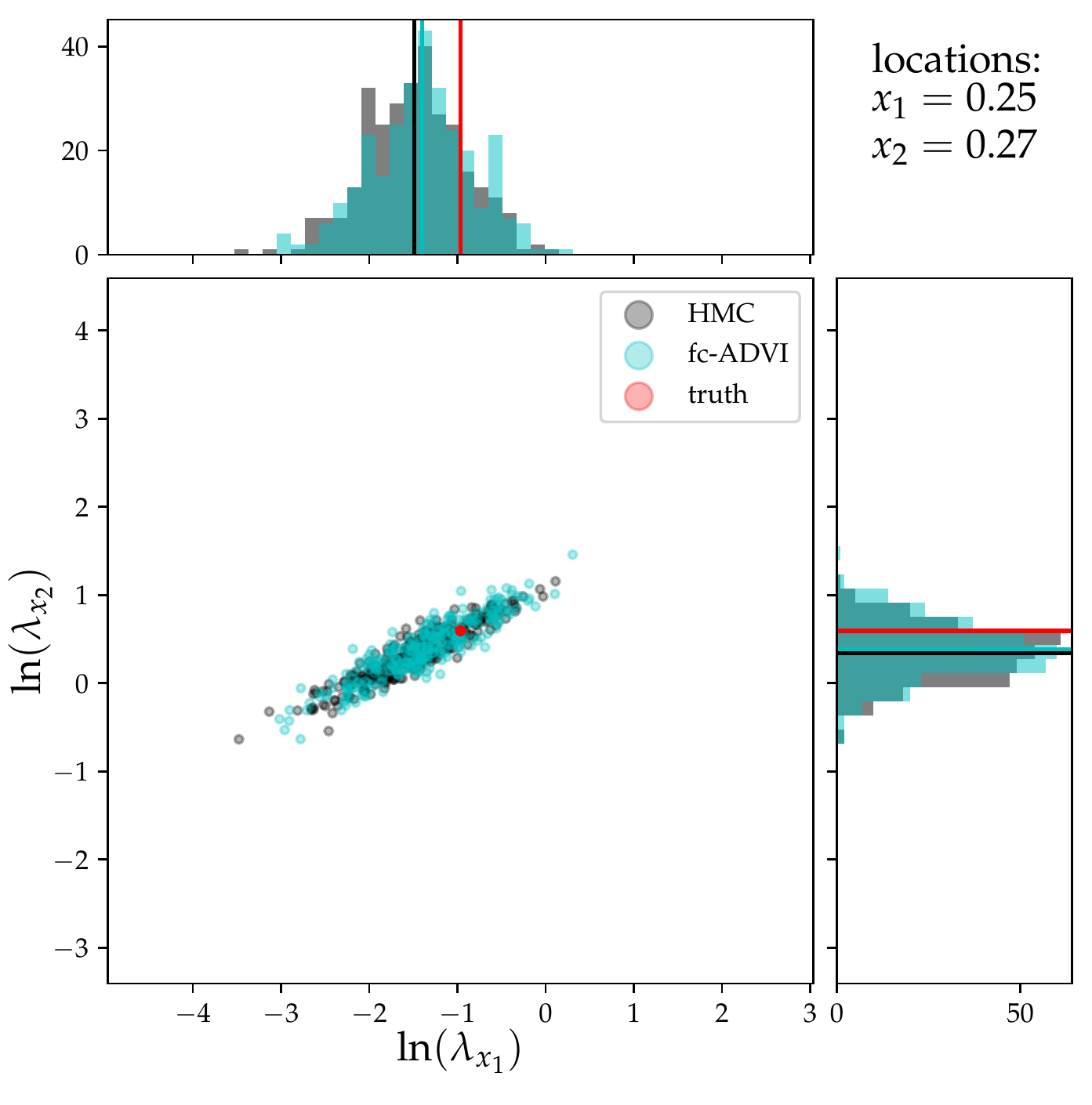}
\end{subfigure}
\begin{subfigure}[b]{0.32\textwidth}
	\includegraphics[width=\textwidth]{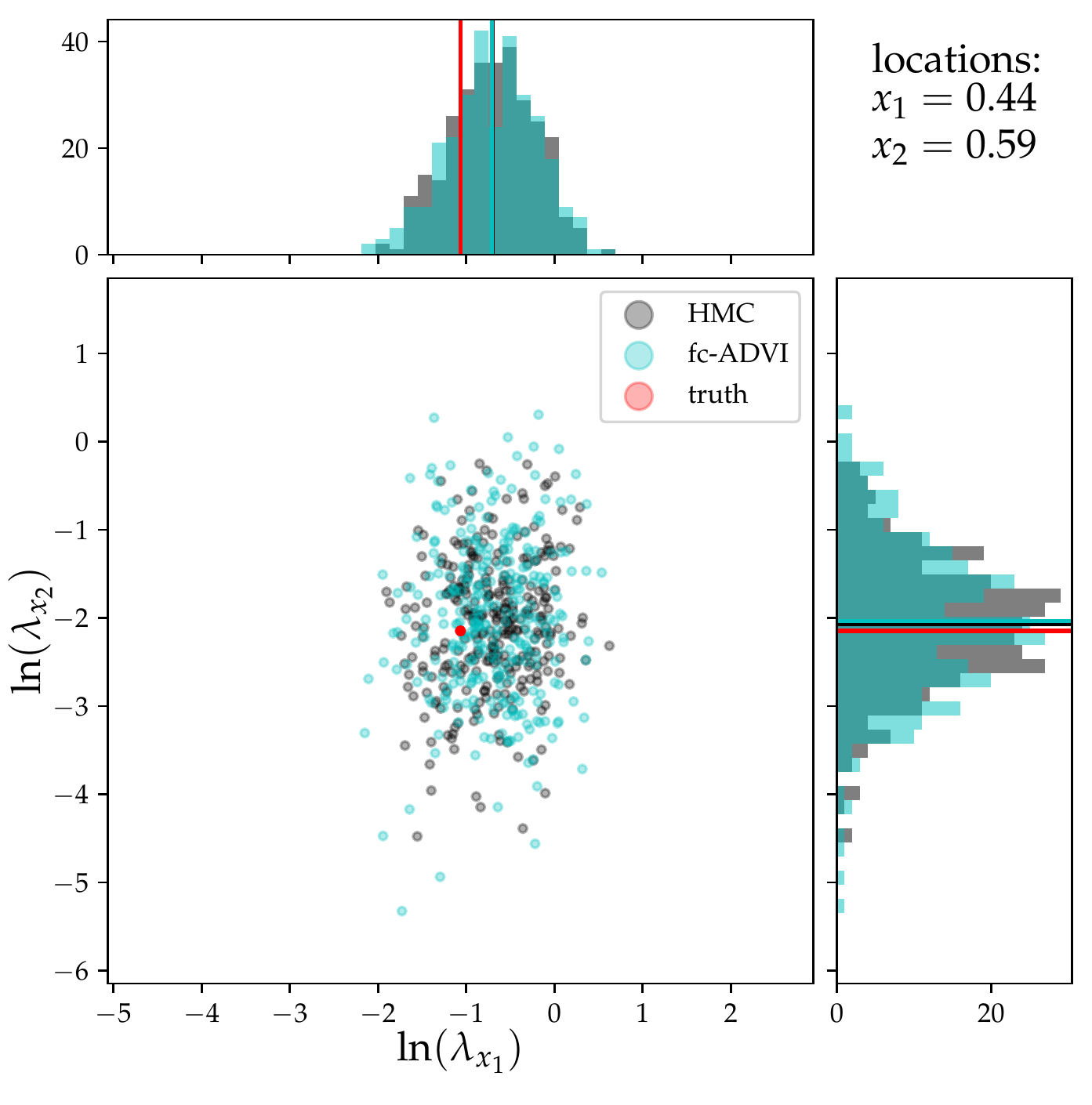}
\end{subfigure}
\begin{subfigure}[b]{0.32\textwidth}
	\includegraphics[width=\textwidth]{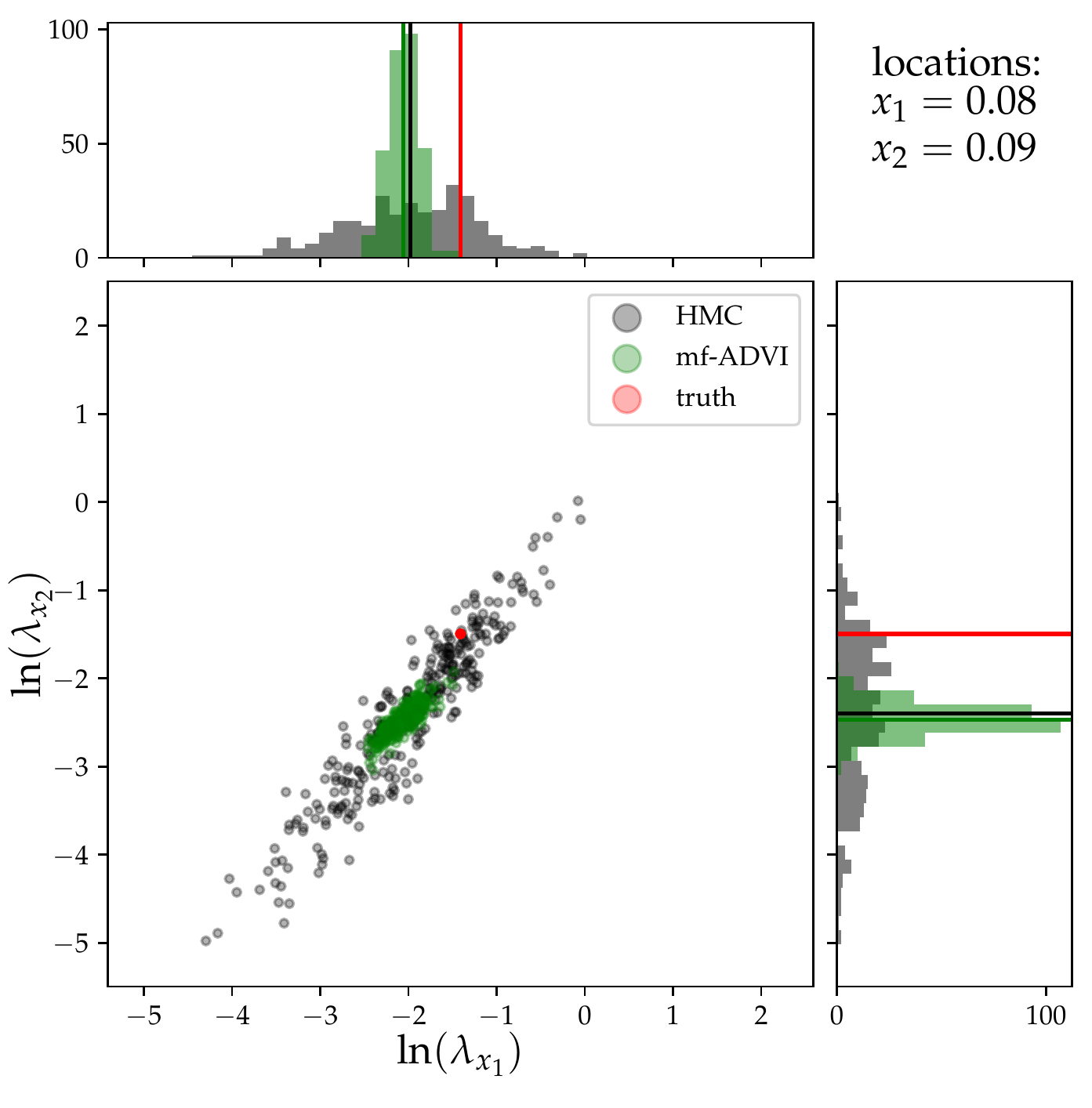}
\end{subfigure}
\begin{subfigure}[b]{0.32\textwidth}
	\includegraphics[width=\textwidth]{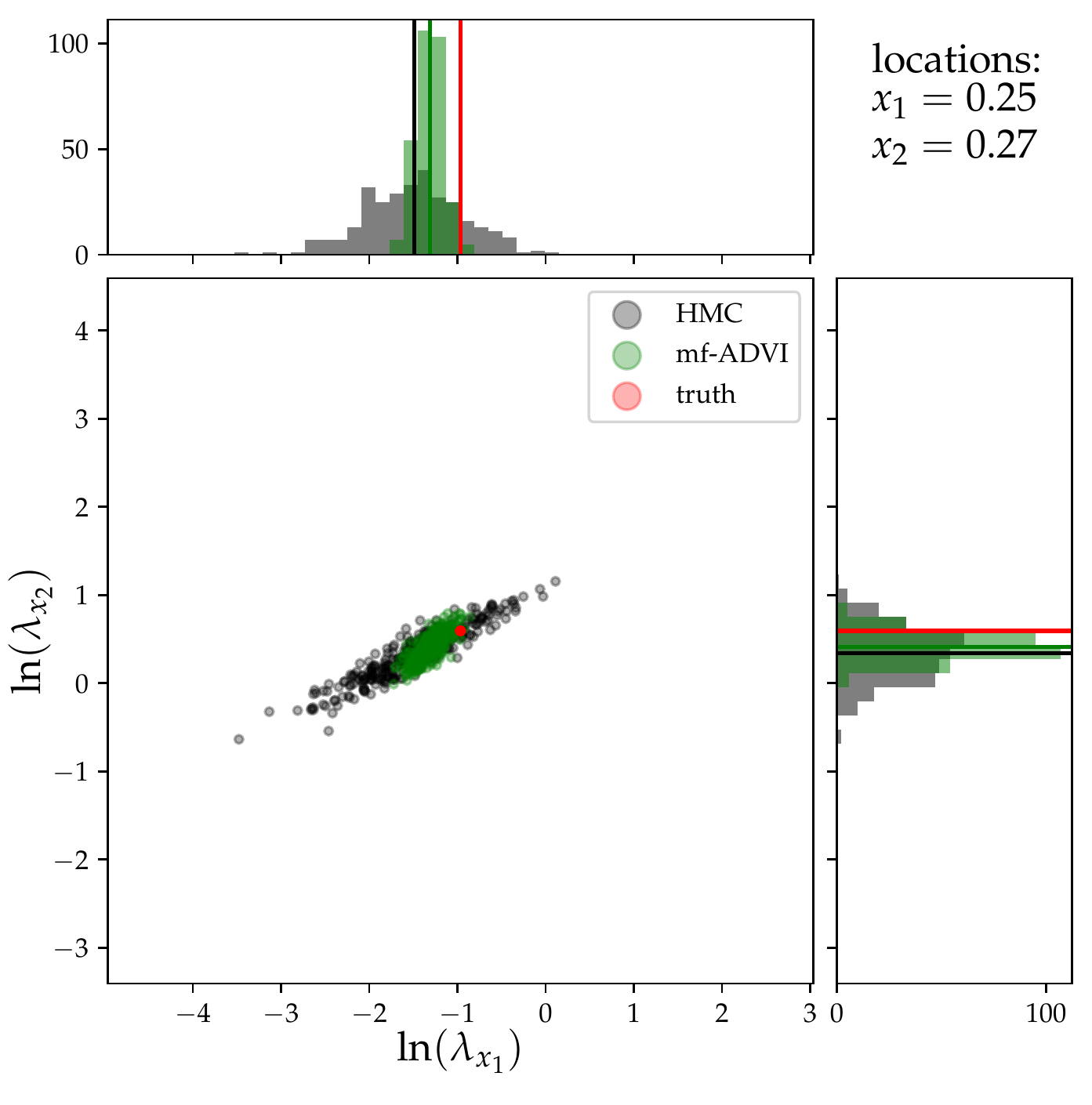}
\end{subfigure}
\begin{subfigure}[b]{0.32\textwidth}
	\includegraphics[width=\textwidth]{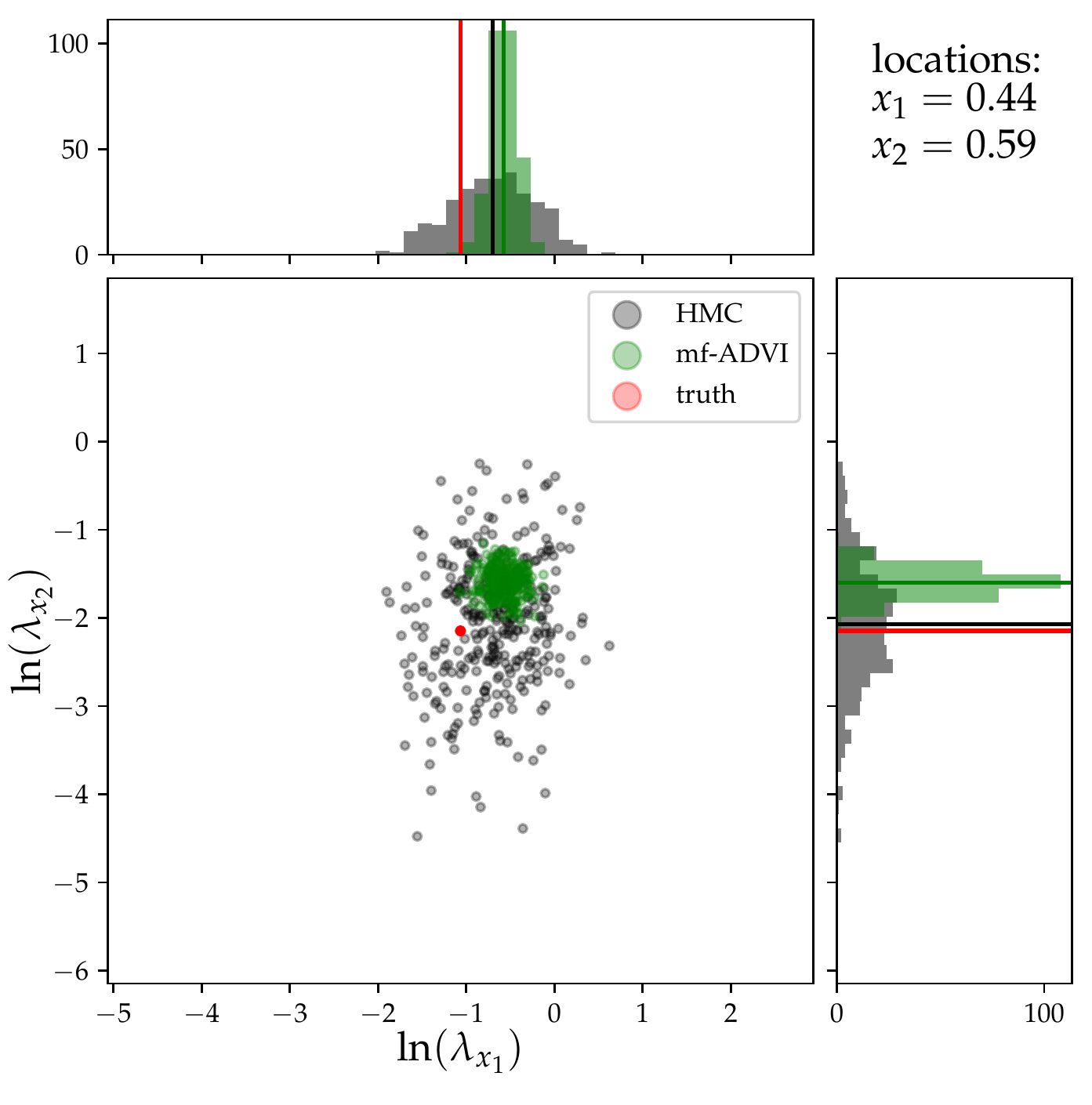}
\end{subfigure}
\begin{subfigure}[b]{0.32\textwidth}
	\includegraphics[width=\textwidth]{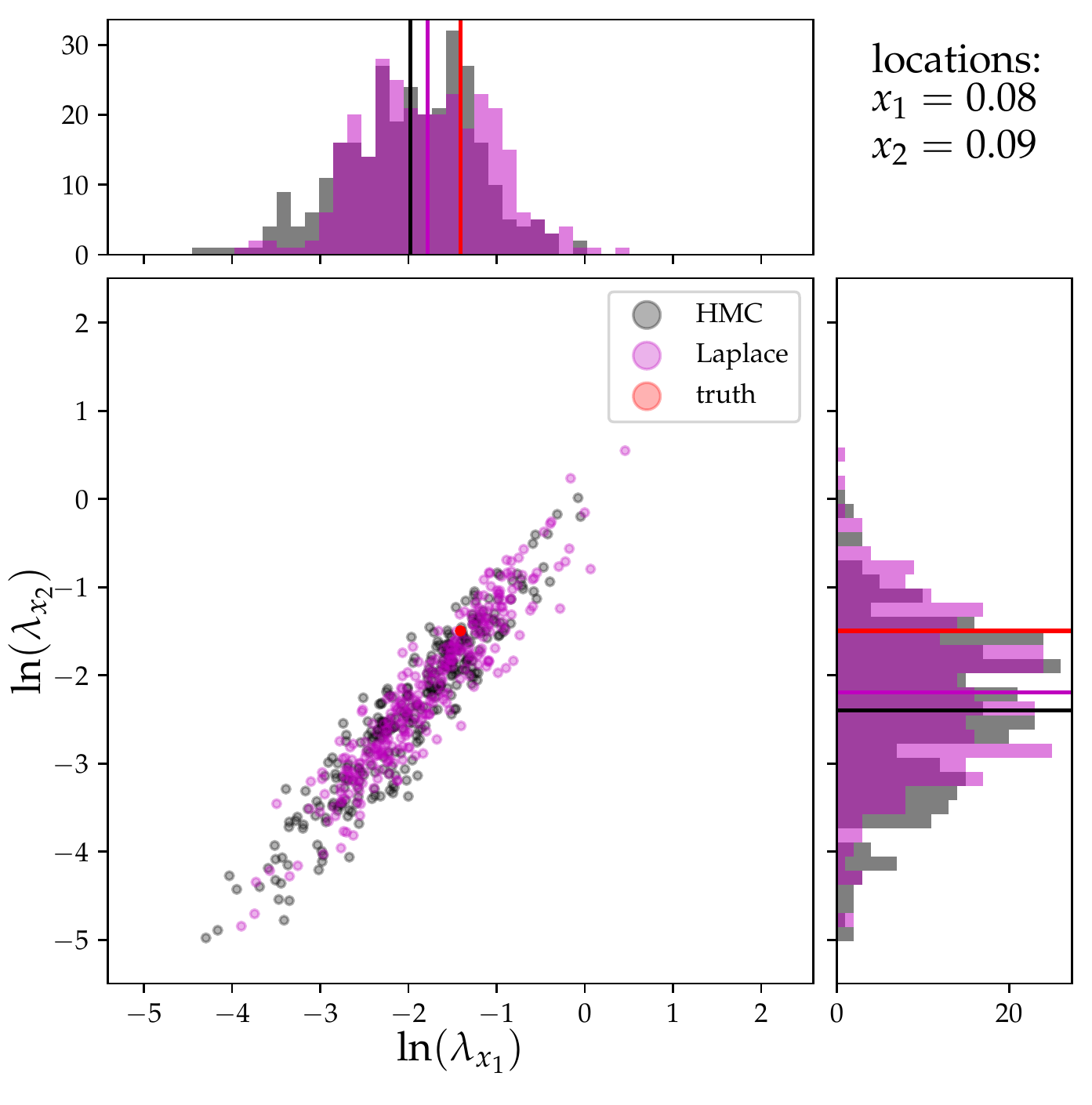}
\end{subfigure}
\begin{subfigure}[b]{0.32\textwidth}
	\includegraphics[width=\textwidth]{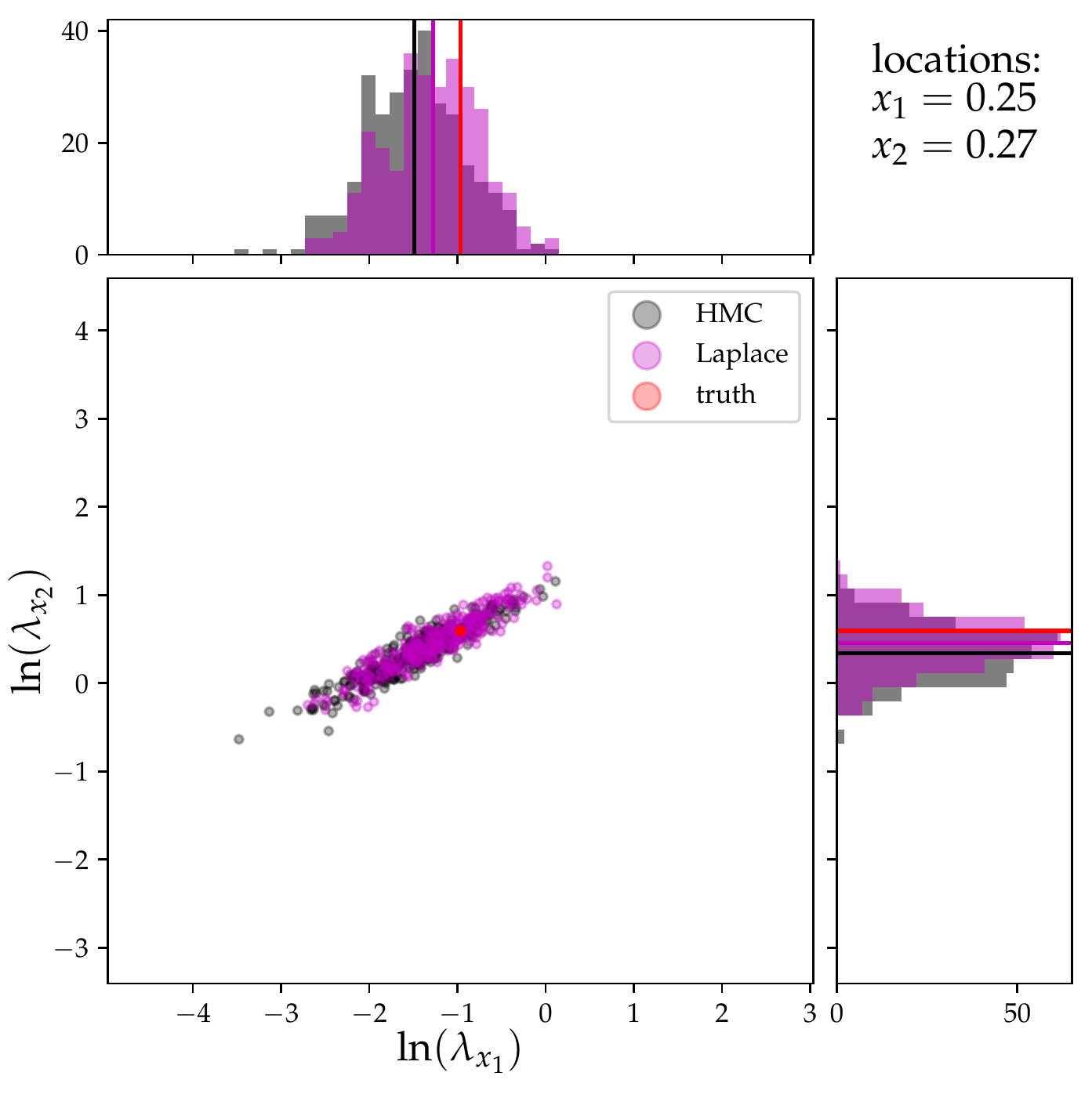}
\end{subfigure}
\begin{subfigure}[b]{0.32\textwidth}
	\includegraphics[width=\textwidth]{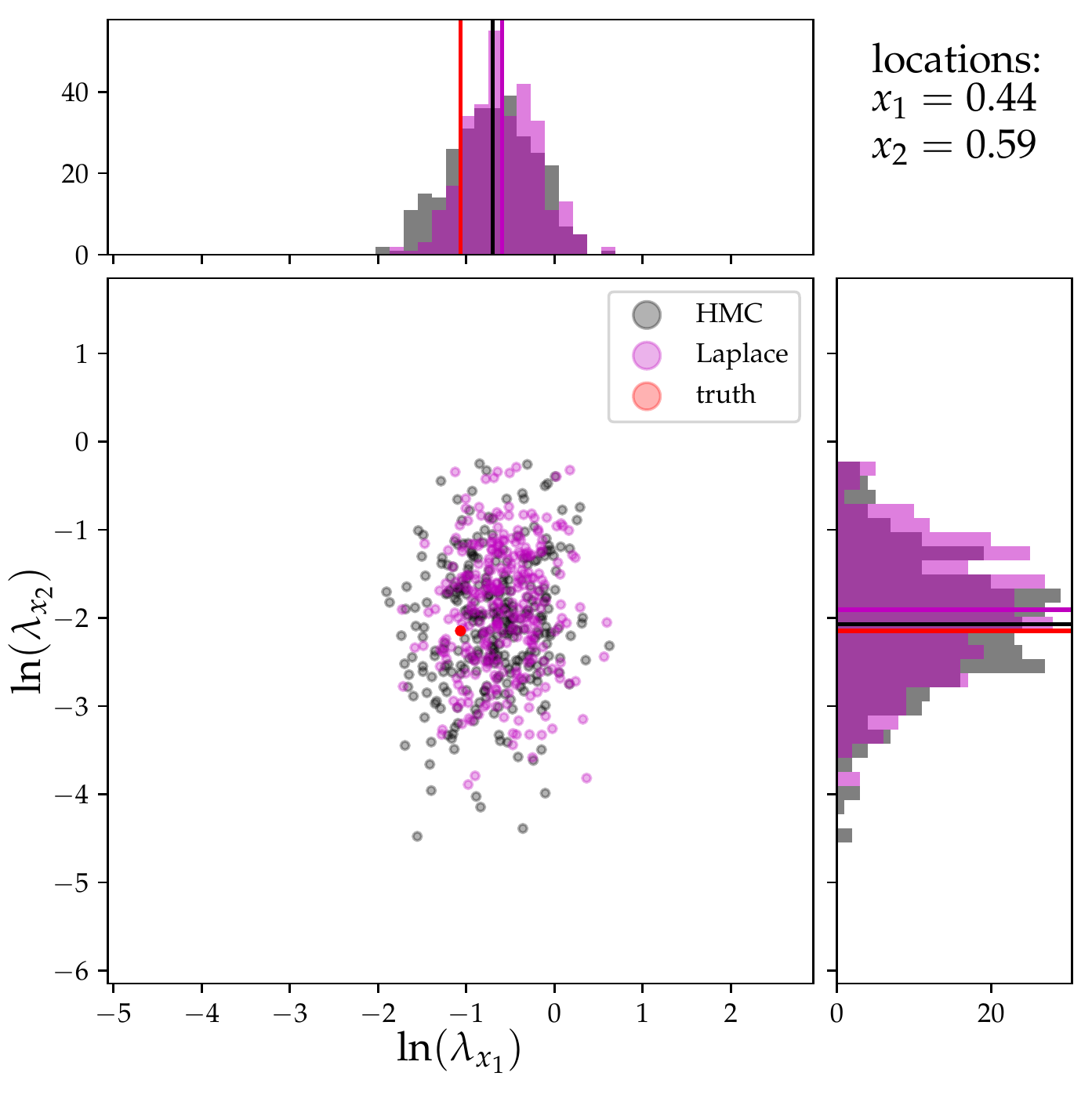}
\end{subfigure}
\caption{Scatter-plots 
of the logarithmic  posterior rates at two close-by locations in a low-count region.
The posterior samples from MGVI are compared to those of all other methods that provide posterior samples.
The true rates are indicated as well.}
	\label{fig:PLN_scatters}
\end{figure}

 \begin{figure}
	\centering
	\begin{subfigure}[b]{1\textwidth}
		\includegraphics[width=\textwidth]{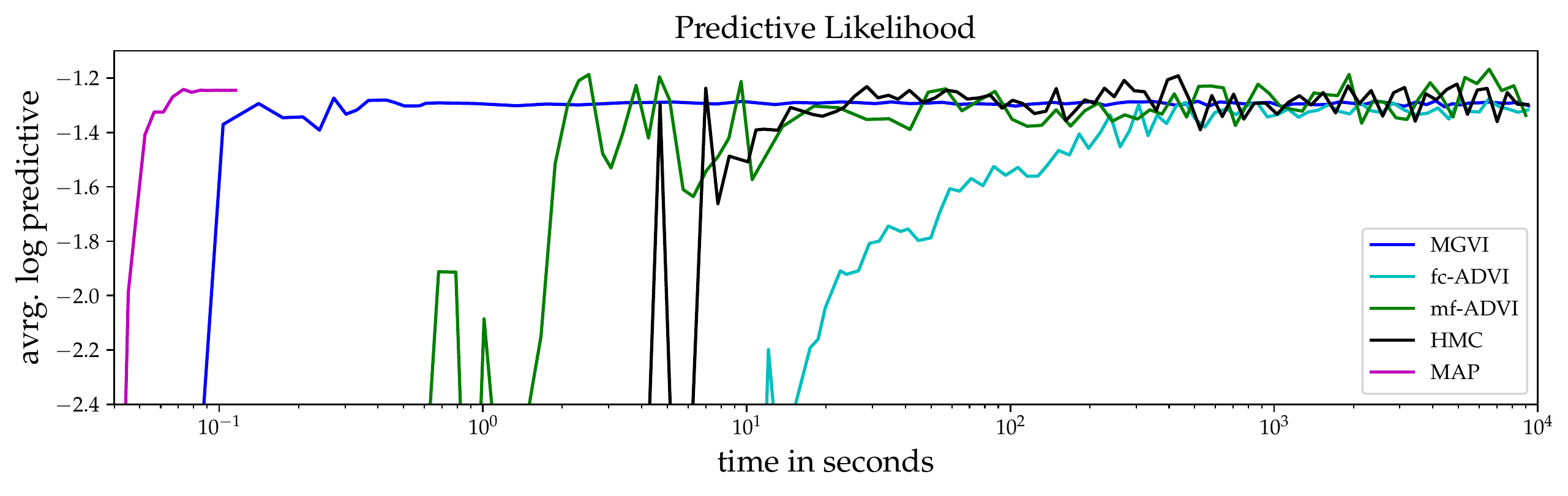}
	\end{subfigure}
	\begin{subfigure}[b]{1\textwidth}
		\includegraphics[width=\textwidth]{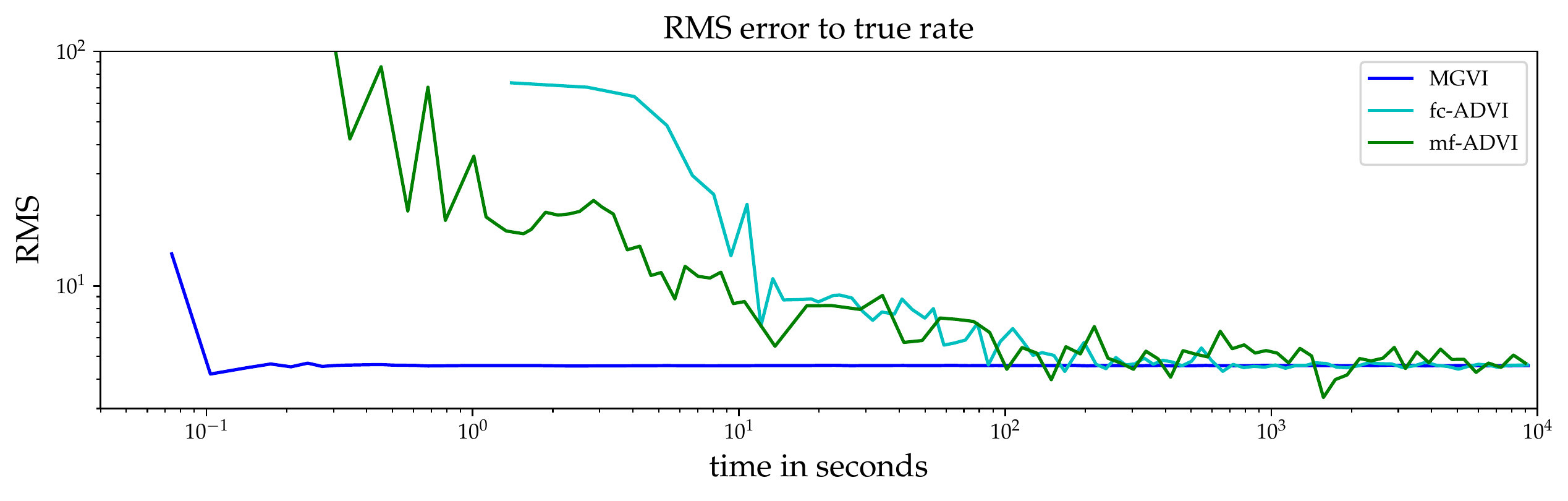}
	\end{subfigure}
	\begin{subfigure}[b]{1\textwidth}
	\includegraphics[width=\textwidth]{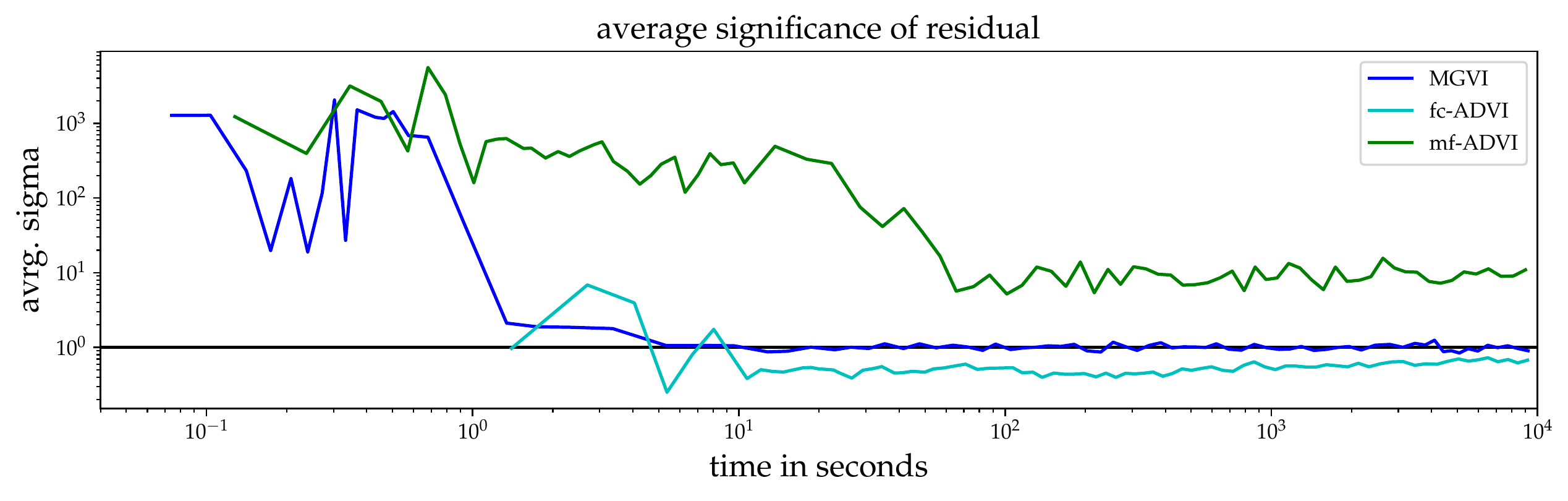}
\end{subfigure}
\caption{The performance metrics for the Poisson log-normal problem for all methods. The curves are smoothed by a moving average after the first ten points and equidistantly sampled on a logarithmic scale.}
 	\label{fig:performancemetrics}
\end{figure}

\subsubsection{Convergence behavior}
The first quantity we monitor during the optimization is the predictive likelihood on unobserved data. For this purpose we withheld $10\%$ of the data points to track how well those are explained by the current state of some method.
In this example we do have access to the underlying true rate, which allows us to monitor the RMS error to this ground truth, as well as how well the remaining residual is captured by the predicted uncertainty in terms of significance, which, for a Gaussian would be one sigma. For the definition of these quantities see Sec.~\ref{sec:performance_metrics}. All results are shown in Fig.~\ref{fig:performancemetrics}. For the predictive likelihood, by far the fastest method is a MAP estimate. Using second order natural gradient descent, this method converges within less than $0.06$ seconds. MGVI is significantly slower, but also rapidly converges in terms of the predictive likelihood. After only $0.1$ seconds and drawing new samples twice it no longer changes significantly for the remaining time. After roughly $2$ second, the next method to converge is mf-ADVI. Ten times longer is required by HMC, which takes roughly $20$ seconds to complete its burn-in. This is consistent with the one order of magnitude speedup reported in \citet{ADVI}. By far the slowest method is fc-ADVI, requiring roughly $1000$ seconds (or $16.6$ minutes) to achieve comparable predictivity. In this example MGVI is slower, but comparable to a MAP estimate, more than one order of magnitude faster than mf-ADVI, two orders of magnitude faster than HMC and four orders of magnitude than fc-ADVI, which on this problem scale is barely feasible. 

The RMS error to the true rate can be used as another indicator how fast the methods converges. In the case of MGVI, not much happens after the first global iteration, requiring $0.1$ seconds. The RMS error of both ADVI methods steadily drop down to the final level, mf-ADVI being initially faster, but fc-ADVI catches up before final convergence.

Interesting is the behavior of the average significance, characterizing how well the deviations from the true rate are explained by a Gaussian approximation using the samples provided by the methods. It also allow us to evaluate how well the covariance of each method has converged. For MGVI this seems to be the case after $10$ seconds. This coincides with the increase of samples used to estimate the KL-divergence. With more samples the uncertainties are probed better and the average significance of the residuals are spot on the one sigma level, hinting at a quite Gaussian posterior. Although mf-ADVI converges quickly in terms of the predictive likelihood, here we observe drifts within the first $100$ seconds. This is even more extreme in the case of fc-ADVI, which only drifts gradually and probably did not fully converge, still slightly overestimating the variance. This is consistent with the shift relative to the HMC standard deviations.

Overall, MGVI is fast because it has intrinsically fewer parameters and (quasi-) second order optimization can be used. Also the observation that means of Gaussian approximations converge fast and a covariances slowly might also contribute to the rapid convergence behavior of MGVI. Only the mean has to be optimized for a given covariance, and once it converged a new, plausible covariance is adapted, without having to laboriously optimize for it.

\subsection{Binary Gaussian Process Classification with non-parametric Kernel}
In the second example we apply MGVI to a much higher dimensional problem and more complex context, making it unfeasible for a fully parametrized covariance. Binary Gaussian process classification is used to attribute regions to certain classes and identify boundaries between them. A comprehensive overview can be found in \citet{kuss2005assessing} and \citet{nickisch2008approximations}. In addition to the typical formulation, we also infer the underlying kernel non-parametrically. With this extension a Laplace approximation will not provide reasonable results as parameters are degenerate. In this example we compare MGVI to mf-ADVI, which is still capable of coping with such extremely high dimensional problems. We consider binary data in two spatial dimensions, measured only at certain locations. The likelihood is a Bernoulli distribution and its rate parameter is described by a sigmoid function applied to an underlying Gaussian process. The kernel of this process is unknown and will be modeled non-parametrically as well. We assume a stationary, isotropic kernel and model it by two spectral components. The first component follows a power law that is modified by the second component, a  log-Gaussian process  with a smooth kernel. Overall the spectral density is parametrized by two power-law parameters, an amplitude and the spectral index, and the Gaussian process parameters for the component modifying this power-law. This model is inspired by systems with underlying processes favoring certain length-scales and is well-suited for imaging applications.
\subsubsection{Setup}

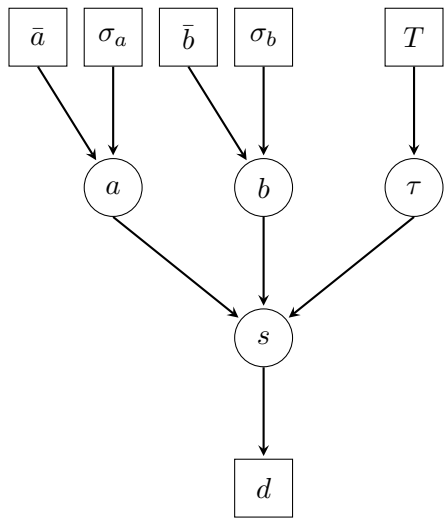
\begin{figure}[h]
	\centering
	\begin{tikzpicture}
	[c/.style={circle,minimum size=2em,text centered,thin},
	r/.style={rectangle,minimum size=2em,text centered,thin},
	v/.style={->,shorten >=1pt,>=stealth,thick}, 
	arrow/.style={-latex, shorten >=1ex, shorten <=1ex, bend angle=45}]
	
	\node(abar)at(-2,6)[r,draw]{$\sigma_a$};
	\node(asig)at(-3,6)[r,draw]{$\bar{a}$};
	
	\node(bbar)at(0,6)[r,draw]{$\sigma_b$};
	\node(bsig)at(-1,6)[r,draw]{$\bar{b}$};
	
	\node(T)at(2,6)[r,draw]{$T$};
	
	\node(a)at(-2,4)[c,draw]{$a$};
	\node(b)at(0,4)[c,draw]{$b$};
	\node(tau)at(2,4)[c,draw]{$\tau$};
	\node(s)at(0,2)[c,draw]{$s$};
	\node(d)at(0,0)[r, draw]{$d$};
	\draw[v](T.south)--(tau);
	\draw[v](asig.south)--(a);
	\draw[v](abar.south)--(a);
	\draw[v](bsig.south)--(b);
	\draw[v](bbar.south)--(b);
	\draw[v](a.south)--(s);
	\draw[v](b.south)--(s);
	\draw[v](tau.south)--(s);
	\draw[v](s.south)--(d);
	\end{tikzpicture}
	\flushleft
	\caption{The graphical structure of the binary Gaussian process classification with non-parametric kernel.}
	\label{fig:diagram} 
\end{figure}

The likelihood in this example is the Bernoulli distribution that reads
\begin{align}
\label{eq:Bernoulli}
\mathcal{P}(d\vert \mu) &= \prod_i \mathcal{P}(d_i \vert \mu_i)  \text{\quad , with}\\
\mathcal{P}(d_i\vert \mu_i) &= \mu^d_i (1-\mu_i)^{1-d_i}\text{\quad .}
\end{align}
for some rate parameter on the unit interval $\mu \in(0,1)$ and binary outcome $d \in \{0,1\} $.
The Fisher information metric for this likelihood is
\begin{align}
\label{eq:BernoulliMetric}
I_d(\mu) = \widetilde{\mu (1-\mu)}^{-1} \text{.}
\end{align}
 The rate $\mu$ is linked to a Gaussian process $s \sim \mathcal{G}(s\vert 0,S)$ by a sigmoid function and a linear response:
\begin{align}
\mu &= R\sigma(s)\\
&= R \:\frac{1}{2}(1+\mathrm{tanh}\left(s\right))\text{\quad .}
\end{align}
The kernel $S$ of this process is assumed to be stationary and isotropic and can be expressed as $S = \mathbb{F}^{-1} \widetilde{\mathbb{P}{p}}\, \mathbb{F} $ with spectral density $p$. This quantity itself is to be learned and it is modeled according to 
\begin{align}
p(k) = e^{a\: \mathrm{ln}k+b +\tau_k} \text{\quad . }
\end{align}
The first two terms in the exponent model a power-law kernel, which is linear on double-logarithmic scale, with power $a$ and amplitude $b$, both of which get a Gaussian prior with assumed mean ($\bar{a}$ and $\bar{b}$) and variance ($\sigma_{a}$ and $\sigma_{b}$). The last term in the exponent follows an integrated Wiener process on logarithmic spatial scale, ergo a differentiable function,  according to the known kernel $T = AA^\dagger$. The integrated Wiener process follows a power-law kernel with power four. We treat it analogously to the other correlation kernel $S$. This prior is standardized by performing a Fourier transformation on logarithmic coordinates and multiplication with the square root of the power-law spectrum. The graphical structure of this described model is shown in Fig.~\ref{fig:diagram}. 

Reparametrizing the model parameters yields the following relation to the original rate $\mu$:
\begin{align}
\mu &= f(\xi) \\
&= R\sigma \left( \mathbb{F}^{-1} \left(\widetilde{\mathbb{P} e^{\left(\bar{a}+ \sigma_a \xi_a\right)\: \mathrm{ln}k +\bar{b}+ \sigma_b \xi_b+A\:\xi_\tau}}\right)\xi_s\right)\text{\quad .}
\end{align} 
This reparametrized model has therefore a highly non-linear likelihood in terms of its parameters, where $\xi = \left(\xi_s,\xi_a,\xi_b,\xi_{\tau}\right)^\dagger$. This expression has to be read from the right to the left, which is the direction of the generative model. A series of linear and point-wise non-linear operations are performed on the latent model parameters to generate $\mu$. $\mathbb{F}$ and $\mathbb{P}$ are again the  Fourier transformation and the isotropic projection of a $1\mathrm{D}$ spectrum to a $2\mathrm{D}$ Fourier space. Here again, the tilde indicate that the quantity below is raised to a  diagonal operator. Obtaining this function is tedious but straightforward and can be done automatically, given the hierarchical structure of the model. We spare the reader the expressions of the Jacobian of the function with respect to its parameters $J(\xi) = \frac{\partial f(\xi)}{\partial \xi}$ as this should be implemented using auto-differentiation, which NIFTy5 \citep{Nifty5} provides to us. Structurally, the problem is now identical to the previous one with information and approximate covariance:

\begin{align}
\mathcal{H}(d,\xi) \:&\widehat{=}\: -d^\dagger \mathrm{ln} f(\xi) - (1-d)^\dagger \mathrm{ln}\left(1-f(\xi)\right) + \frac{1}{2} \xi^\dagger \mathbb{1} \xi \\ 
\Xi(\widehat{\xi}) &= \left( J(\widehat{\xi})^\dagger  \widetilde{\left(f(\widehat{\xi}) (1-f(\widehat{\xi})\right)}^{-1} J(\widehat{\xi}) + \mathbb{1} \right)^{-1}\text{\quad .}
\end{align}
Regarding the numerical setup, we consider $2^{19}$ binary data points on a two dimensional plane organized in a checkerboard. We use $1024 \times 1024$ parameters to describe the Gaussian process underlying this rate. The spectral density is parametrized by additional two parameters for the power-law and $64$ for the non-parametric part $\xi_\tau$, resulting in overall more than a million parameters, which is completely out of reach for explicit covariance parametrization. For simplicity periodic boundaries were assumed. Due to the large parameter dimension, we initially choose to use $100$ conjugate gradient iterations, and increase it towards $400$ at the end. Otherwise we use the setup from the previous example.

\subsubsection{Results}
\begin{figure}
	\centering
	\begin{subfigure}[b]{0.49\textwidth}
		\includegraphics[width=\textwidth]{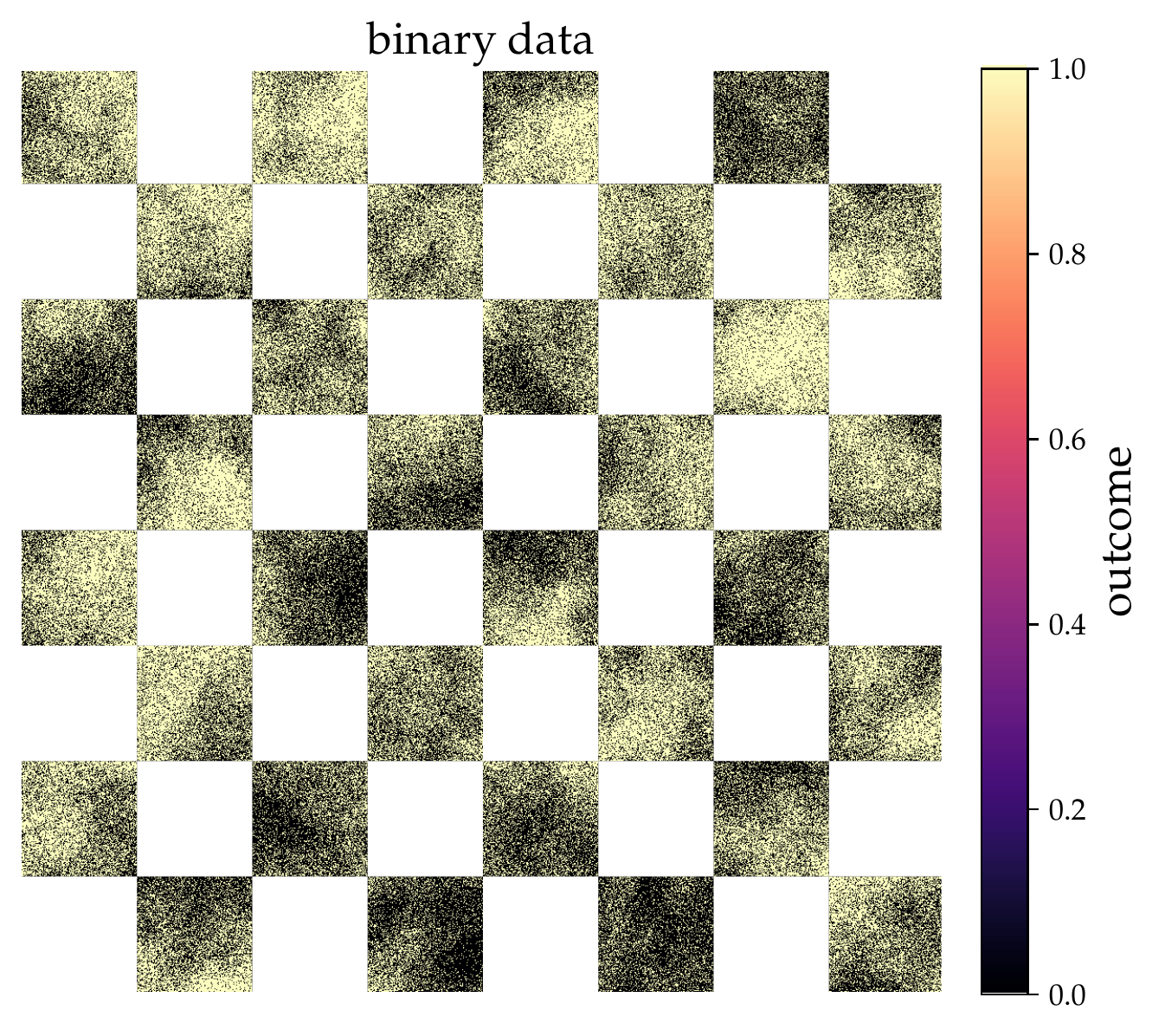}
	\end{subfigure}
	\begin{subfigure}[b]{0.49\textwidth}
		\includegraphics[width=\textwidth]{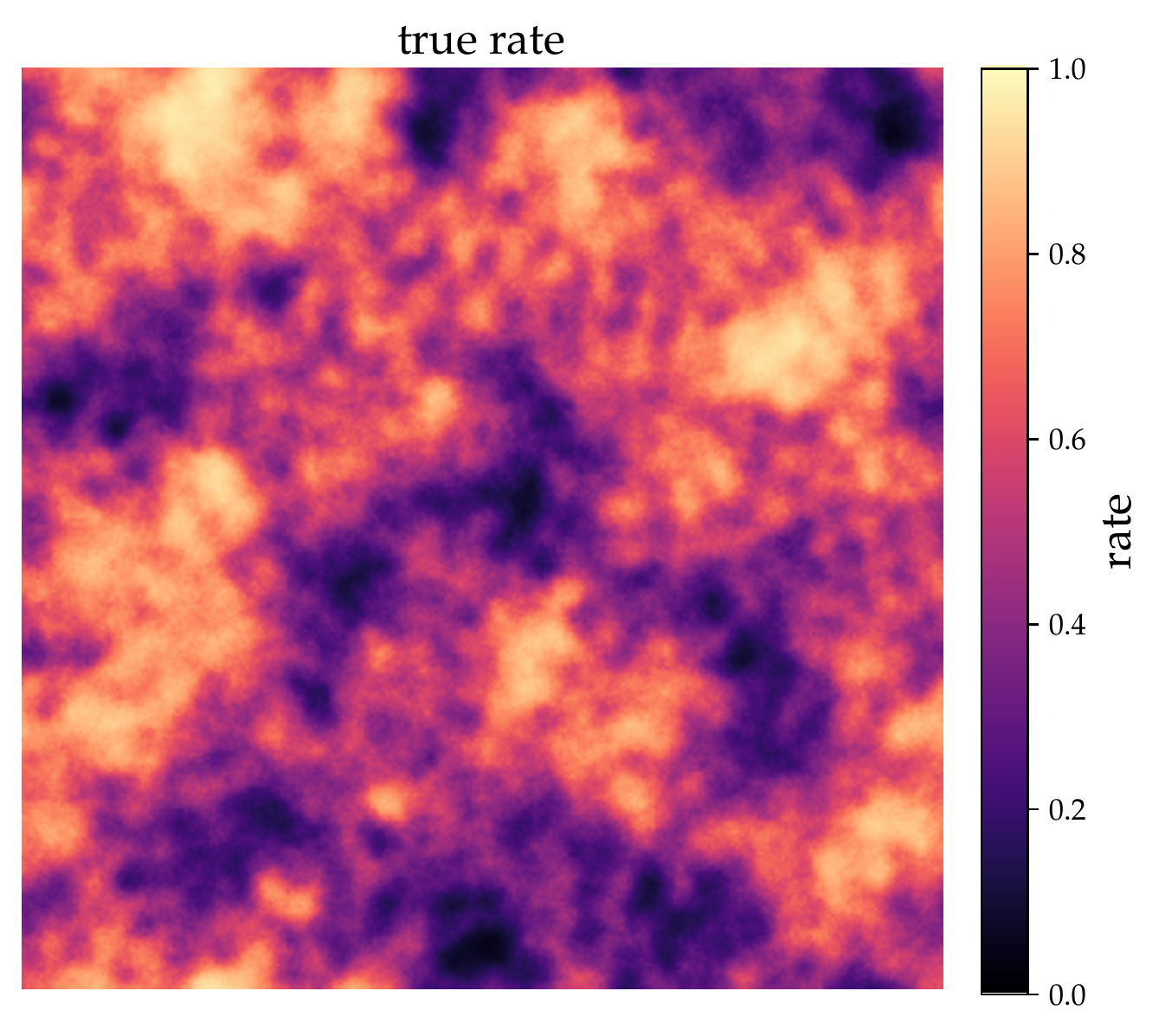}
	\end{subfigure}
	\begin{subfigure}[b]{0.49\textwidth}
		\includegraphics[width=\textwidth]{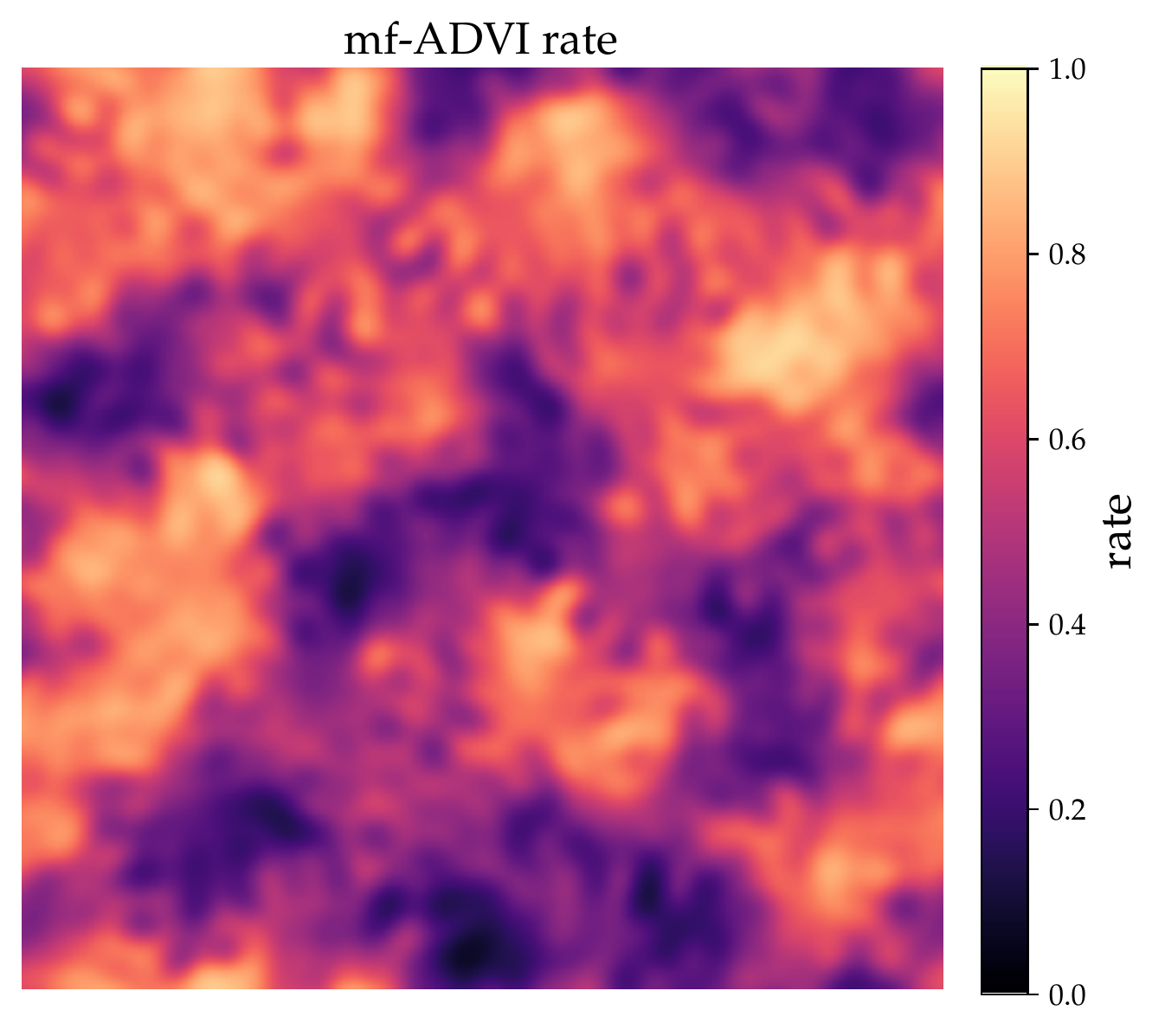}
	\end{subfigure}
	\begin{subfigure}[b]{0.49\textwidth}
	\includegraphics[width=\textwidth]{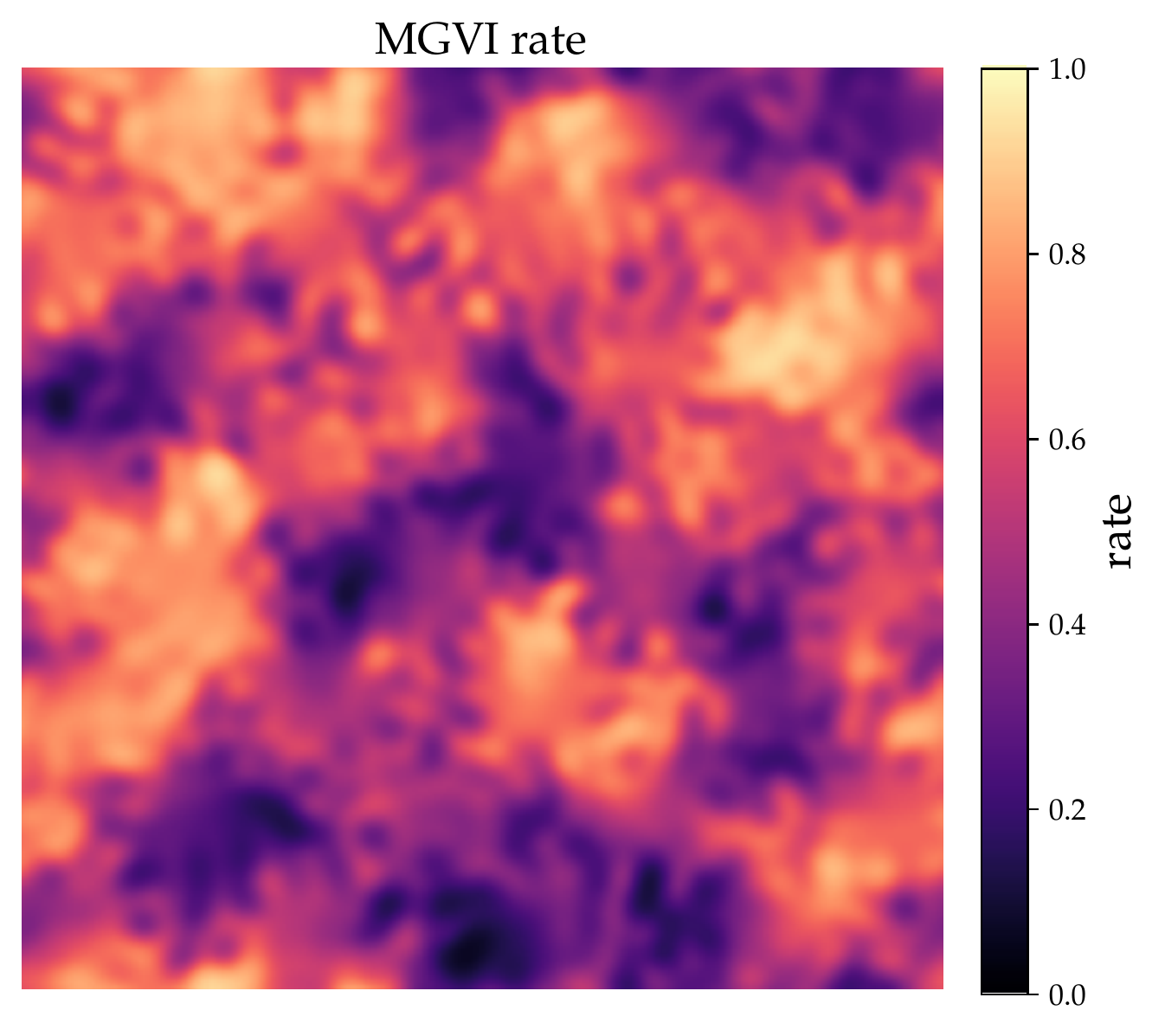}
\end{subfigure}
			\begin{subfigure}[b]{0.49\textwidth}
		\includegraphics[width=\textwidth]{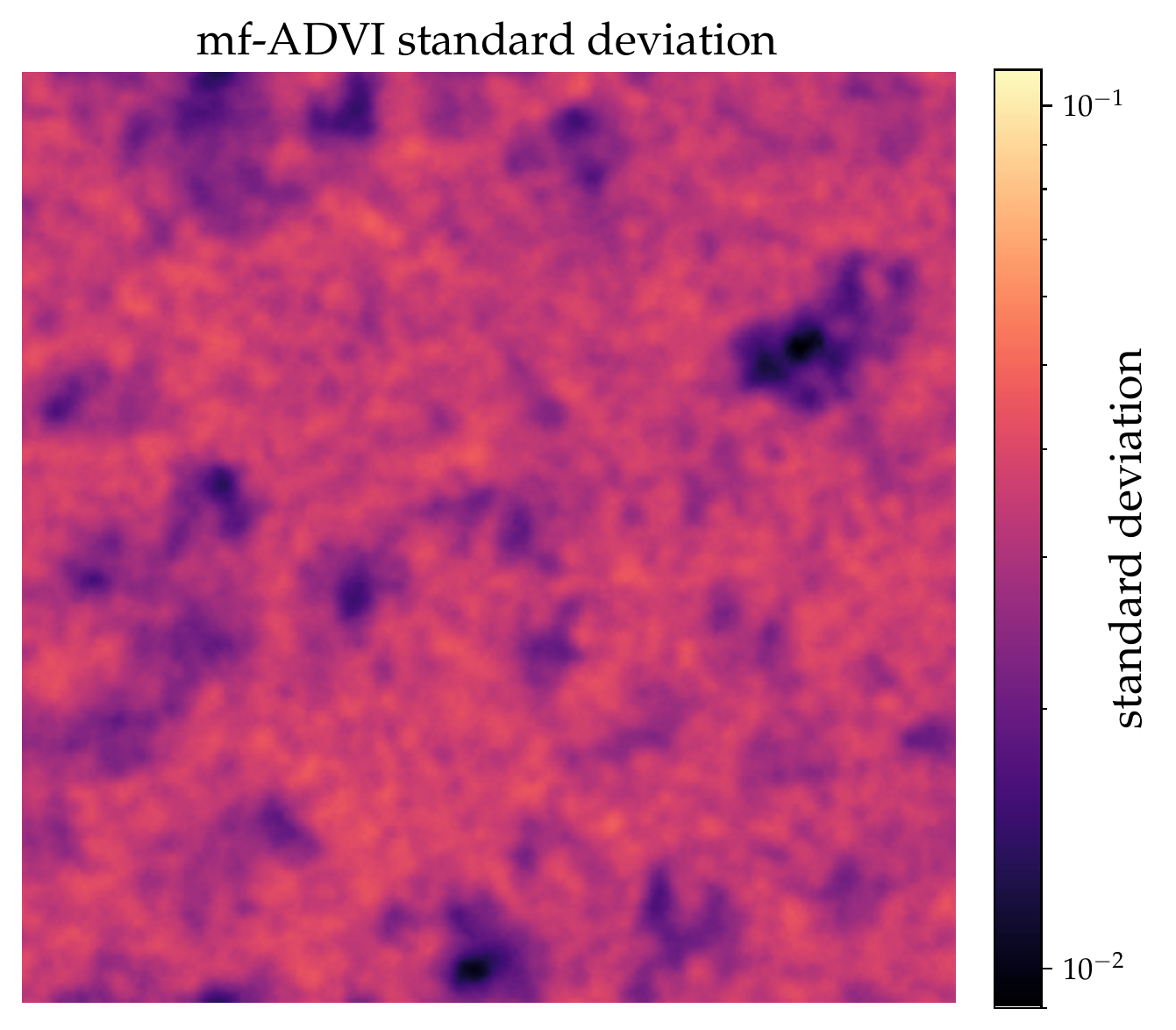}
	\end{subfigure}
		\begin{subfigure}[b]{0.49\textwidth}
		\includegraphics[width=\textwidth]{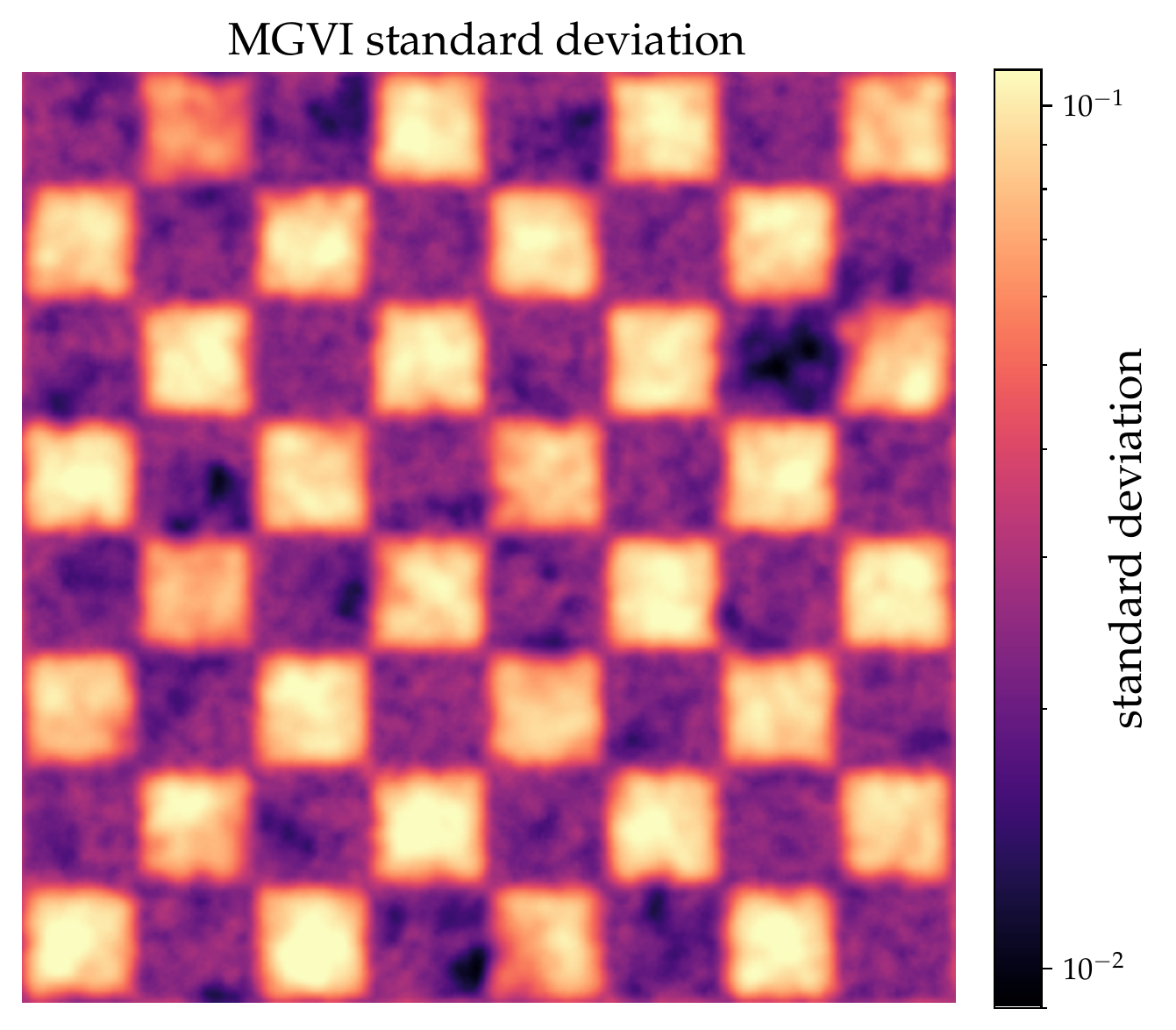}
	\end{subfigure}
	\caption{The data and true rate, as well as the mf-ADVI and MGVI means and standard deviations. Note that the small-scale noise in the data can lead to a color blend that does not seem to be part of the used color scheme.}
\label{fig:large_example}
\end{figure}

The synthetic data, the true underlying rate, the mf-ADVI, as well as MGVI results are shown in Fig.~\ref{fig:large_example}. The data is only sampled at certain locations and due to the binary output appears noisy. It exhibits spatial characteristics, predominately showing one class over the other in certain regions. The true rate, from which the data was drawn, shows rich features on all scales. The largest of them can also be seen in the data directly, but small-scale features are washed out due to the Bernoulli noise. The MAP solution to this problem (not shown) does not provide a plausible posterior estimate and completely over-fits the data.

The mean rate recovered by MGVI matches up to a certain scale exceptionally well to the true rate. Even in unobserved areas the structures are recovered correctly to some extent (as can be seen e.g. in the top right and bottom left corners). Small scales cannot be recovered as the data does not contain much information on them. This is also reflected in the standard deviation at each location. The highest uncertainty is, as expected, in the not observed areas, reproducing the checkerboard pattern. The standard deviation is also modulated by the rate itself. The more a certain region is attributed to one class, the lower its uncertainty. The uncertainty is especially high at the boundaries between the classes.

Also mf-ADVI recovers the underlying rate quite well, with maybe slightly less sharp features, but certainly comparable to the MGVI result. The main difference lies in the uncertainty estimate, which completely lacks the spatial features attributed to the incomplete checkerboard sampling of the data. Nevertheless, it shows the error associated with the nonlinear error propagation from the Gaussian process to the rate. This is similar to the behavior observed in the previous example. Compared to the standard deviation from MGVI, the uncertainty seems to be larger in areas with observed data and significantly smaller in unobserved regions. 

\begin{figure}
	\centering
		\begin{subfigure}[b]{0.49\textwidth}
		\includegraphics[width=\textwidth]{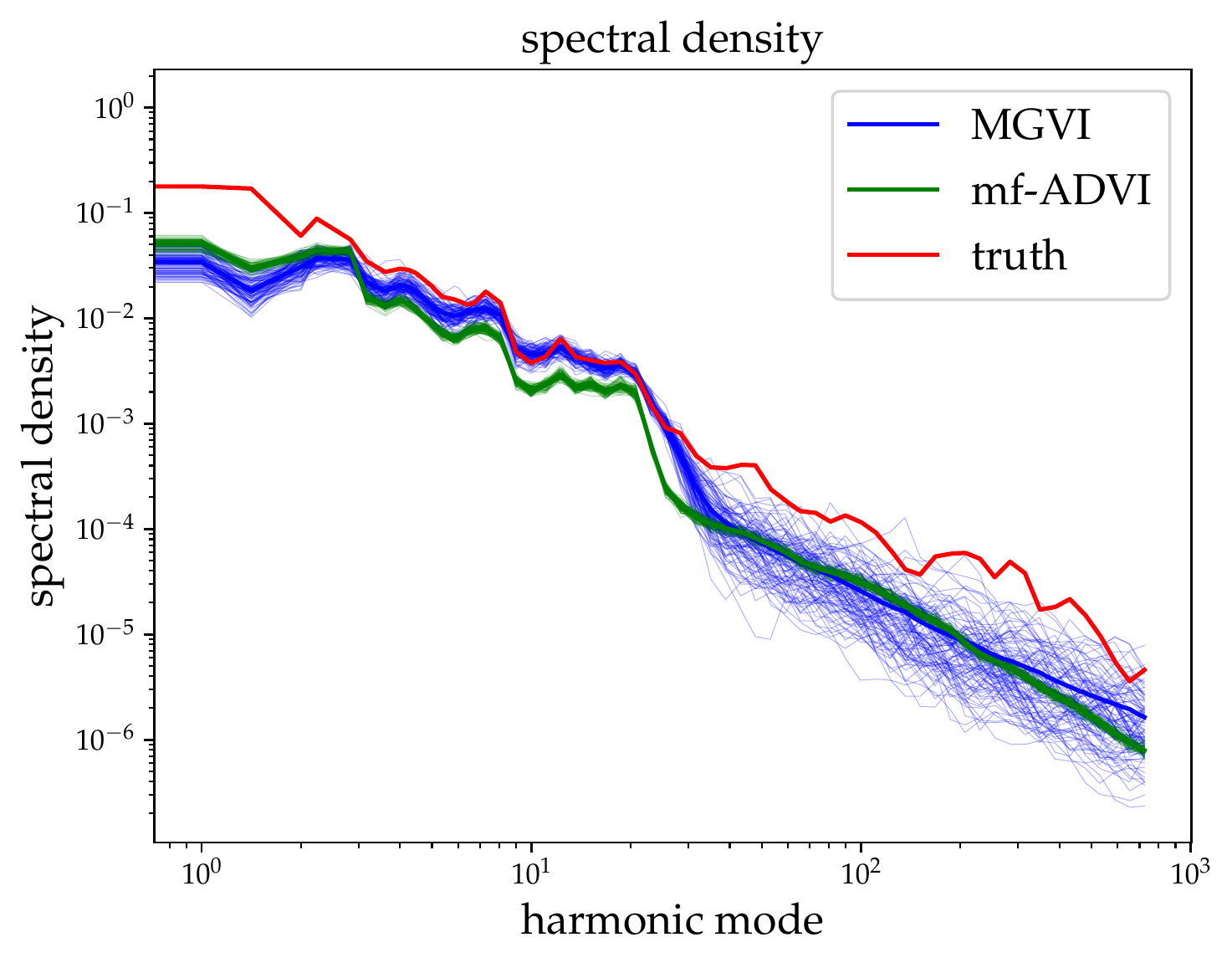}
	\end{subfigure}
		\begin{subfigure}[b]{0.49\textwidth}
	\includegraphics[width=\textwidth]{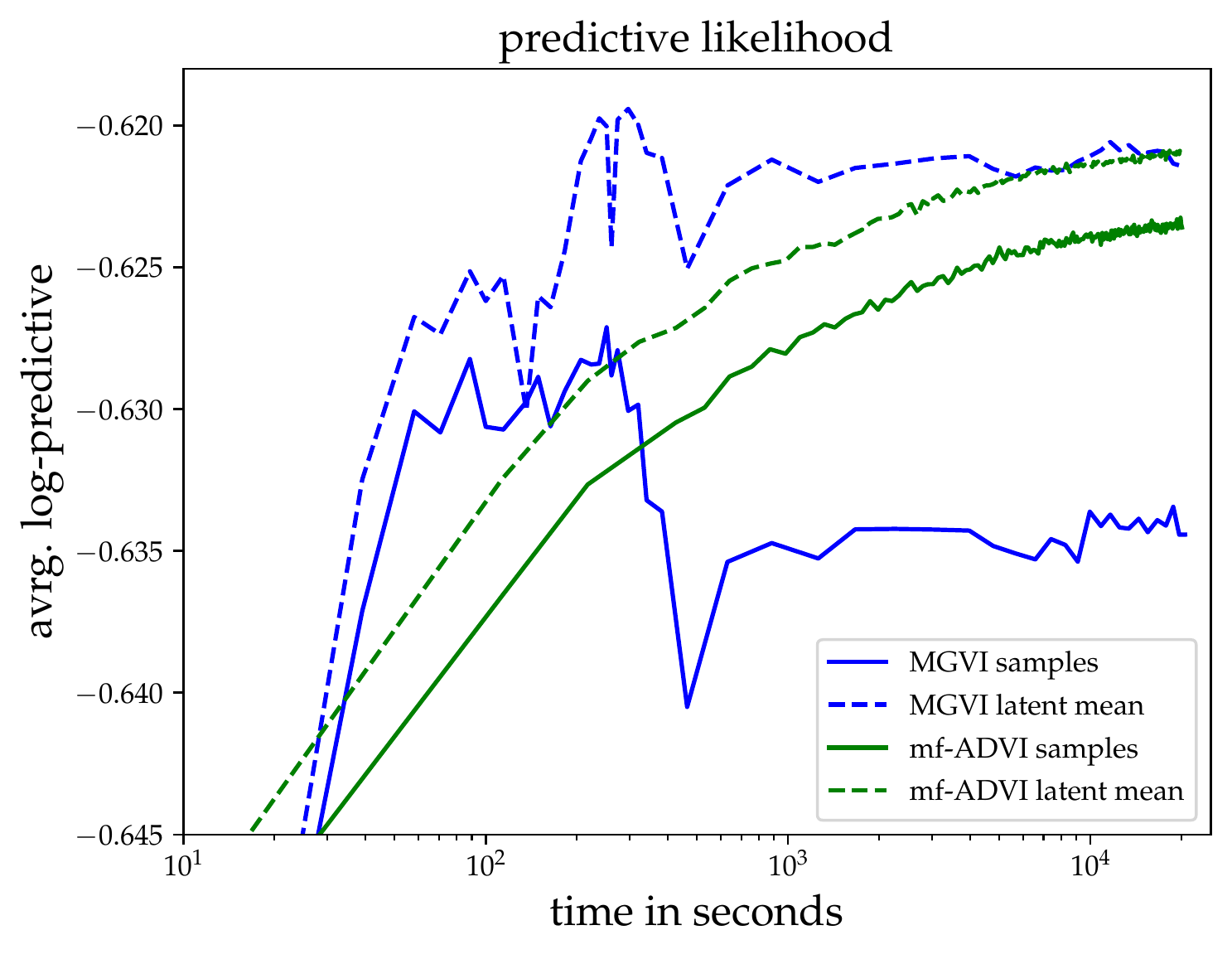}
	\end{subfigure}
			\begin{subfigure}[b]{0.49\textwidth}
		\includegraphics[width=\textwidth]{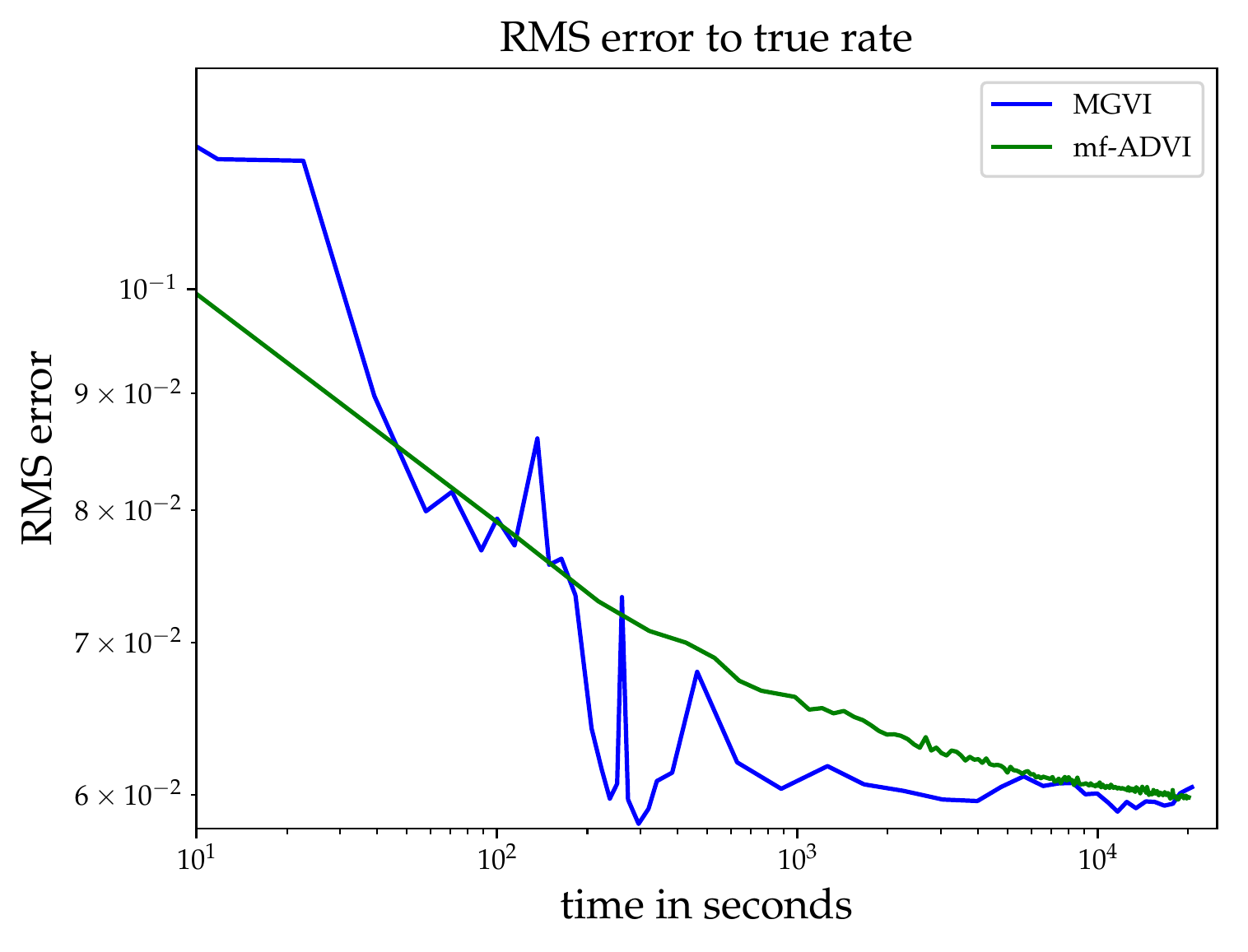}
	\end{subfigure}
	\begin{subfigure}[b]{0.49\textwidth}
		\includegraphics[width=\textwidth]{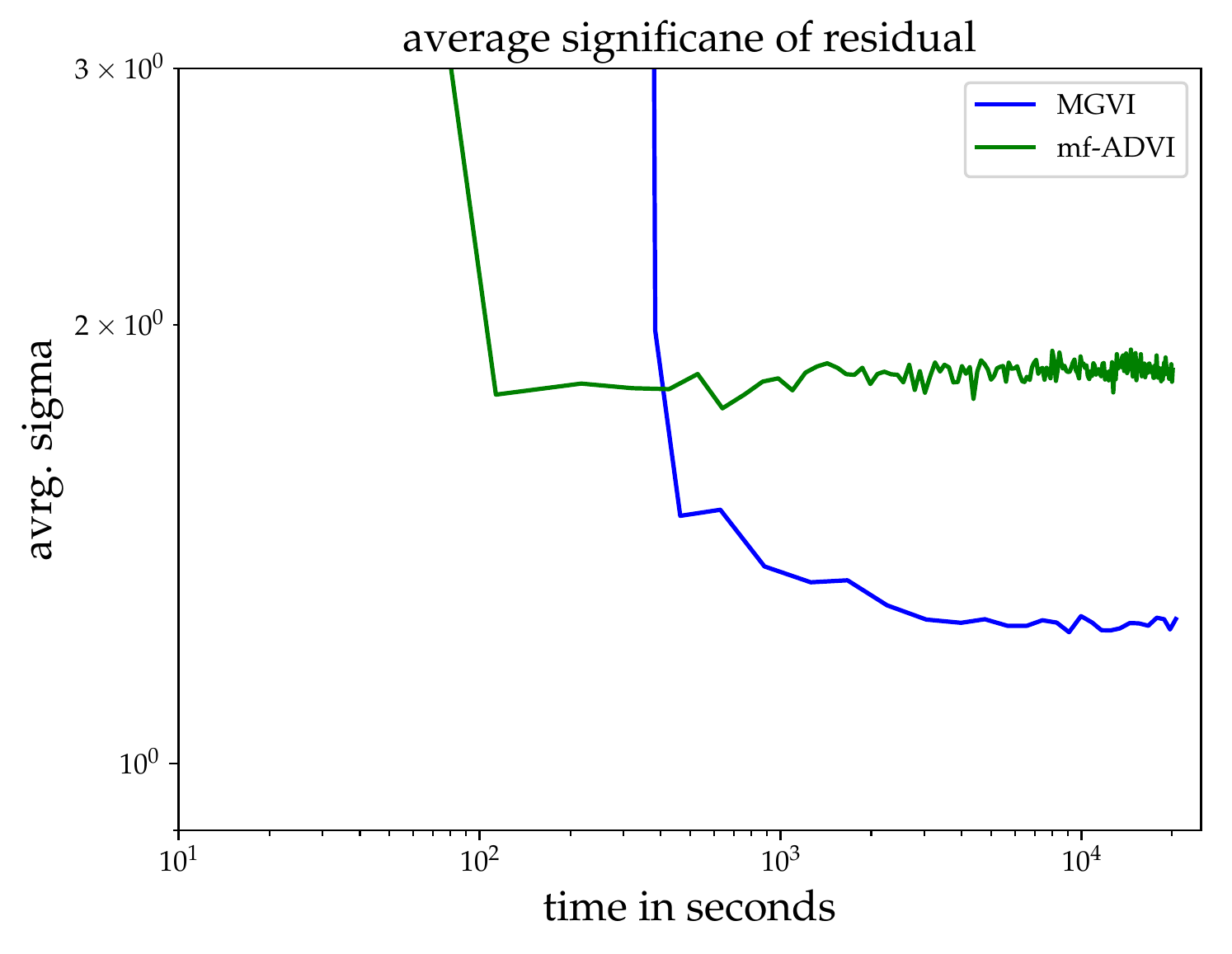}
\end{subfigure}
\caption{Recovered spectral density (top left), and the performance metrics for the binary Gaussian process classification problem, comparing MGVI to mf-ADVI.}
\label{fig:power_and_metrics}
\end{figure}

The recovered spectral density  is shown in Fig.~\ref{fig:power_and_metrics}. At the largest scales, and therefore the smallest modes, the true correlation structure is correctly recovered within the error by MGVI, indicated by a set of samples. Even most large-scale spectral features  are identified correctly by the algorithm. At a some point towards smaller scales the uncertainty increases significantly. This is also the point where the recovered spectrum diverges from the true one. This might indicate incomplete convergence, however, those highly uncertain parameters are the last ones to converge anyway and are affected the most by the stochastic estimation of the KL-divergence. Even on those scales the trues spectrum is not completely out of the error bound and seems consistent with the recovered spectrum. The mean spectrum obtained by mf-ADVI is similar to MGVI, but is slightly shifted down for the most part, but the error estimates are again not spatially modulated, not reflecting the deviations from the truth. 

Regarding the result, MGVI seems to be better at describing the true posterior distribution with slightly more accurate means but superior uncertainties. Also important is how fast MGVI achieved this result, compared to mf-ADVI. For this we track again, as in the previous example, the predictive likelihood of all the unobserved data, filling in the checkerboard. Additionally we track the RMS error of the mean to the ground truth, as well as the average significance of the residual. 

The convergence behavior of MGVI and mf-ADVI are also shown in  Fig.~\ref{fig:power_and_metrics} as well. Here MGVI shows the first accurate results after roughly $200$ seconds in terms of RMS error and predictive likelihood. Shortly after this, the predictive likelihood of the mean and samples significantly diverge. This coincides with the increase from a single pair of antithetic samples to a gradually larger number, shown by the steep drop in the average significance. We suspect that the system found a self-consistent solution with that single sample and it got strongly disturbed by the presence of more samples. Finally the system recovers and converges to a similar predictivity and consistent error. In this example, mf-ADVI is significantly slower to achieve comparable predictivity and RMS error by roughly one order of magnitude, but those levels are achieved.

\subsubsection{Discussion of meta parameters}

\begin{figure}
	\centering
	\begin{subfigure}[b]{0.32\textwidth}
		\includegraphics[width=\textwidth]{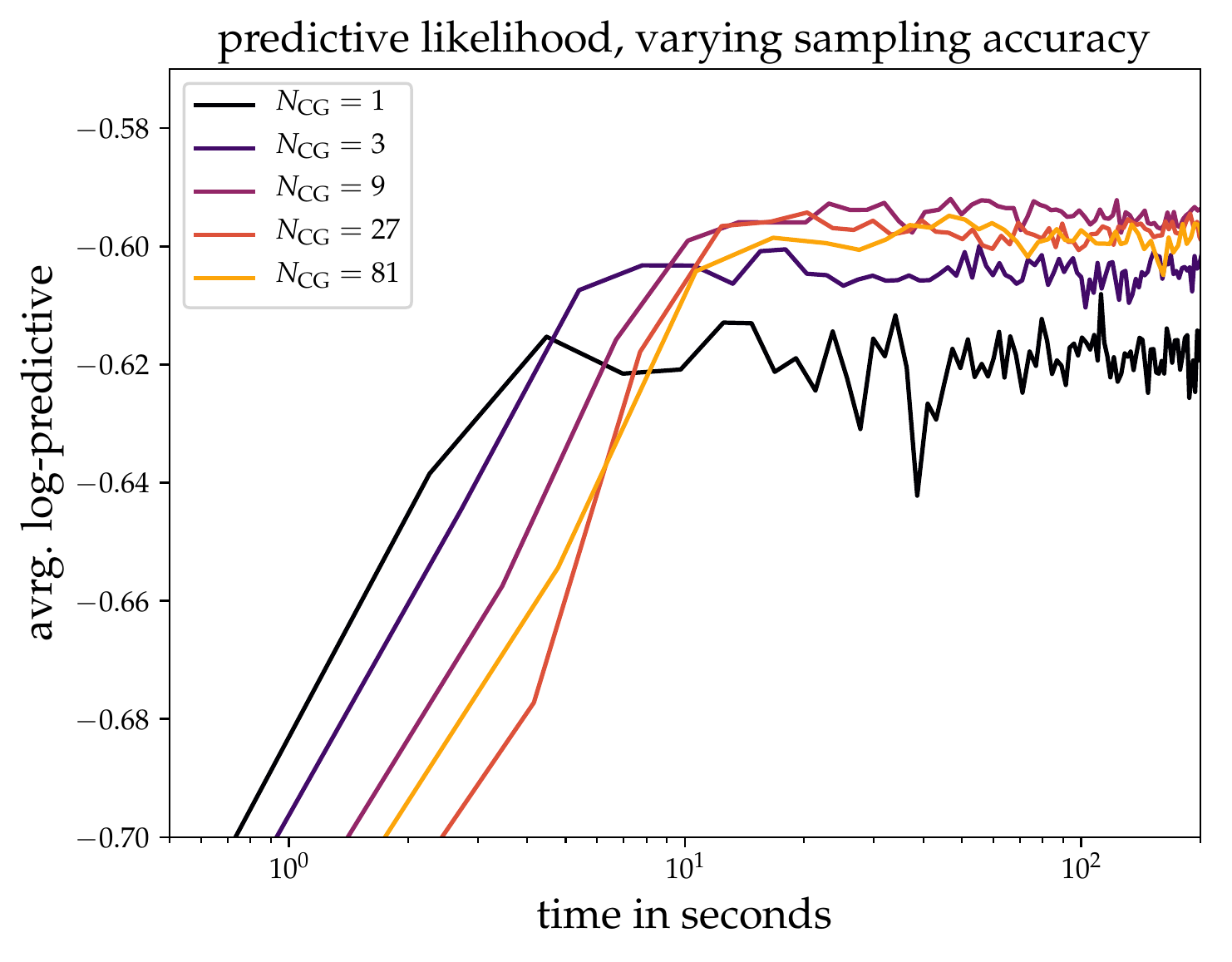}
	\end{subfigure}
	\begin{subfigure}[b]{0.32\textwidth}
	\includegraphics[width=\textwidth]{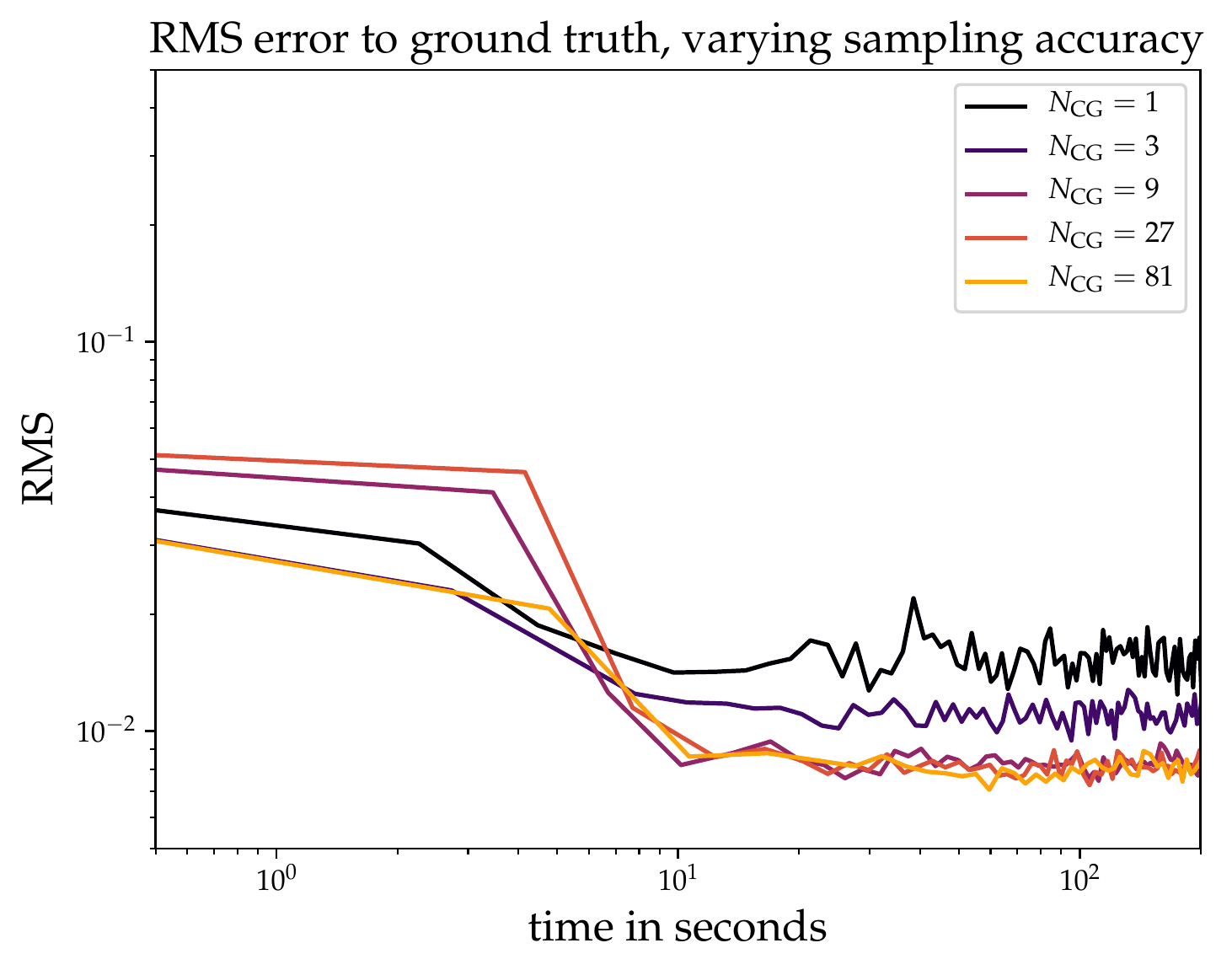}
\end{subfigure}
	\begin{subfigure}[b]{0.32\textwidth}
	\includegraphics[width=\textwidth]{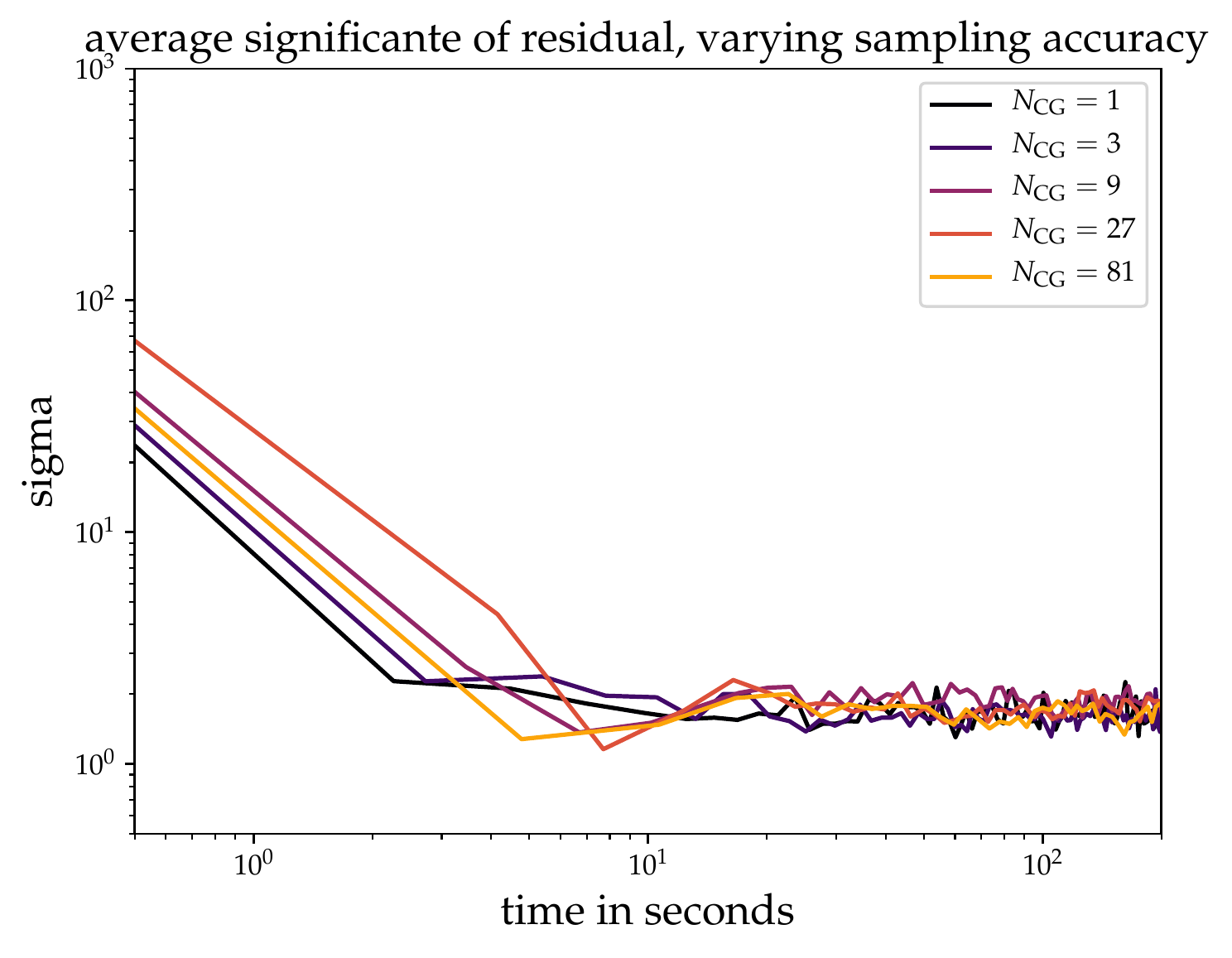}
\end{subfigure}
	\begin{subfigure}[b]{0.32\textwidth}
	\includegraphics[width=\textwidth]{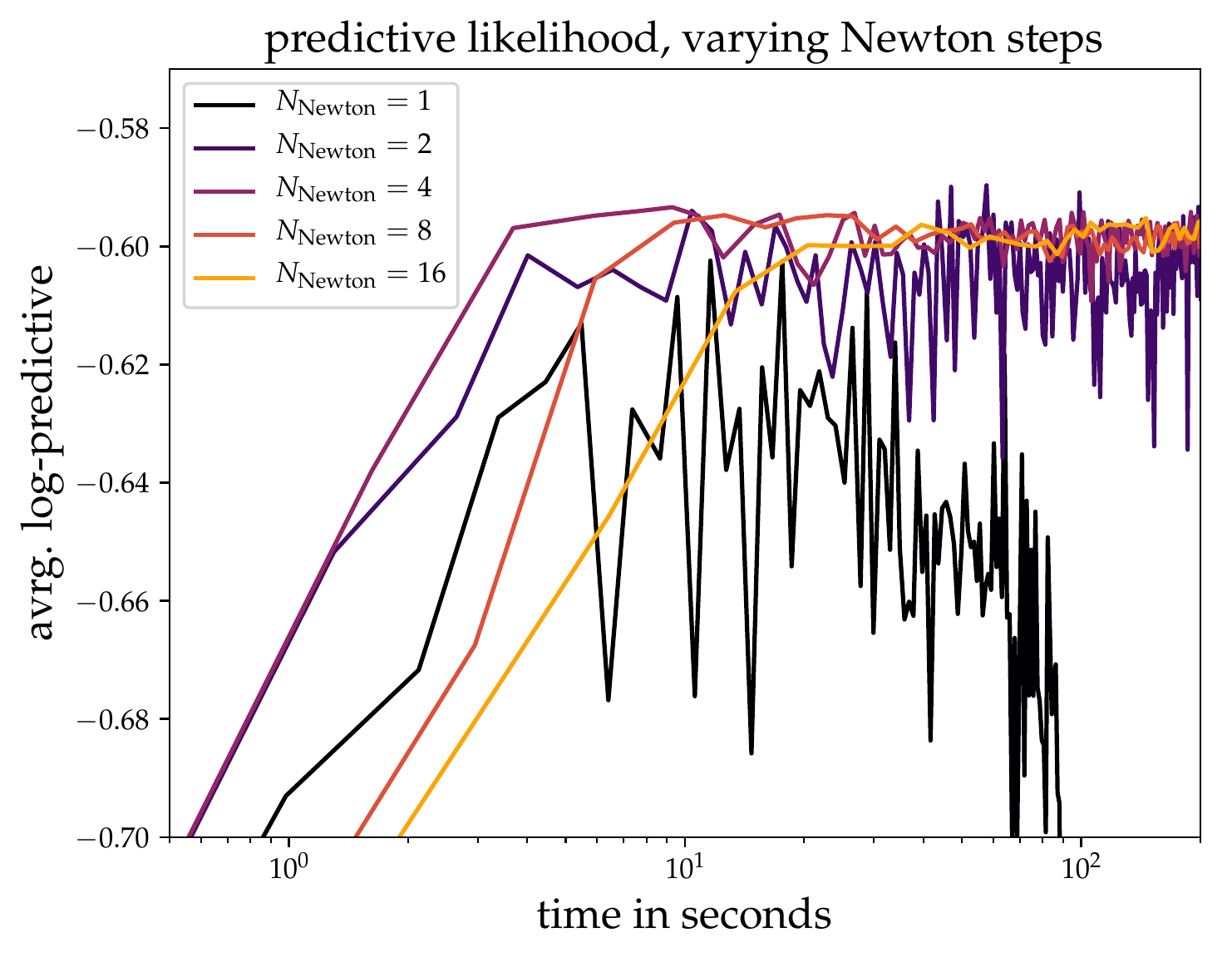}
\end{subfigure}
\begin{subfigure}[b]{0.32\textwidth}
	\includegraphics[width=\textwidth]{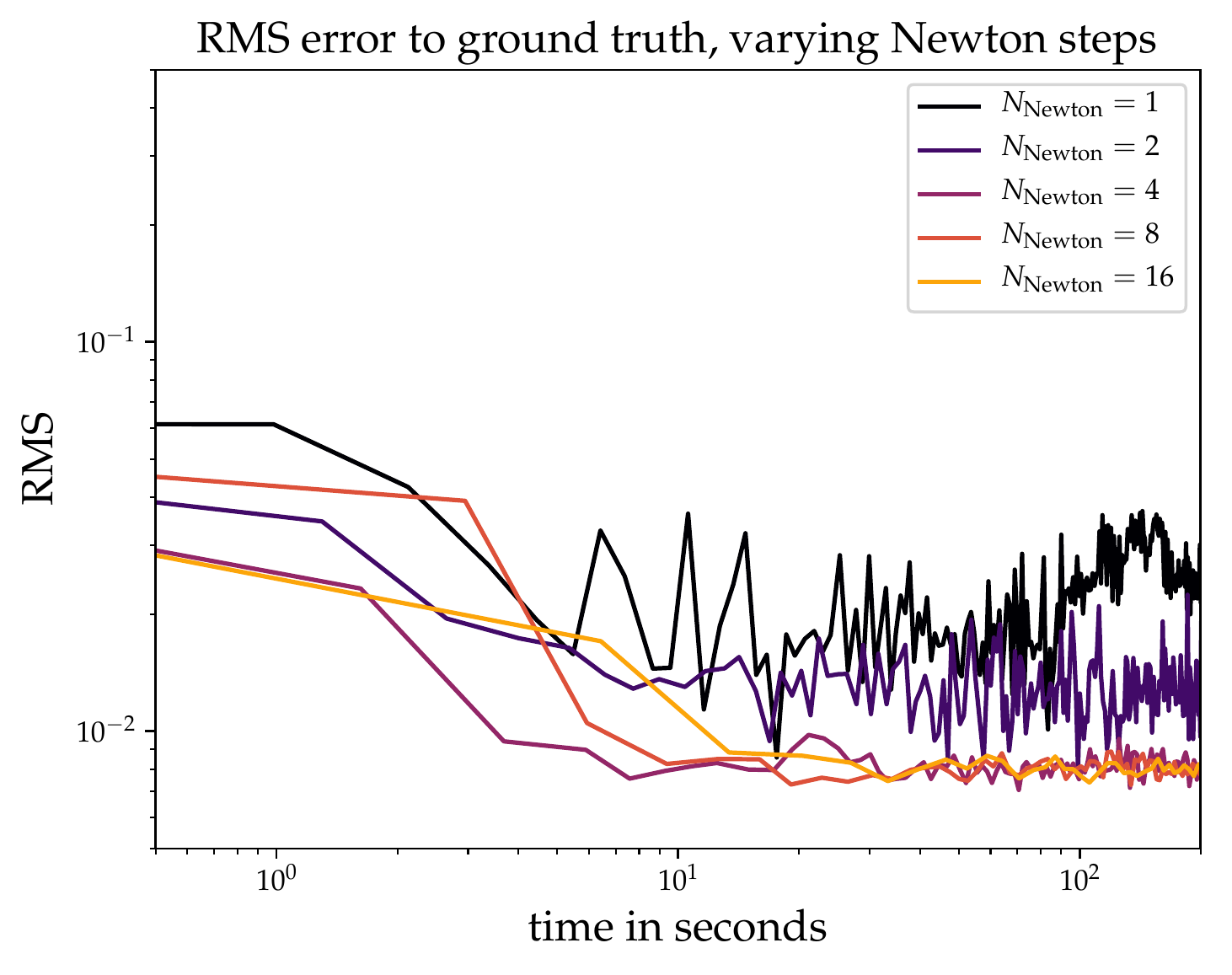}
\end{subfigure}
\begin{subfigure}[b]{0.32\textwidth}
	\includegraphics[width=\textwidth]{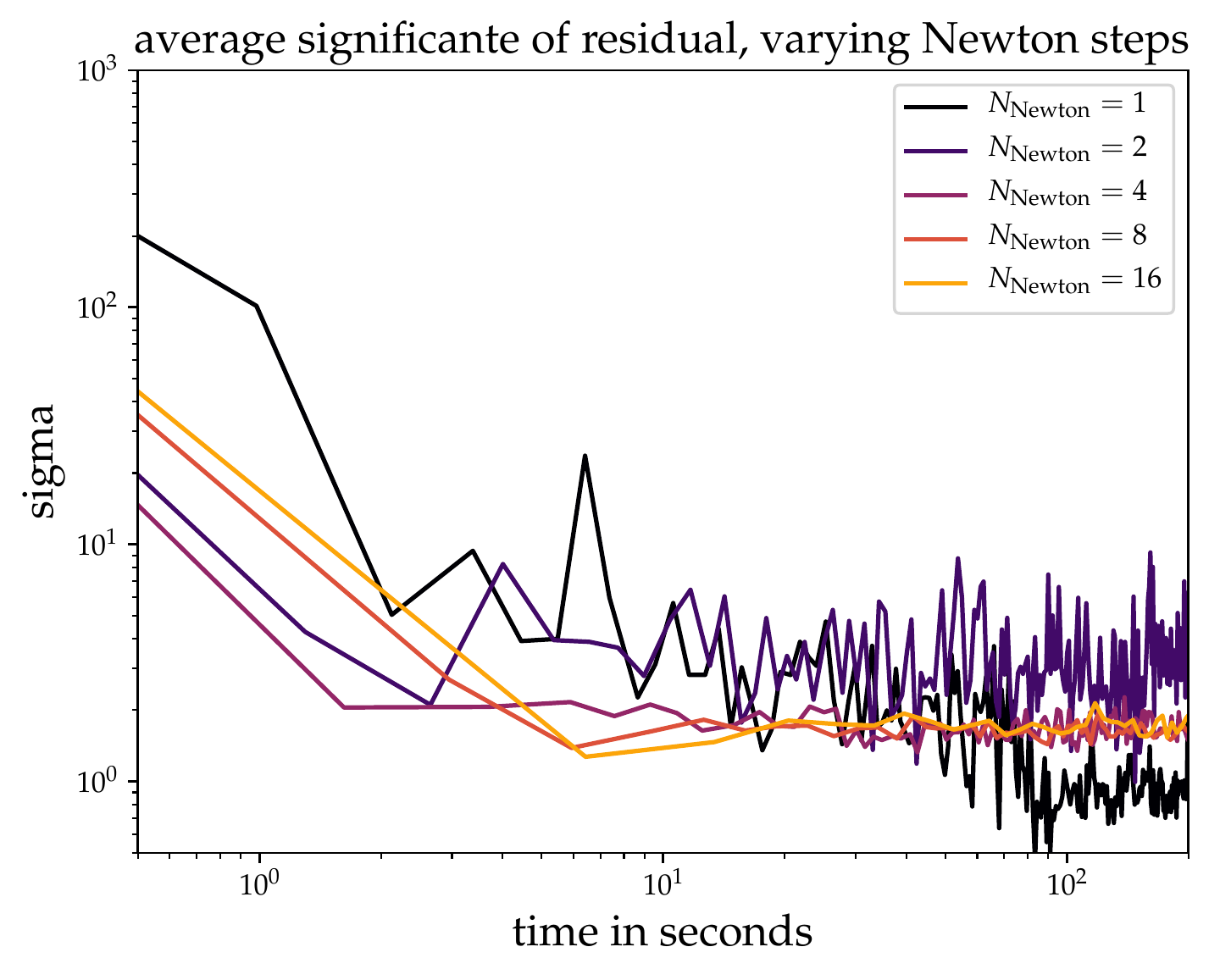}
\end{subfigure}
	\begin{subfigure}[b]{0.32\textwidth}
	\includegraphics[width=\textwidth]{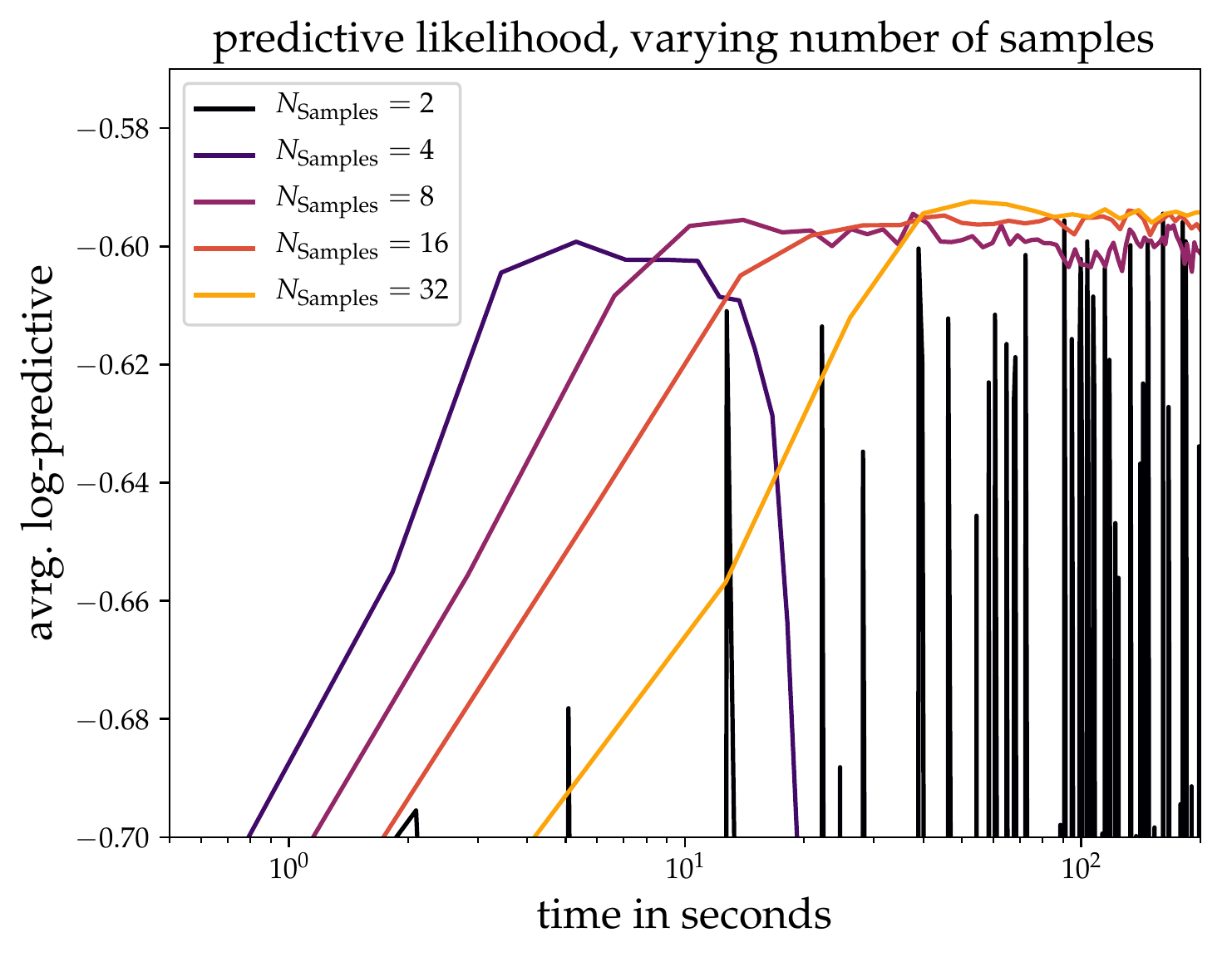}
\end{subfigure}
\begin{subfigure}[b]{0.32\textwidth}
	\includegraphics[width=\textwidth]{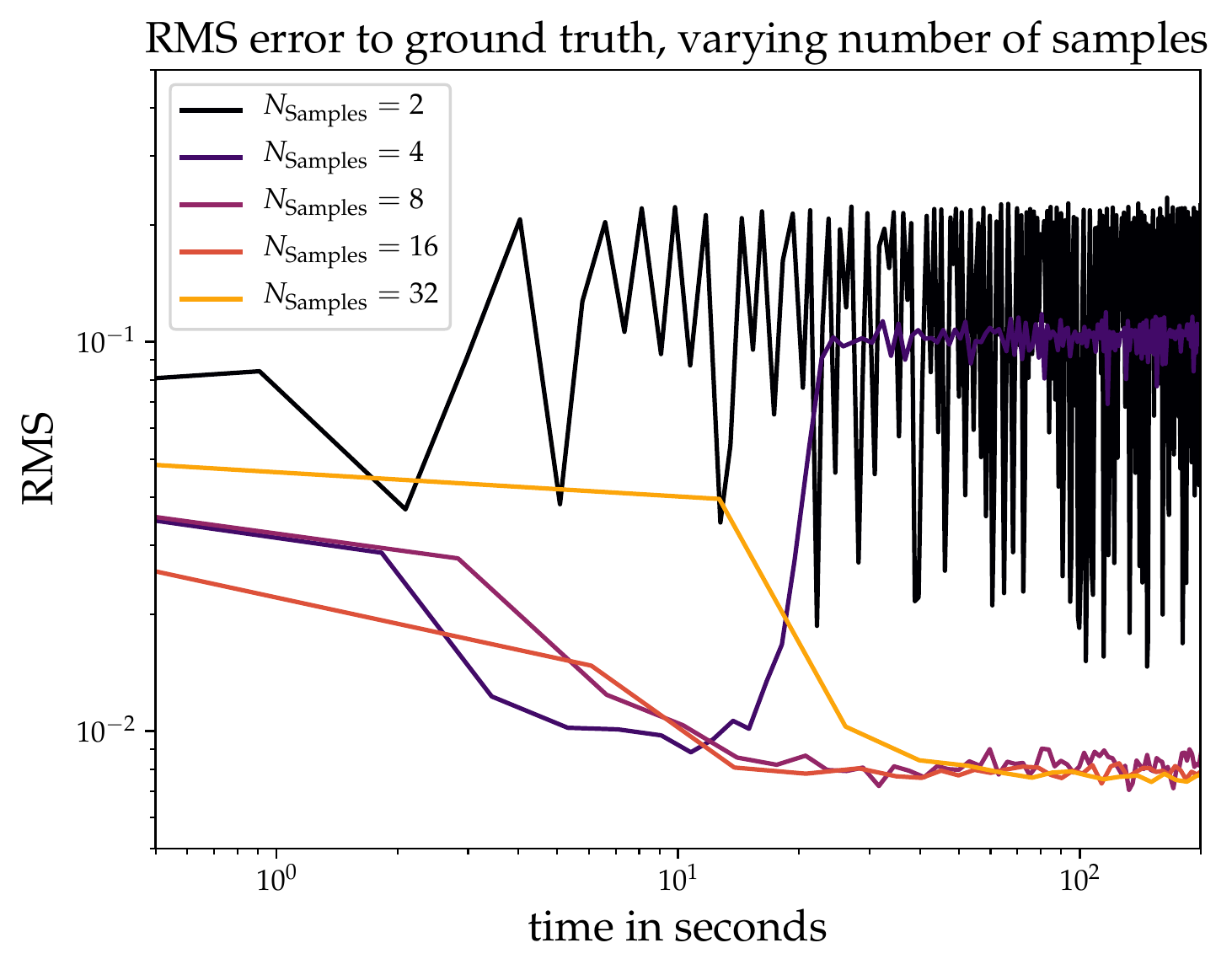}
\end{subfigure}
\begin{subfigure}[b]{0.32\textwidth}
	\includegraphics[width=\textwidth]{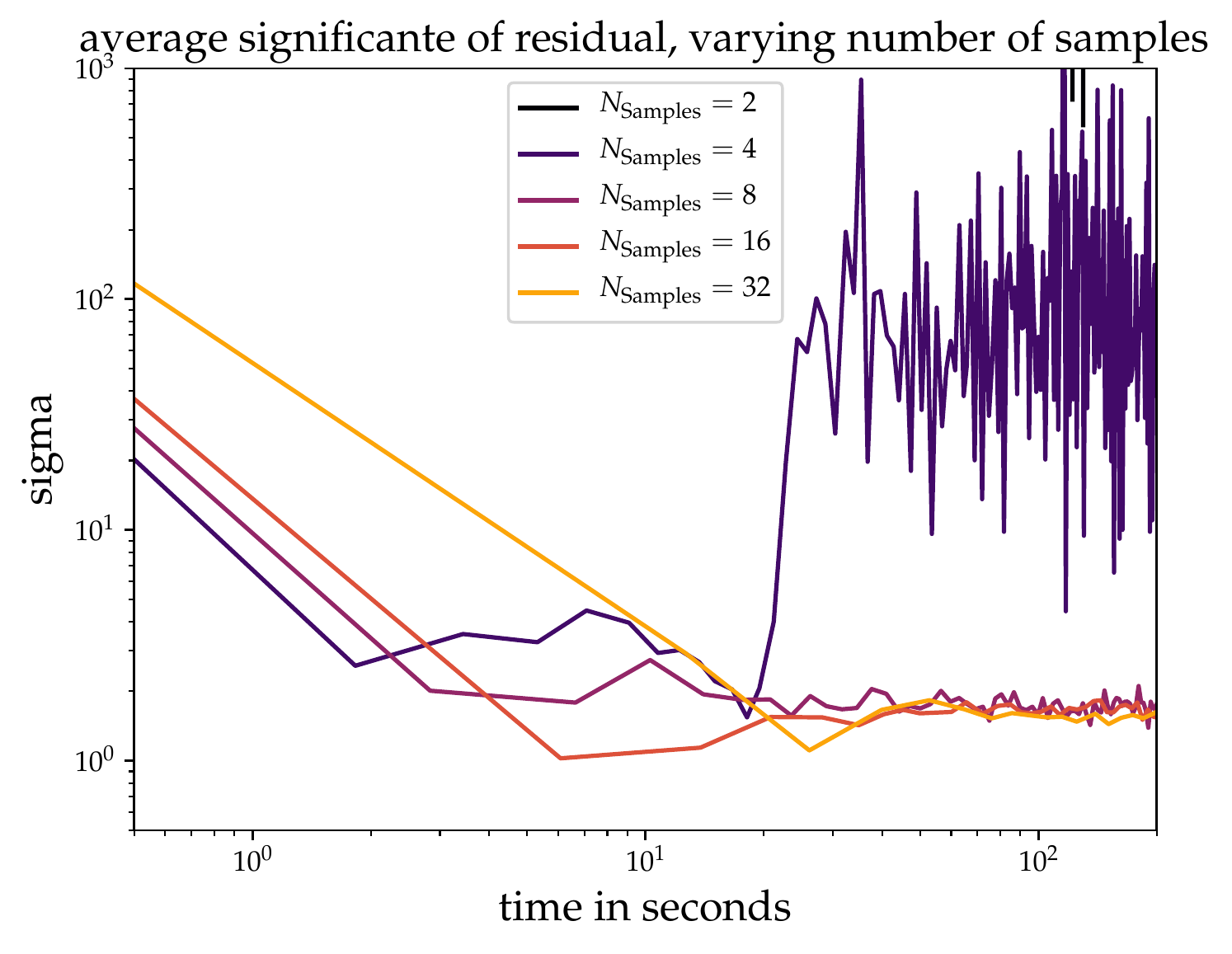}
\end{subfigure}
	\begin{subfigure}[b]{0.32\textwidth}
	\includegraphics[width=\textwidth]{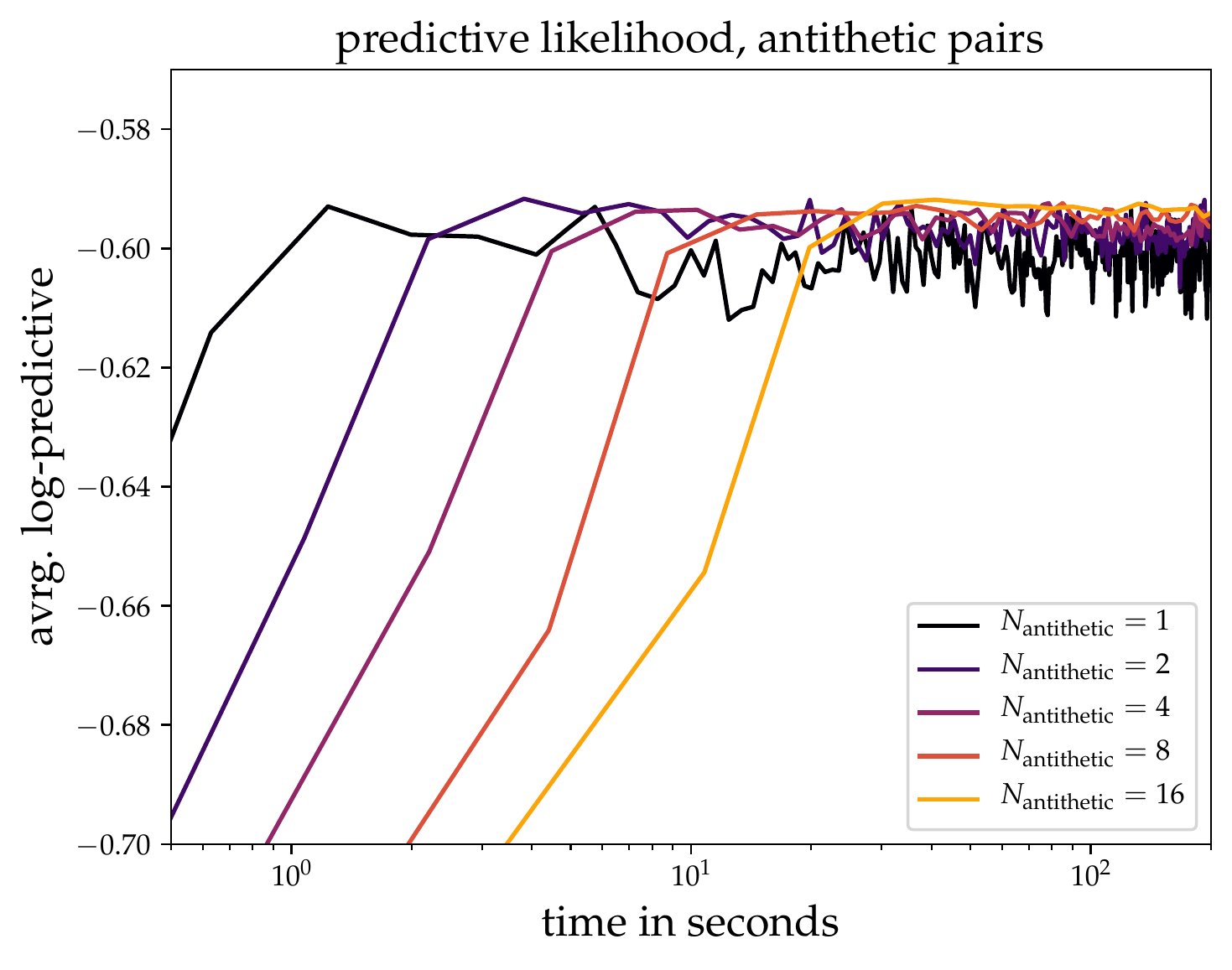}
\end{subfigure}
\begin{subfigure}[b]{0.32\textwidth}
	\includegraphics[width=\textwidth]{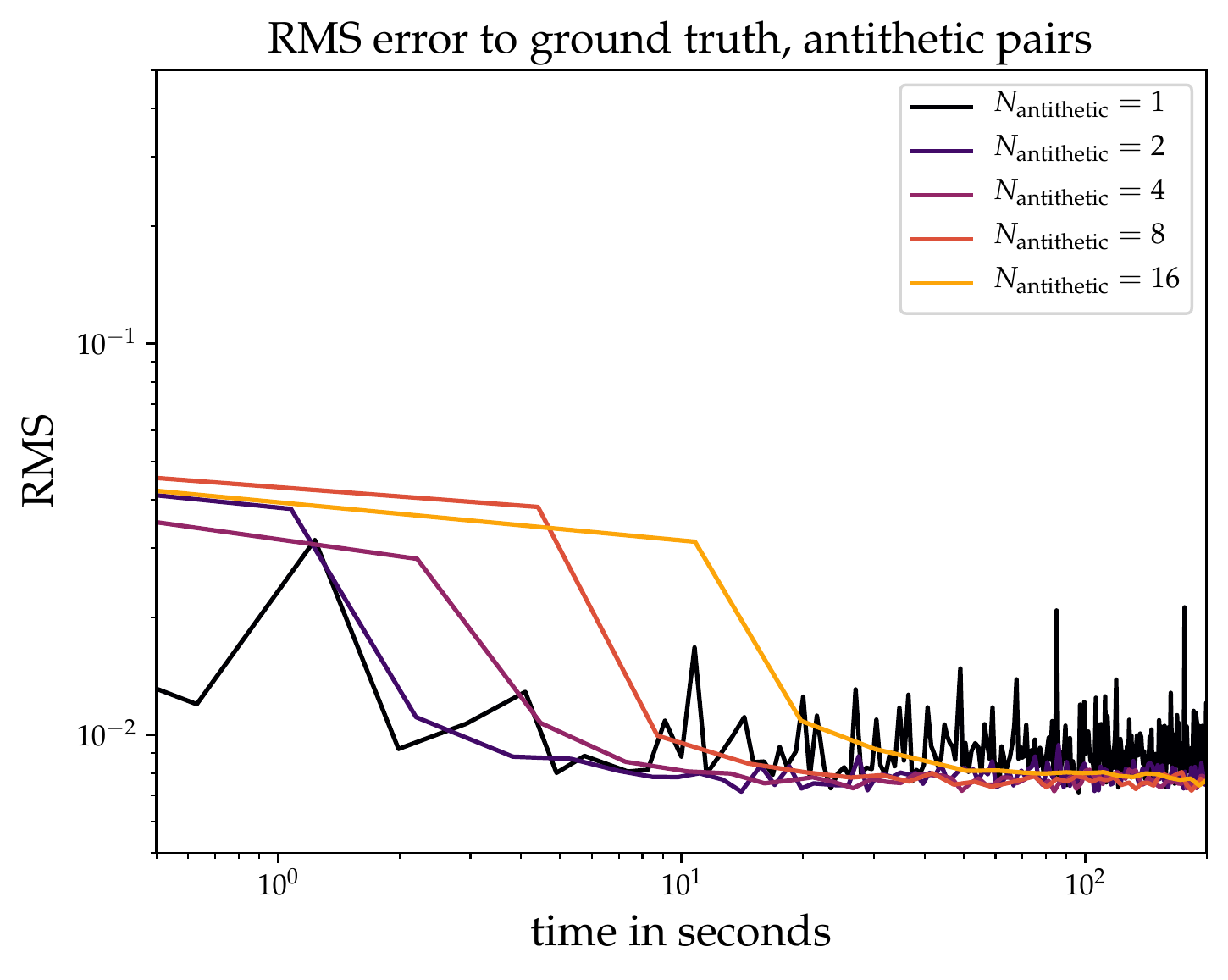}
\end{subfigure}
\begin{subfigure}[b]{0.32\textwidth}
	\includegraphics[width=\textwidth]{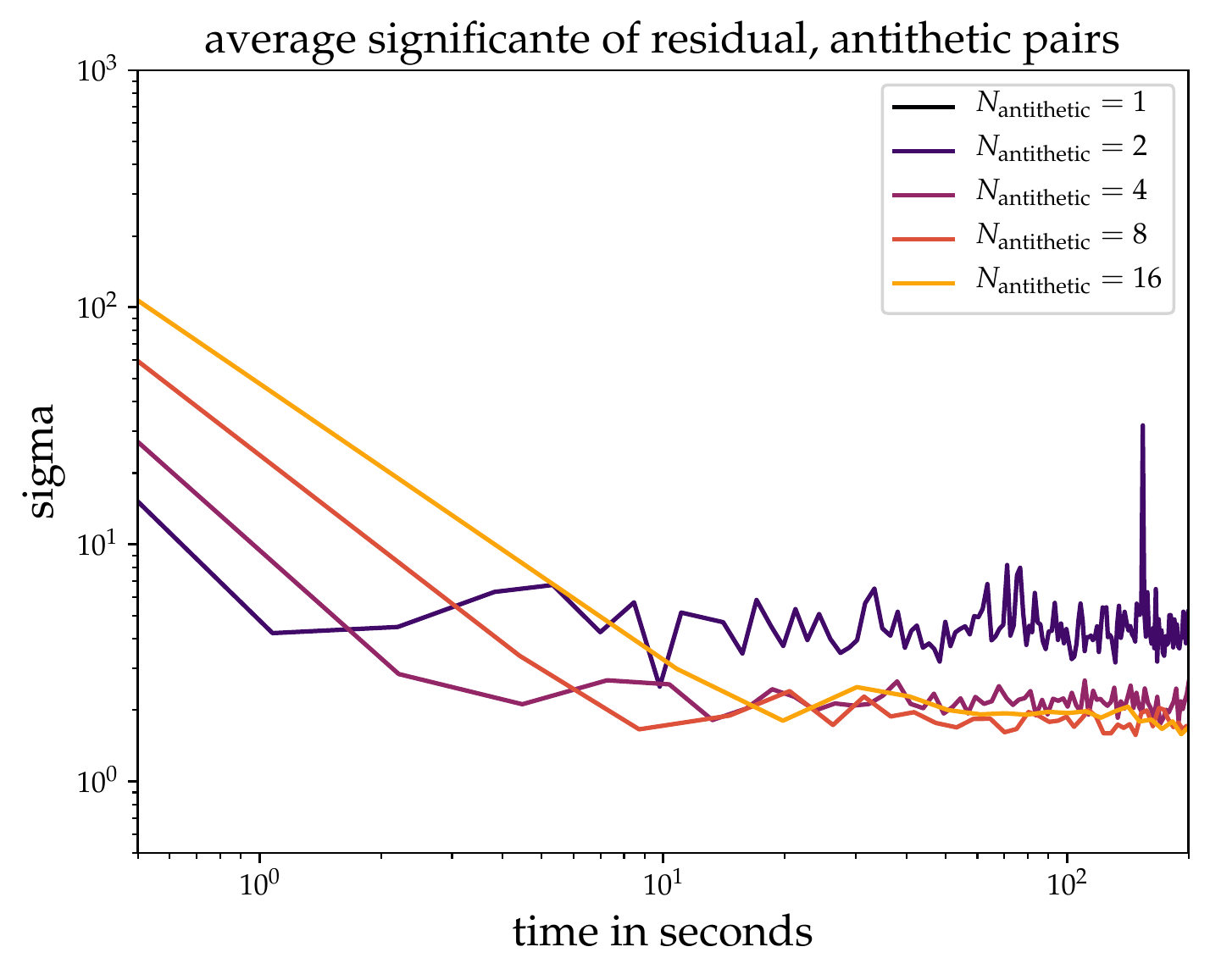}
\end{subfigure}
\caption{The results of the meta-parameter exploration. The different performance metrics are shown from left to right, the different meta-parameters from top to bottom.}
	\label{fig:BGPC_meta}
\end{figure}

MGVI requires a number of meta parameters that will affect the performance and accuracy of the method and we cannot provide a universally applicable recipe on how to set them. Here we want to showcase how the choice of a parameter tends to impact the method, but only in an isolated case. We also restrict ourselves to a discussion on individual parameters, not their interactions, as the possible combinations are overwhelming. To illustrate their impact we use the identical setup as in the last example, just a factor of $64$ smaller with $128\times128$ Gaussian process parameters and half that data points. We again track the predictive likelihood, RMS error to the ground truth, and average significance of the residual for different meta parameter settings.  Here we discuss the sampling accuracy, the number of natural gradient steps for a set of samples, the number of overall samples and the effect of using antithetic samples. We only vary one parameter per example, keeping all other parameters at some reasonable value. The default sampling accuracy are $30$ conjugate gradient steps, for a set of samples we make $10$ natural gradient steps and use $10$ independent samples without antithetic counterparts. The results are shown in Fig.~\ref{fig:BGPC_meta}.

The first meta parameter is the sampling accuracy, describing how many conjugate gradient steps are used to draw an approximate sample according to the covariance. Here one starts with a prior sample and every conjugate gradient iteration removes variance along the eigendirections corresponding to the consecutive largest eigenvalues of the metric. How many steps are required will strongly depend on the problem at hand and especially on the eigenspectrum of the metric. If it drops fast, only a few iterations are sufficient, otherwise more are required. The predictive likelihood, as well as the RMS error in the top row of Fig.~\ref{fig:BGPC_meta} show that too few iterations will affect the result, but increasing the number rapidly converges towards a common plateau. Already $9$ iterations seem to be sufficient in this example. Extremely interesting is the average significance in this case. Regardless of the sampling accuracy, the result will have a consistent error estimate, absorbing insufficient convergence into uncertainties and avoiding a misleading result. 

The next meta parameter is the number of natural gradient steps for a given set of samples. This number essentially controls how well an intermediate approximation converges before new samples are drawn at the obtained location. Newly drawn samples will usually not match the true posterior as well as the old samples, for which the KL was optimized, as they probe other directions and it takes some optimization steps to catch up, during which the problem itself is not yet further optimized. Taking too few steps will not lead to good results, as only the sampling stochasticity is chased.  This can be observed in the plots. Overall, a deeper convergence for an intermediate approximation reduces the variance of the results and converges better overall. One danger is the over-fitting of the sample realization, not collecting the progress in the mean parameter. This can mainly occur for a small number of samples, as the variances are not probed well.

The number of samples to estimate the KL divergence critically impacts the performance of MGVI. In terms of required computations, everything scales linearly with the number of used samples, so using as few as possible is desired. A single sample is certainly insufficient, as it does not define a variance and the result will be a MAP estimate shifted by the sampled residual, if the KL is fully optimized. The behavior of MGVI for different sample numbers are shown in the third row of Fig.~\ref{fig:BGPC_meta}. Clearly two and four samples are not enough to converge towards a reasonable solution. For more than eight samples it converges and it seems that more samples allow for deeper convergence and reduced stochastic behavior. It is worth noting that stochasticity is sometimes an advantage, as one can escape local minima, making it more reliable to finding good solutions. 

Finally, the last row shows the impact of using antithetic pairs of samples, using not only mean plus the residual as sample, but also minus the residual. This way the mean of the samples and the mean parameter always coincide, stabilizing the gradient estimate significantly, while requiring to draw only half the number of samples. This way MGVI already converges towards reasonable results using only one single sample together with its antithetic counterpart, as shown in the plots. It is still relatively noisy, but compared to using two independent samples, as shown in the row above, far more robust. More samples again reduce the stochasticity even further.

Overall, a higher accuracy, more samples, or more steps are always favorable for higher accuracy of the approximation, but they come at the price of computational effort, so the hard task is to counterbalance those two contradicting goals. Especially towards the beginning of the procedure, high precision might not be needed, as the landscape changes rapidly anyway, but towards the end, as everything starts to settle down and converge, it might be worth to invest into higher accuracy. In our experience we find it useful to gradually tune up the parameters, especially the number of samples to be fast and inaccurate in the beginning and then converge by adding samples, but how to optimally steer MGVI in general is unclear.
\subsection{Non-Negative Matrix Factorization}
In Non-Negative Matrix Factorization models, the data is described as a positive mixture of positive components, or factors. The goal is to find a lower-dimensional description of the data, which can be used to predict unobserved values. The data $d$ should be described by a data matrix $D$, which is the product of a mixture matrix $M$ and a component matrix $C$, which are to be learned:
\begin{align}
D = MC \text{.}
\end{align}
We choose a Gamma-Poisson model, assuming a Poisson likelihood and Gamma-priors on all entries of the matrices, enforcing positivity on all quantities. 
The problem is standardized by reparametrization defined by the inverse CDF of the Gamma distribution and CDF of the standard Gaussian. Both functions do not have an analytic expression, but can be approximated numerically. 

\begin{align}
M &= \mathcal{F}^{-1}_{\mathrm{Gamma}(M\vert \alpha_M,\beta_M)} \circ \mathcal{F}_{\mathcal{G}(\xi_M,\mathbb{1})}(\xi_M) \equiv f_M(\xi_M) \\
C &= \mathcal{F}^{-1}_{\mathrm{Gamma}(C\vert \alpha_C,\beta_C)} \,\,\,\circ \mathcal{F}_{\mathcal{G}(\xi_C,\mathbb{1})}(\xi_C) \,\,\,\equiv f_C(\xi_C) \text{\quad .}
\end{align}
These equations are to be read element-wise for every matrix entry. The standardized problem information then reads
\begin{align}
\mathcal{H}(d,\xi) = d^\dagger \mathrm{ln} \left(f_M(\xi_M)f_C(\xi_C)\right) + 1^\dagger \left(f_M(\xi_M)f_C(\xi_C)\right) + \frac{1}{2} \xi^\dagger \mathbb{1} \xi \text{\quad .}
\end{align}
Here $\xi$ is the concatenation of $\xi_C$ and $\xi_M$. As in the first example, we have again a Poisson likelihood and its metric is given by Eq.~\ref{poisson-metric}.
We apply this model to the Frey face data set, consisting of 1965 images of a sequence of facial expressions in a resolution of 28x20 pixels, assuming ten components. All parameters of the Gamma distribution are chosen to be $1$, and we randomly mask $10\%$ of pixels to calculate the predictivity of different methods per elapsed time. In addition to that, the bottom part of one frame is fully covered  by the mask, and we will show how well it is recovered. Overall the model has $25160$ free parameters to be learned and we compare the performance of MGVI to mean-field ADVI. In this example we do have relatively good data, so it is not as relevant to frequently refresh the samples to explore the uncertainty and we can afford to optimize deeper in each global iteration to achieve overall faster convergence. Therefore we initially perform $10$ natural gradient steps together with one pair of antithetic samples, compared to the three steps in the previous example, but otherwise we also increase the number of samples starting after twenty iterations. The initial sampling accuracy are $50$ iterations, increasing it to $200$ towards the end.

\begin{figure}
	\centering
\begin{subfigure}[b]{0.49\textwidth}
	\includegraphics[width=\textwidth]{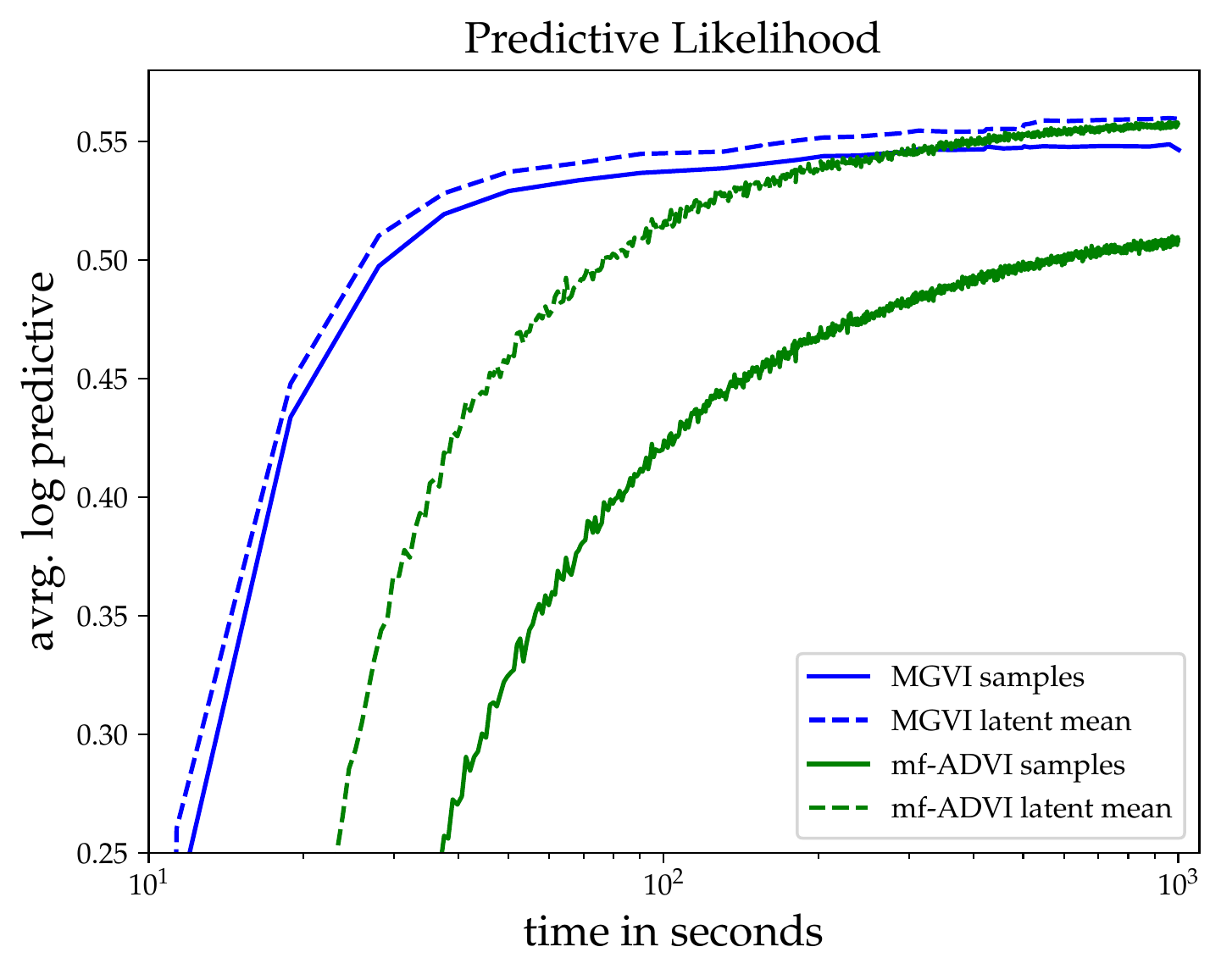}
	\caption{Predictive likelihood of MGVI compared to mf-ADVI in seconds. Dashed lines are the predictivity of the mean, solid lines of the samples.}
\end{subfigure}
\begin{subfigure}[b]{0.42\textwidth}
\begin{subfigure}[b]{0.32\textwidth}
	\flushleft
	\includegraphics[width=\textwidth]{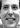}
		\caption{ ground\\ truth.}
\end{subfigure}
\begin{subfigure}[b]{0.32\textwidth}
	\includegraphics[width=\textwidth]{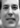}
	\caption{MGVI\\ mean.}
\end{subfigure}
\begin{subfigure}[b]{0.32\textwidth}
	\includegraphics[width=\textwidth]{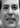}
	\caption{mf-ADVI\\ mean.}
\end{subfigure}

\begin{subfigure}[b]{0.32\textwidth}
	\includegraphics[width=\textwidth]{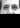}
	\caption{masked\\ data.}
\end{subfigure}
\begin{subfigure}[b]{0.32\textwidth}
	\includegraphics[width=\textwidth]{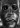}
	\caption{MGVI\\ uncertainty.}
\end{subfigure}
\begin{subfigure}[b]{0.32\textwidth}
	\includegraphics[width=\textwidth]{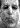}
		\caption{mf-ADVI \\uncertainty.}
\end{subfigure}
\end{subfigure}

\begin{subfigure}[b]{0.45\textwidth}
	\includegraphics[width=\textwidth]{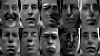}
	\caption{MGVI components.}
	\vspace{5mm}
\end{subfigure}
\begin{subfigure}[b]{0.45\textwidth}
	\includegraphics[width=\textwidth]{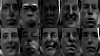}
	\caption{mf-ADVI components.}
	\vspace{5mm}
\end{subfigure}
\caption{The predictivity of both methods (top left), means and standard deviations for a certain frame together with ground truth and data (top right), and recovered components (bottom).}
	\label{fig:freyface}
\end{figure}

 The predictive likelihoods during the optimization for both methods, the results on the half masked frame, as well as the recovered components  are shown in Fig.~\ref{fig:freyface}. The predictivity of the MGVI samples and mean converge rapidly towards the same value, indicating low uncertainty. After $40$ seconds MGVI seems relatively converged as the slope strongly decreases. The predictivity of the  mf-ADVI mean achieves comparable levels to MGVI after $200$ seconds, but the predictive likelihood of the samples is significantly lower. As the discrepancy between predictivity of the mean and the samples are a proxy to the variance of the distribution, it seem that mf-ADVI severely struggles to compress towards the posterior mode, crippling down the overall convergence. MGVI does not have this problem, as the covariance adapts to the environment of the mean, and it can therefore contract towards the posterior mode far more rapidly.
Regarding the half masked frame, the mean for both methods matches very closely. Interesting is the pixel-wise standard deviation, shown below the mean. For MGVI it is clearly structured and aligns with regions of facial variability, for example around the mouth and the eyebrows. The variance is especially high in the masked region, whereas mf-ADVI shows less pronounced features, and even lower variance within the masked half.

\subsection{Hierarchical logistic regression}
In this example we discuss  two hierarchical logistic regression problems involving polling data from the US 1988 presidential election, using the models discussed in \citet{polldata} and we will follow the analysis of \citet{ADVI}. The data set involves $13544$ points on age, gender, ethnicity, education, region, state, and polling behavior. We consider two logistic regression models of different complexity to predict the polling behavior. The smaller model contains only information on state, gender and ethnicity, allowing full posterior sampling with HMC as reference. The larger model utilizes the full data set and it allows us insight into the convergence behavior of MGVI. 

The likelihood is given by a Bernoulli distribution, as stated in Eq.~\ref{eq:Bernoulli} and corresponding metric is given in Eq.~\ref{eq:BernoulliMetric}. The data is the polling result and it is modeled by a rate $\mu$, depending via a logit link on regression coefficients and the design matrix containing $X$.
\begin{align}
\mu = \sigma(X^\dagger \beta)\text{\quad .}
\end{align}

Here $\sigma$ is again a sigmoid function.
 
\subsubsection{A simple model}
For the simple model the rate is described by only a subset of all categories
\begin{align}
\mu = \sigma \left(\beta_0 + x_{\mathrm{gender}} \beta_{\mathrm{gender}}  + x_{\mathrm{ethnicity}} \beta_{\mathrm{ethnicity}} + \beta_\mathrm{state}[x_\mathrm{state}]\right)\text{\quad ,}
\end{align}
with binary data on gender and ethnicity, and multi-class labels on the state. Additionally we set a standard Gaussian prior on $\beta_0$, $ \beta_{\mathrm{ethnicity}}$ and $ \beta_{\mathrm{gender}}$. To make it a hierarchical problem, all $\beta_\mathrm{state}$ coefficients follow also independent Gaussian priors, but with a priori unknown standard deviation $\sigma_\mathrm{state}$, shared among them. We give it a uniform prior on the unit interval. Compared to the model described in \citet{polldata}, we choose more restrictive priors for convergence reasons, especially for HMC, but also MGVI. We will elaborate on this later when discussing the full model. 
\begin{figure}
	\centering
	\begin{subfigure}[b]{0.32\textwidth}
	\includegraphics[width=\textwidth]{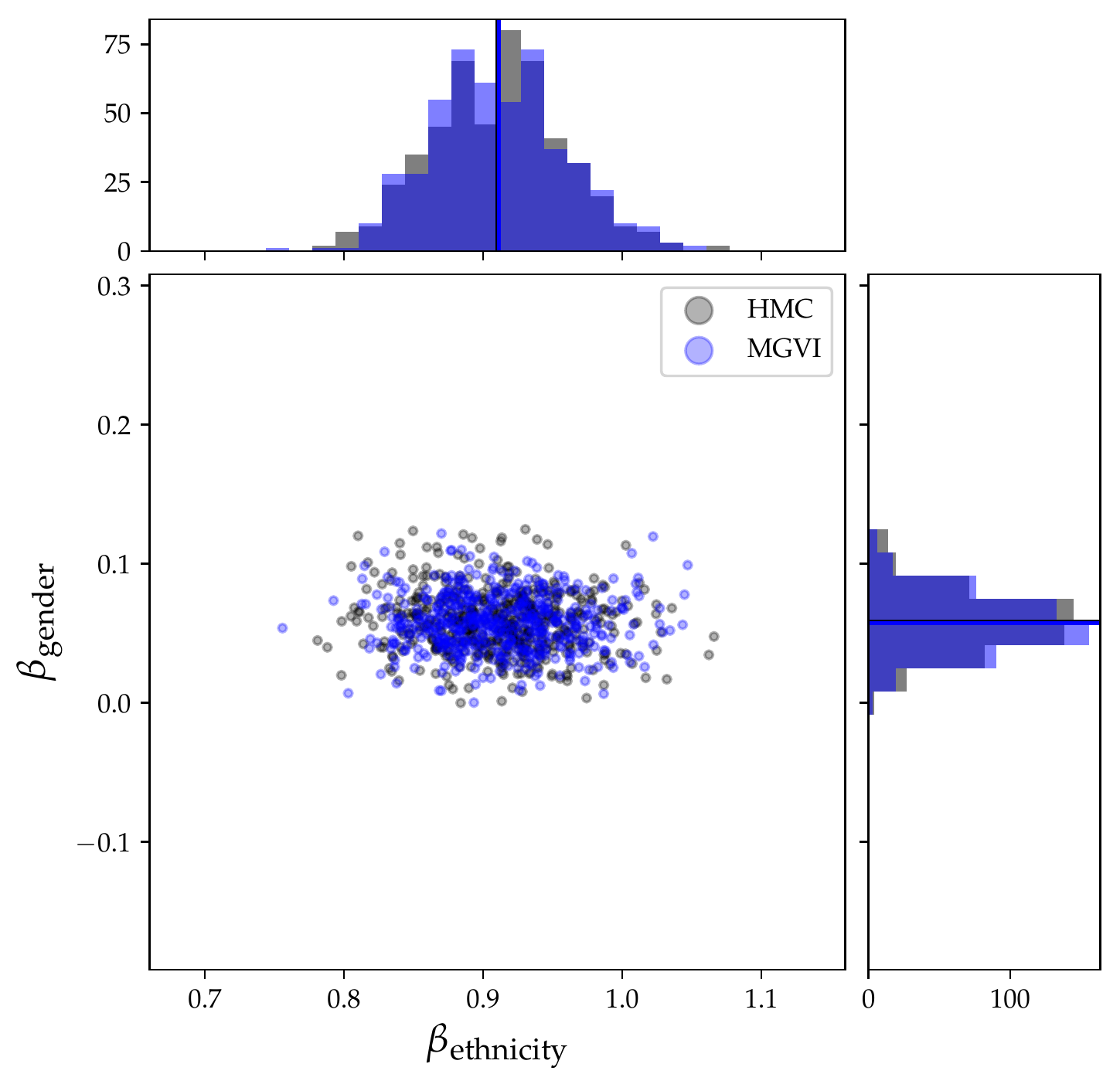}
\end{subfigure}
\begin{subfigure}[b]{0.32\textwidth}
	\includegraphics[width=\textwidth]{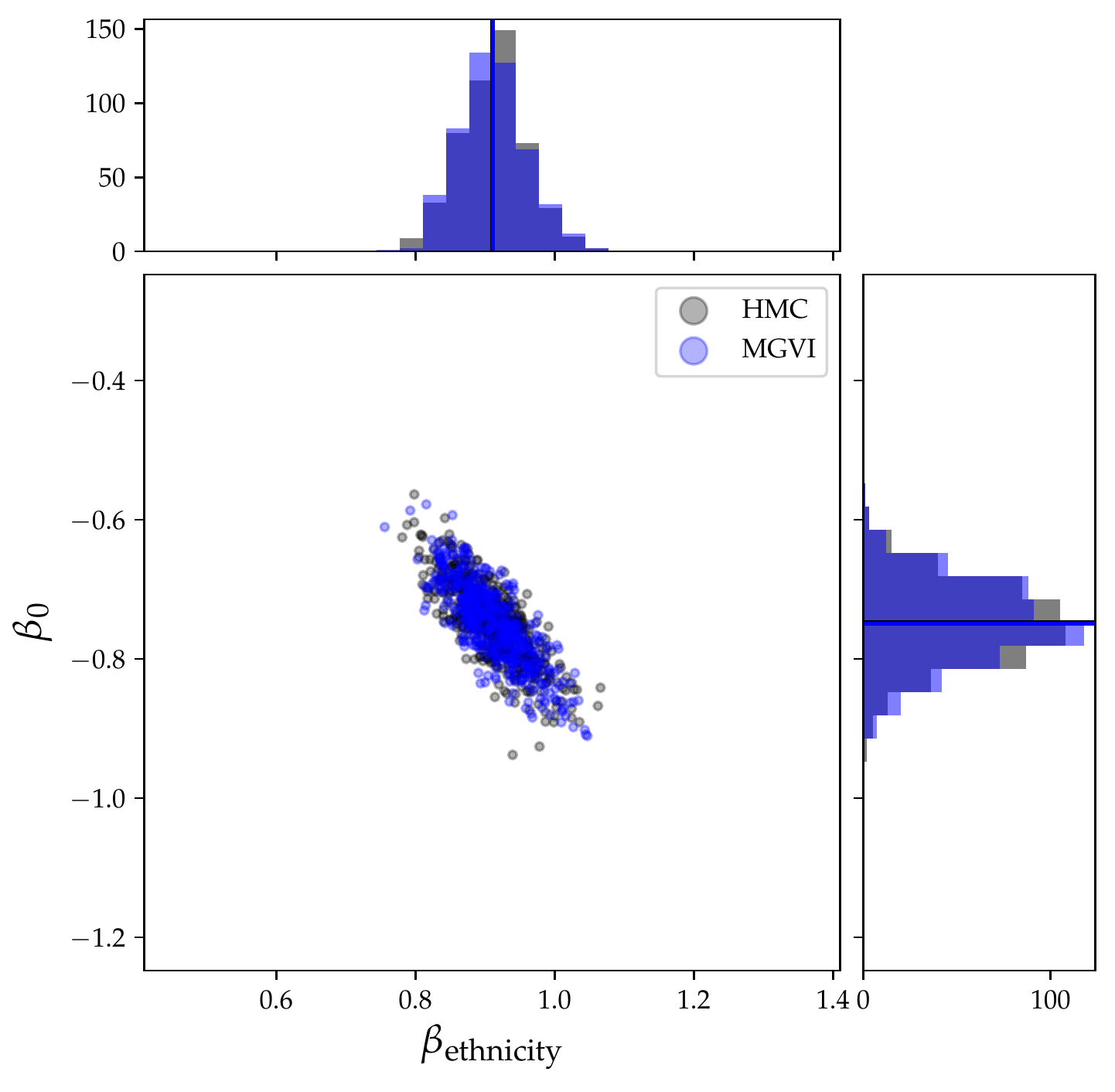}
\end{subfigure}
\begin{subfigure}[b]{0.32\textwidth}
	\includegraphics[width=\textwidth]{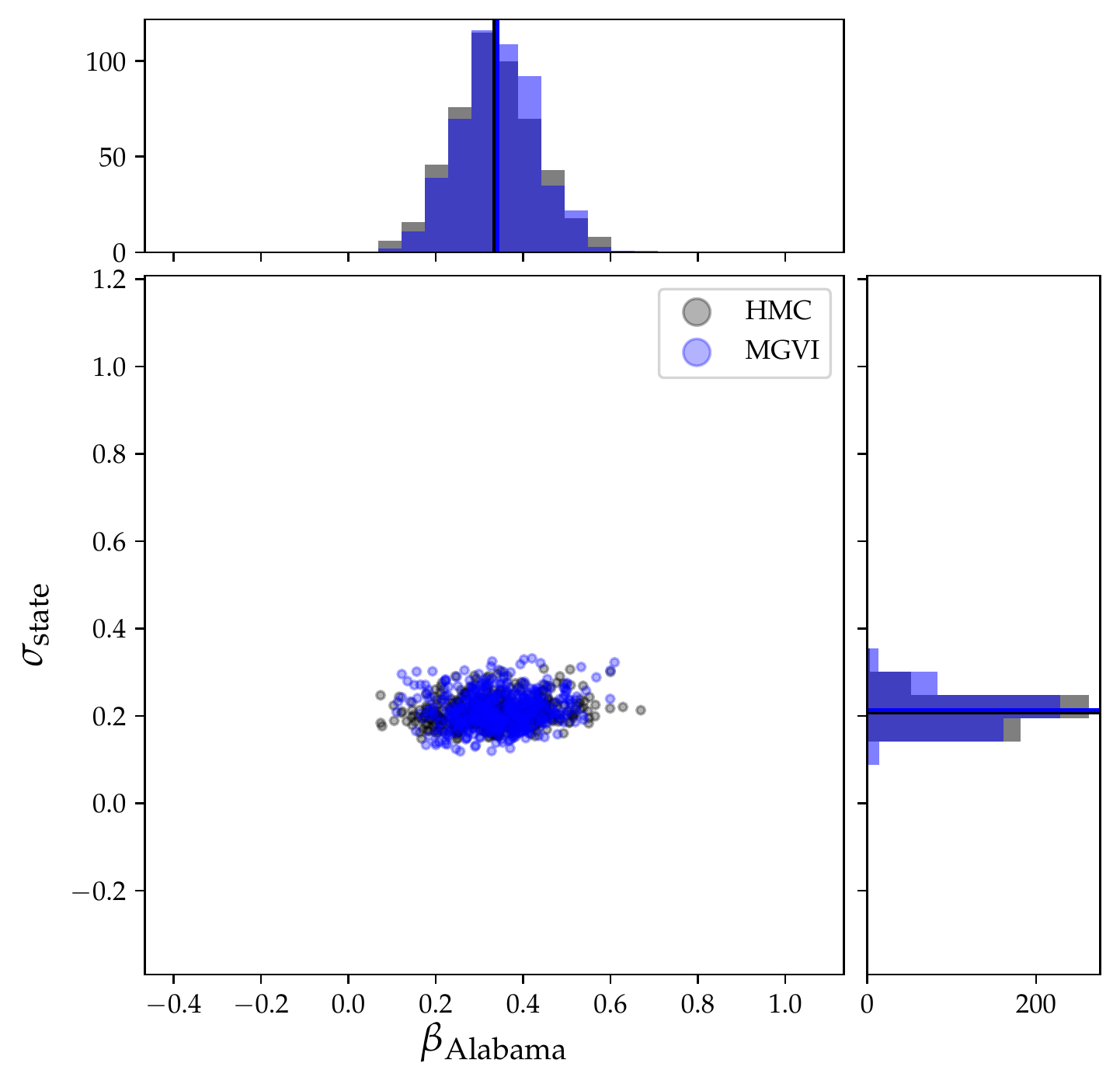}
\end{subfigure}
\begin{subfigure}[b]{0.32\textwidth}
	\includegraphics[width=\textwidth]{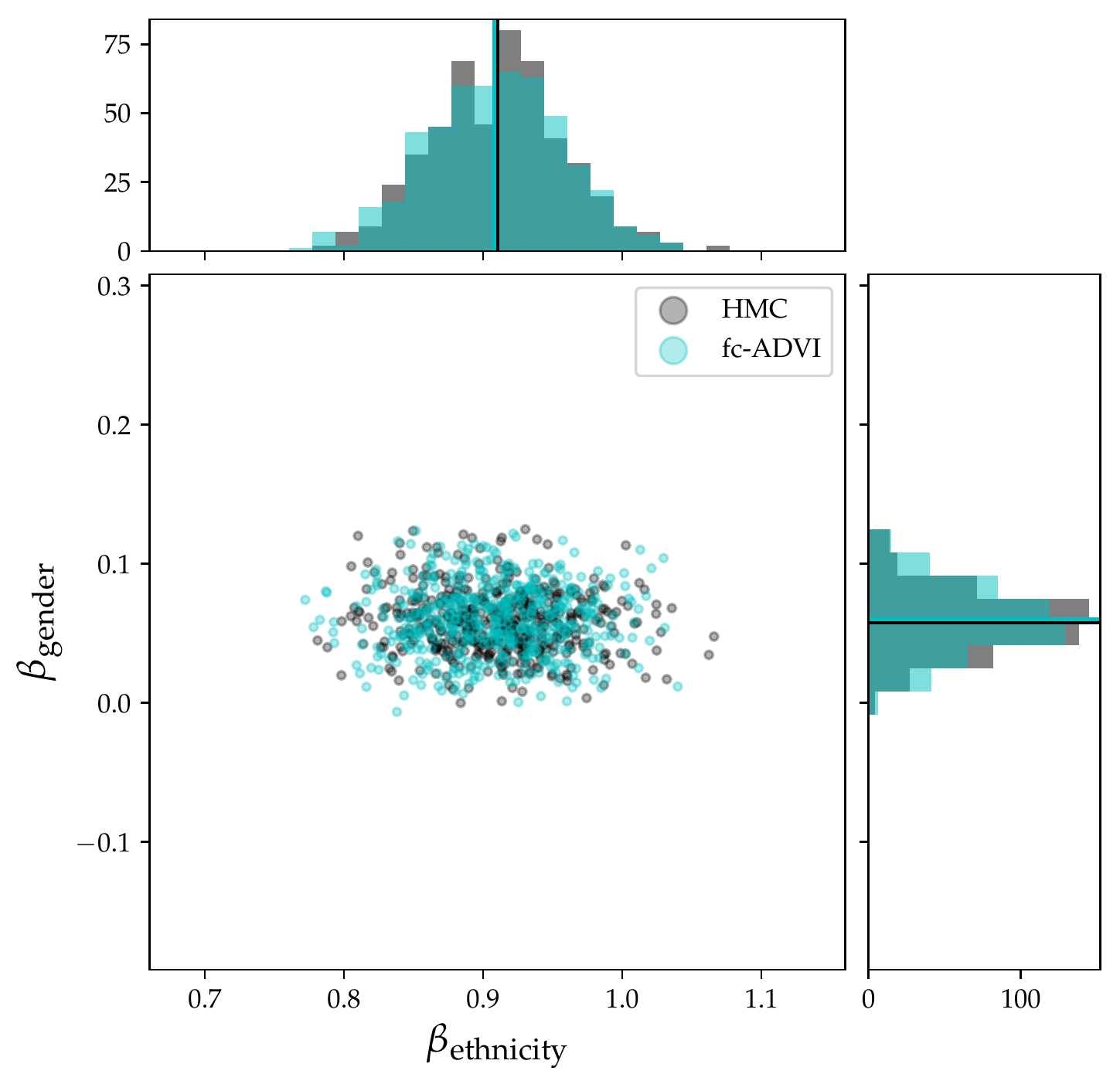}
\end{subfigure}
\begin{subfigure}[b]{0.32\textwidth}
	\includegraphics[width=\textwidth]{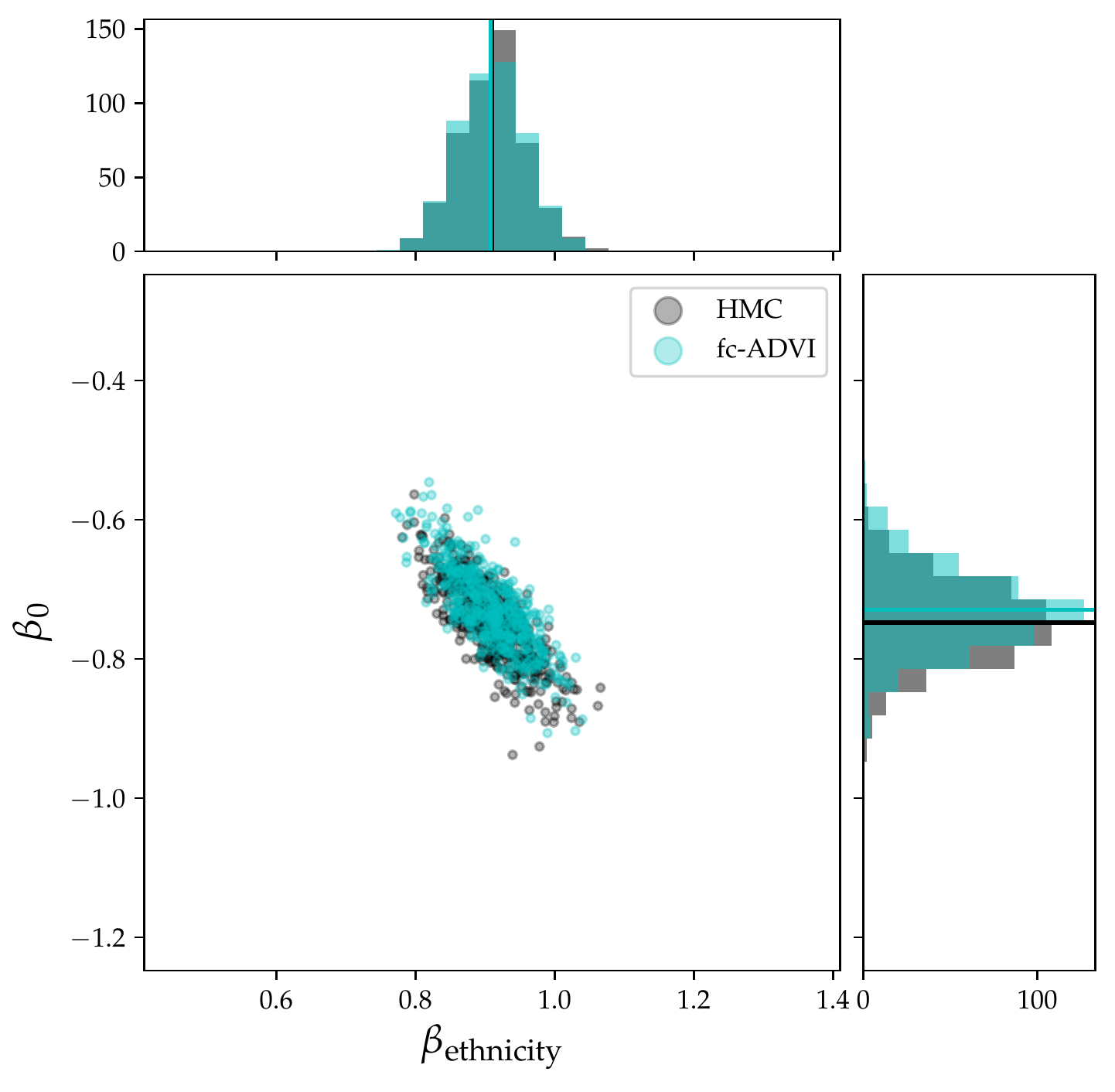}
\end{subfigure}
\begin{subfigure}[b]{0.32\textwidth}
	\includegraphics[width=\textwidth]{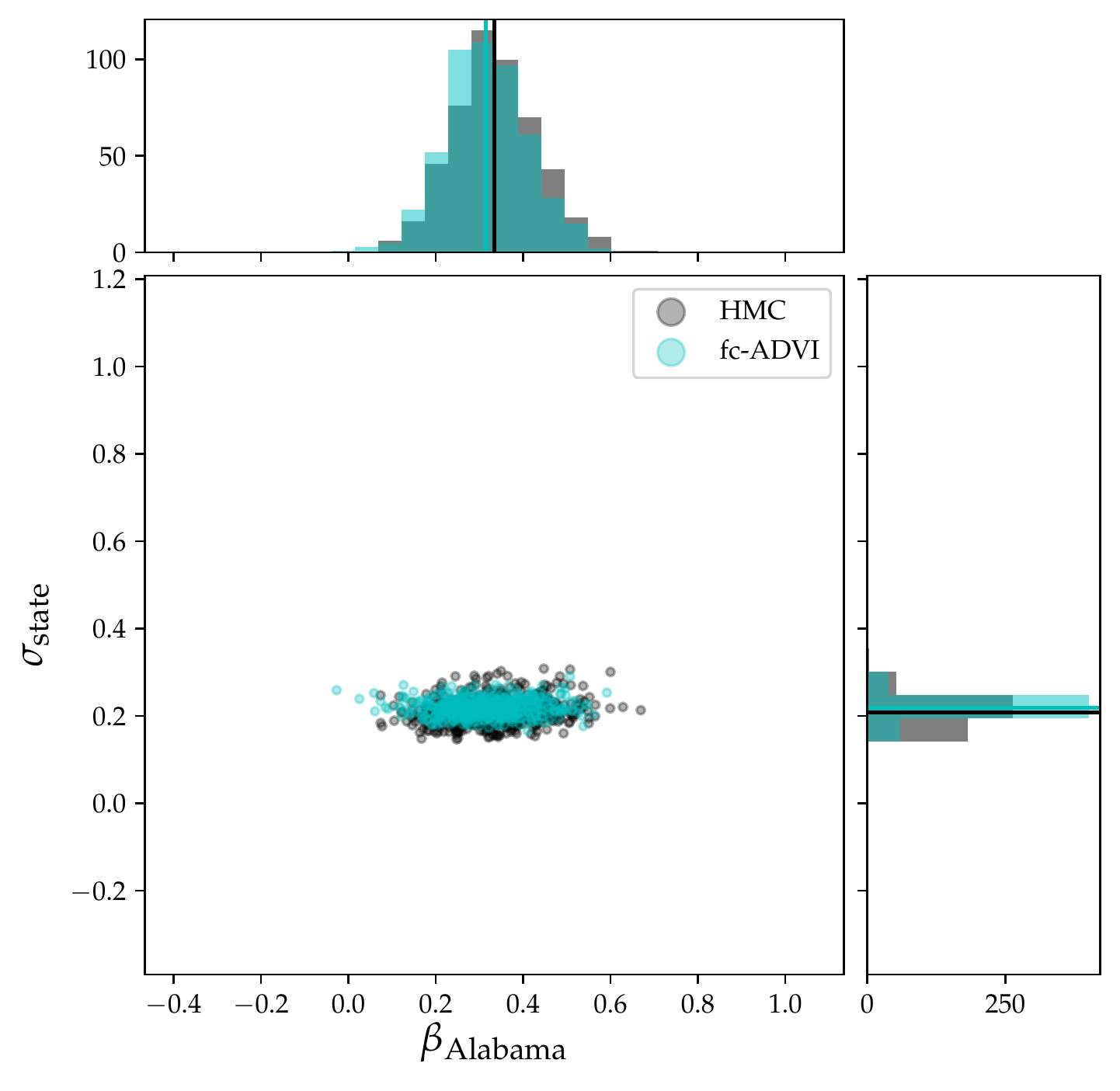}
\end{subfigure}
\begin{subfigure}[b]{0.32\textwidth}
	\includegraphics[width=\textwidth]{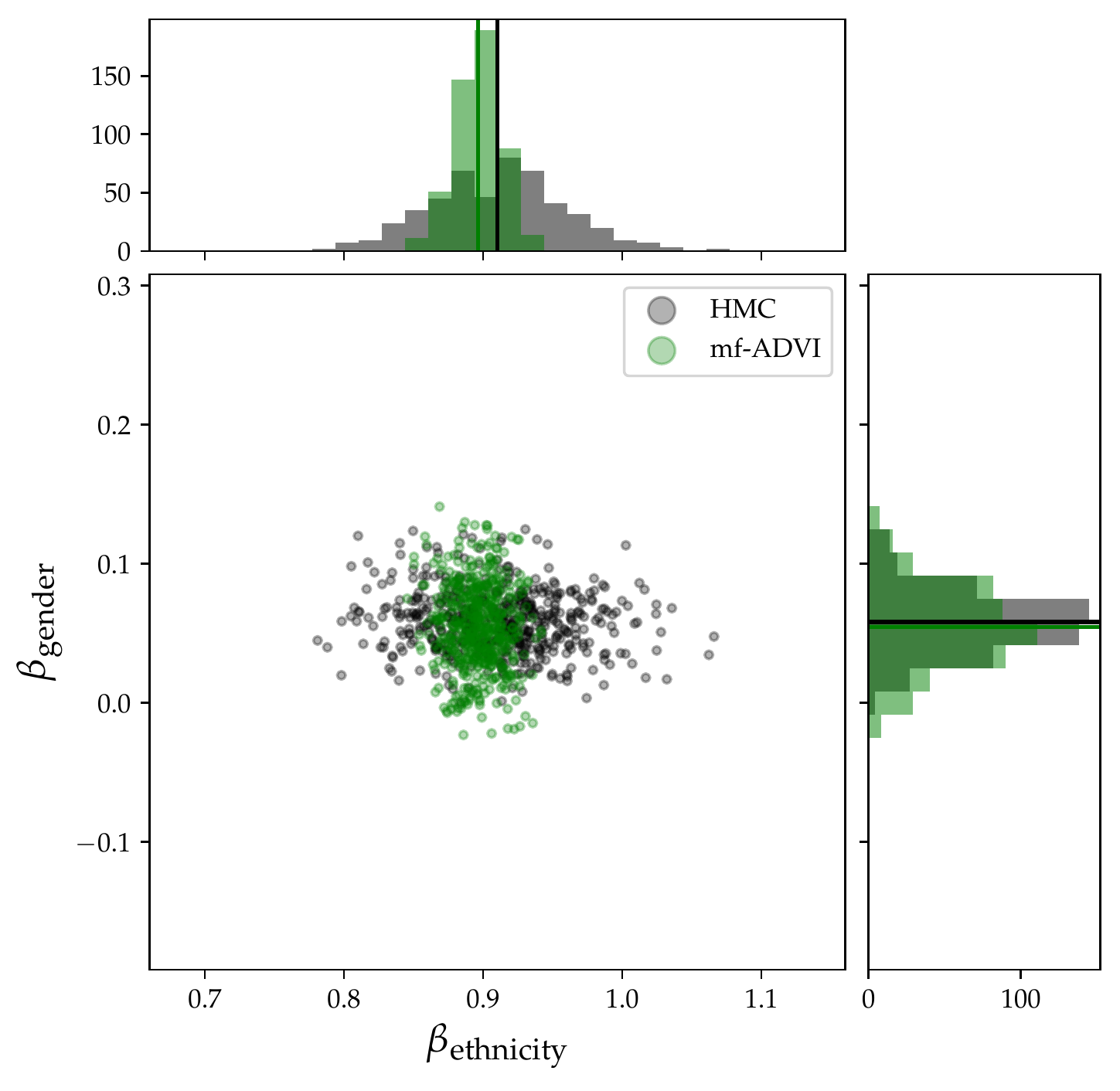}
\end{subfigure}
\begin{subfigure}[b]{0.32\textwidth}
	\includegraphics[width=\textwidth]{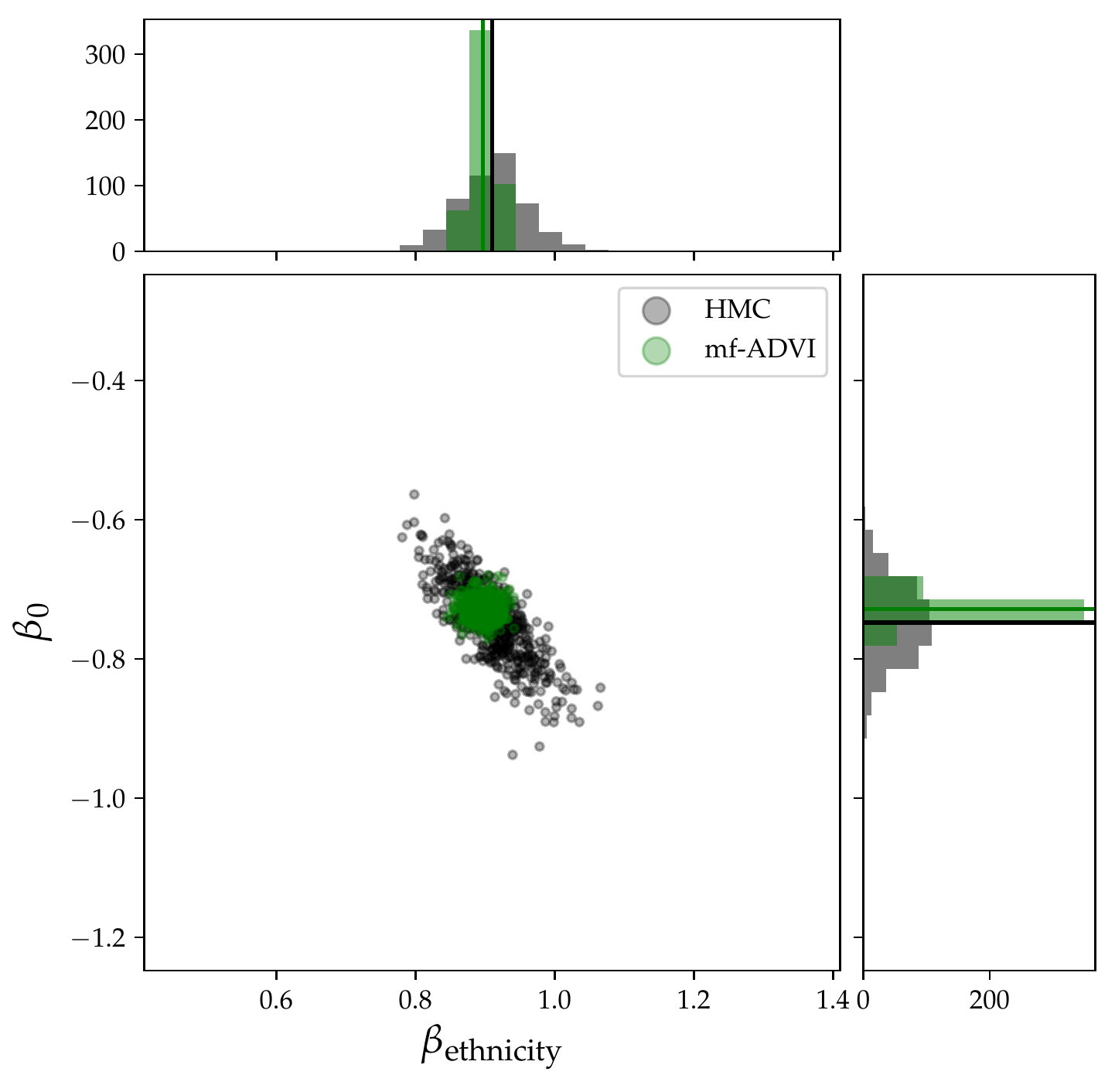}
\end{subfigure}
\begin{subfigure}[b]{0.32\textwidth}
	\includegraphics[width=\textwidth]{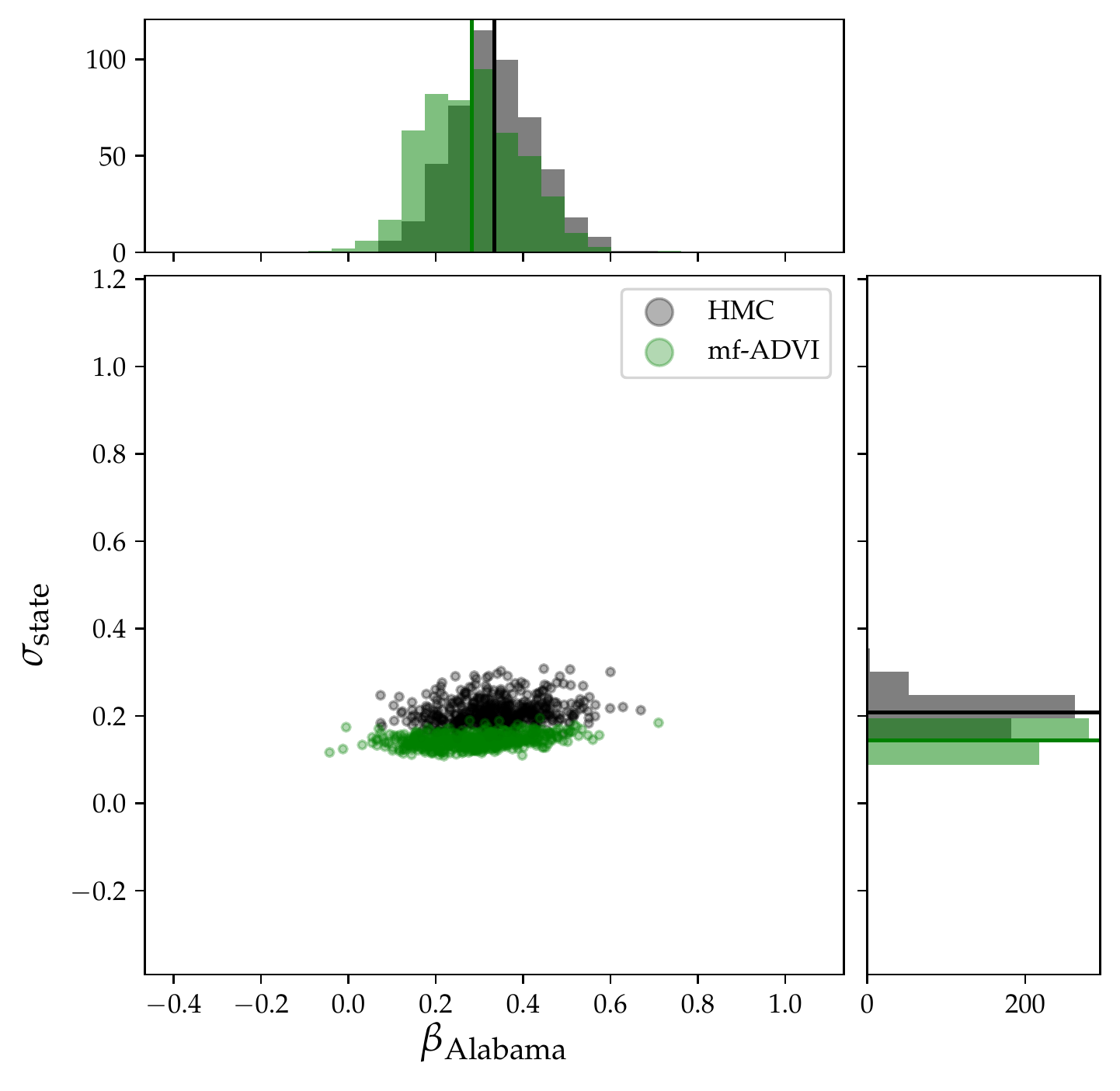}
\end{subfigure}
\begin{subfigure}[b]{0.32\textwidth}
	\includegraphics[width=\textwidth]{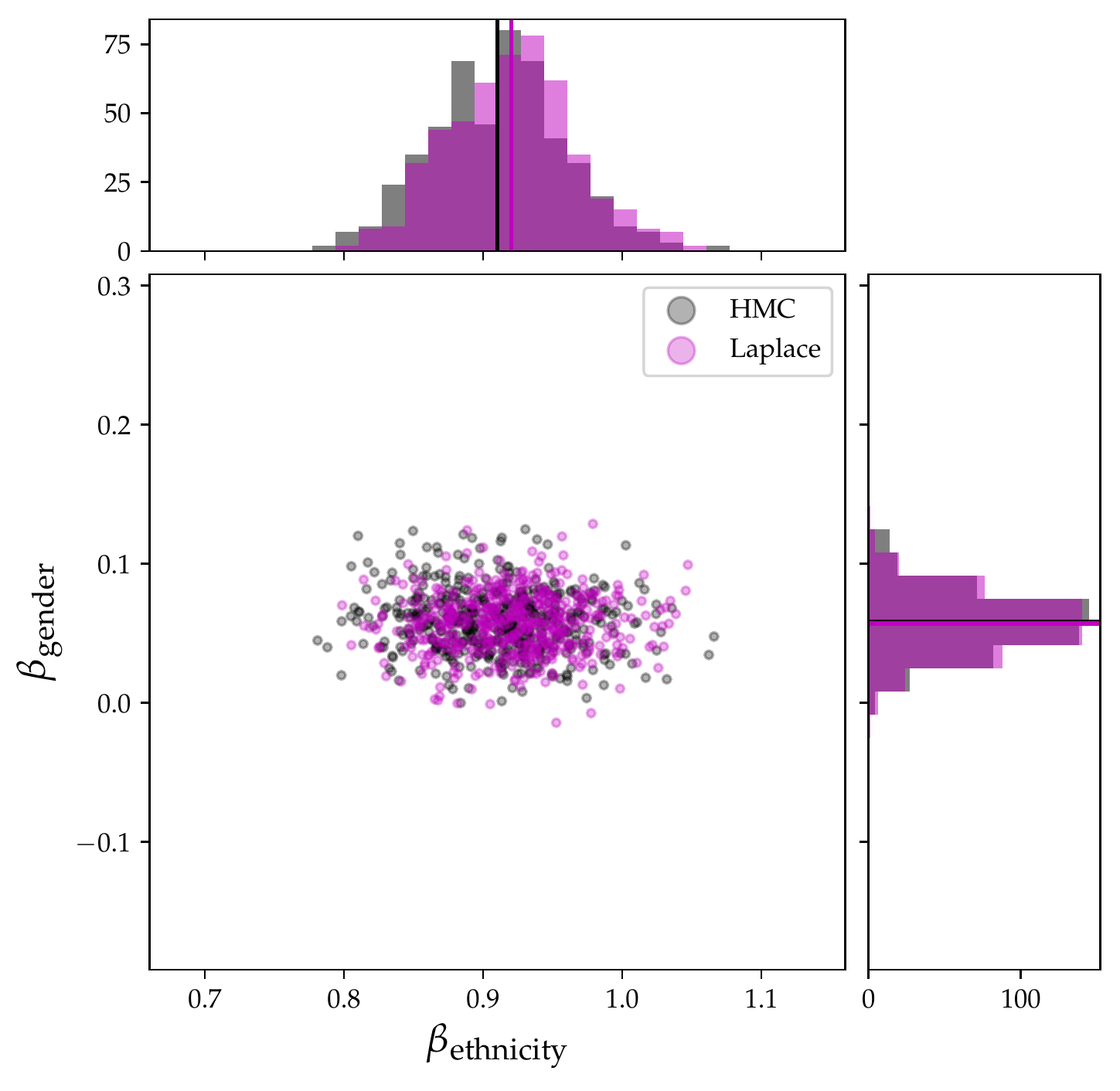}
\end{subfigure}
\begin{subfigure}[b]{0.32\textwidth}
	\includegraphics[width=\textwidth]{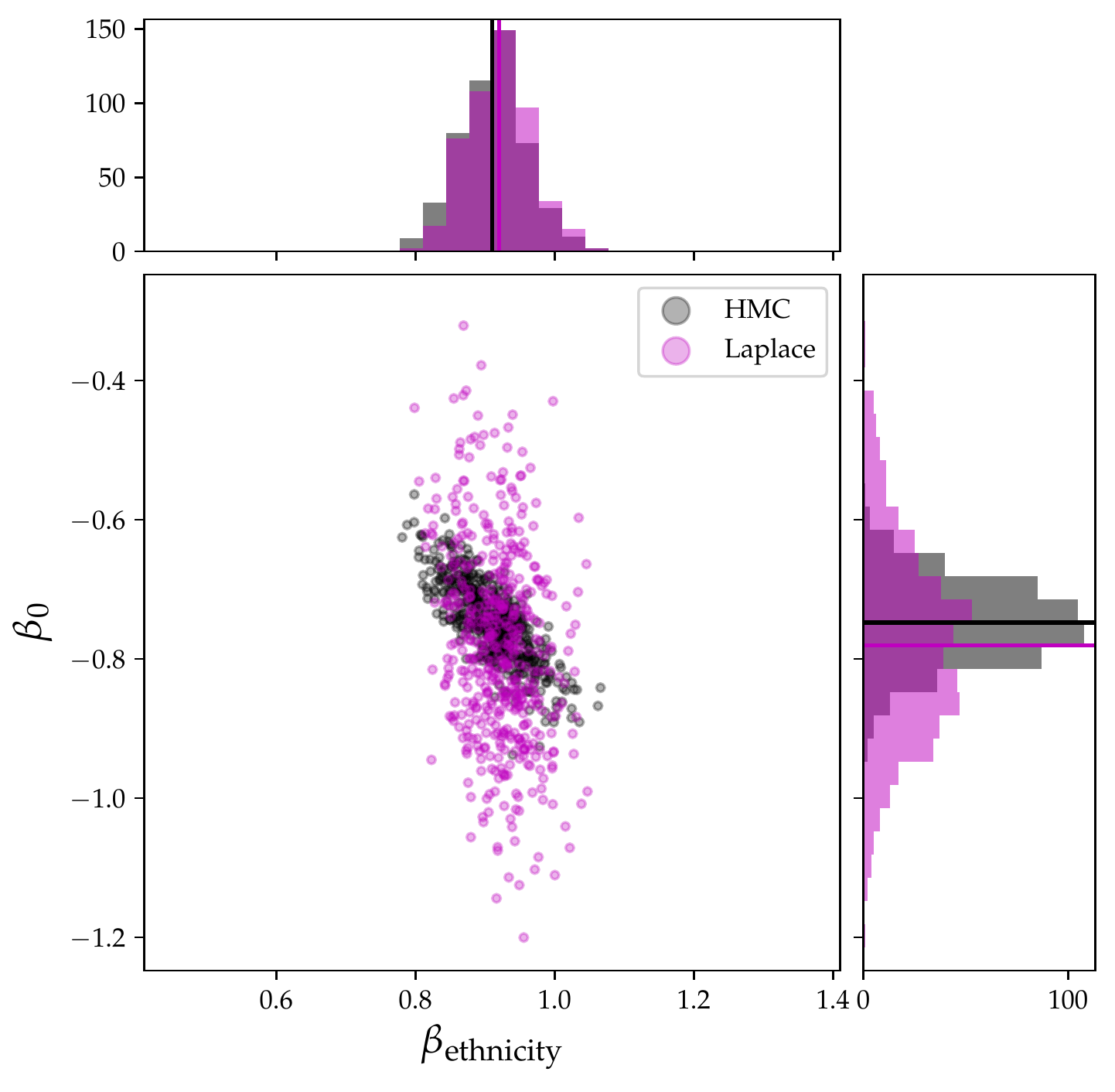}
\end{subfigure}
\begin{subfigure}[b]{0.32\textwidth}
	\includegraphics[width=\textwidth]{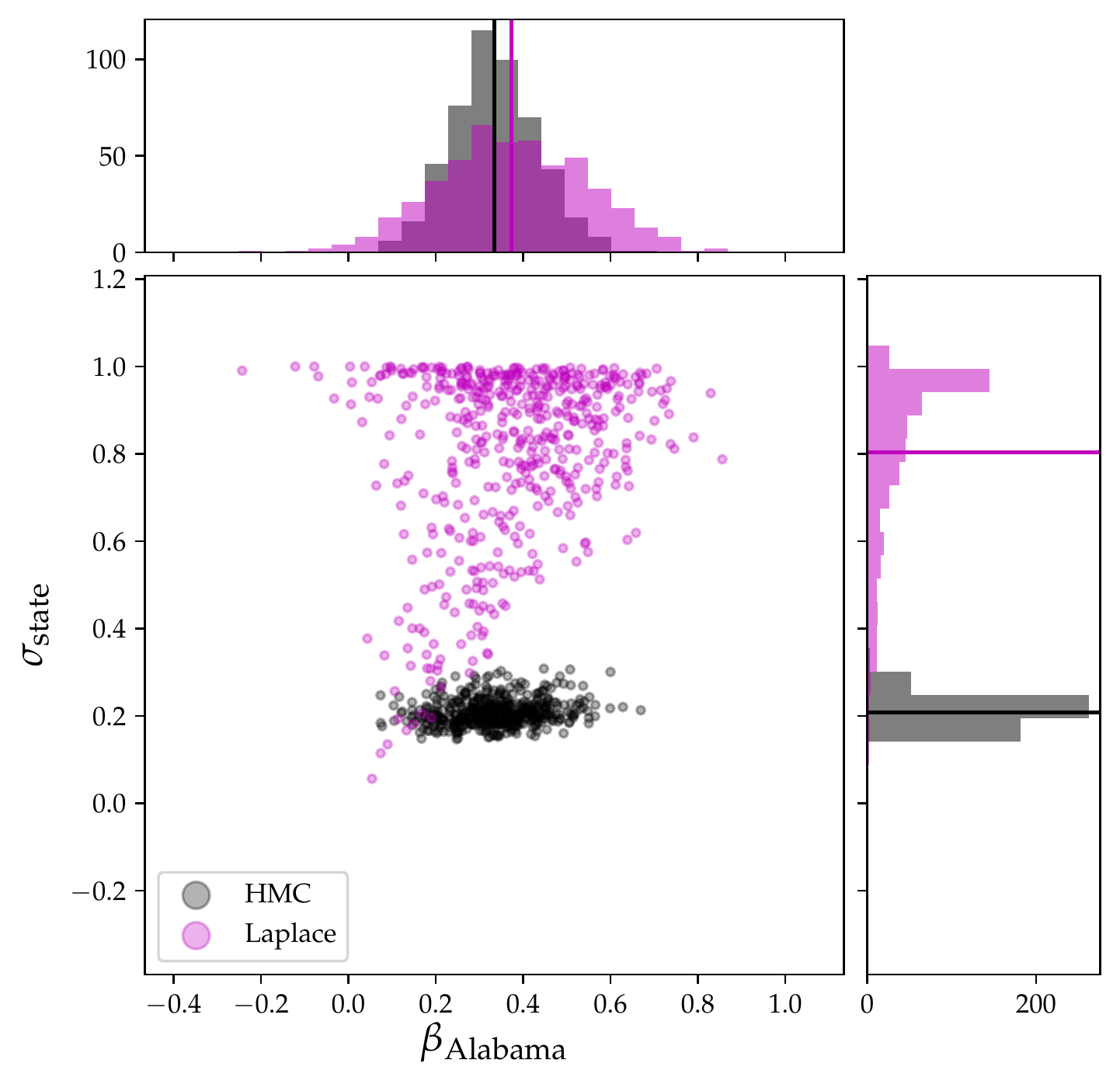}
\end{subfigure}
\caption{Scatter plots for certain parameter combinations for all the different methods in comparison to HMC in the logistic regression example. The parameter pairs vary from left to right, the methods from top to bottom.}
	\label{fig:small_EL}
\end{figure}

\begin{figure}
	\begin{subfigure}[b]{0.49\textwidth}
	\includegraphics[width=\textwidth]{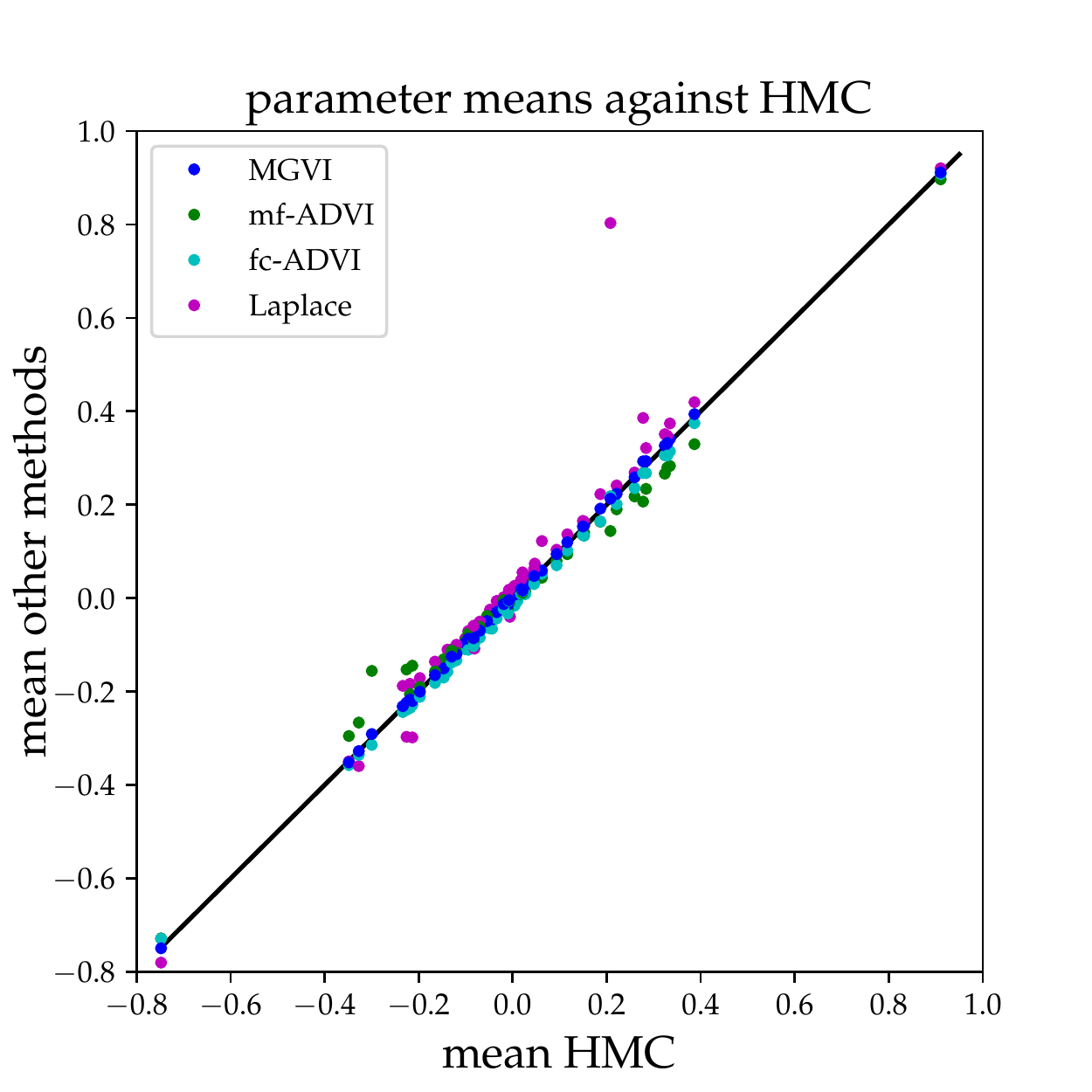}
\end{subfigure}
	\begin{subfigure}[b]{0.49\textwidth}
	\includegraphics[width=\textwidth]{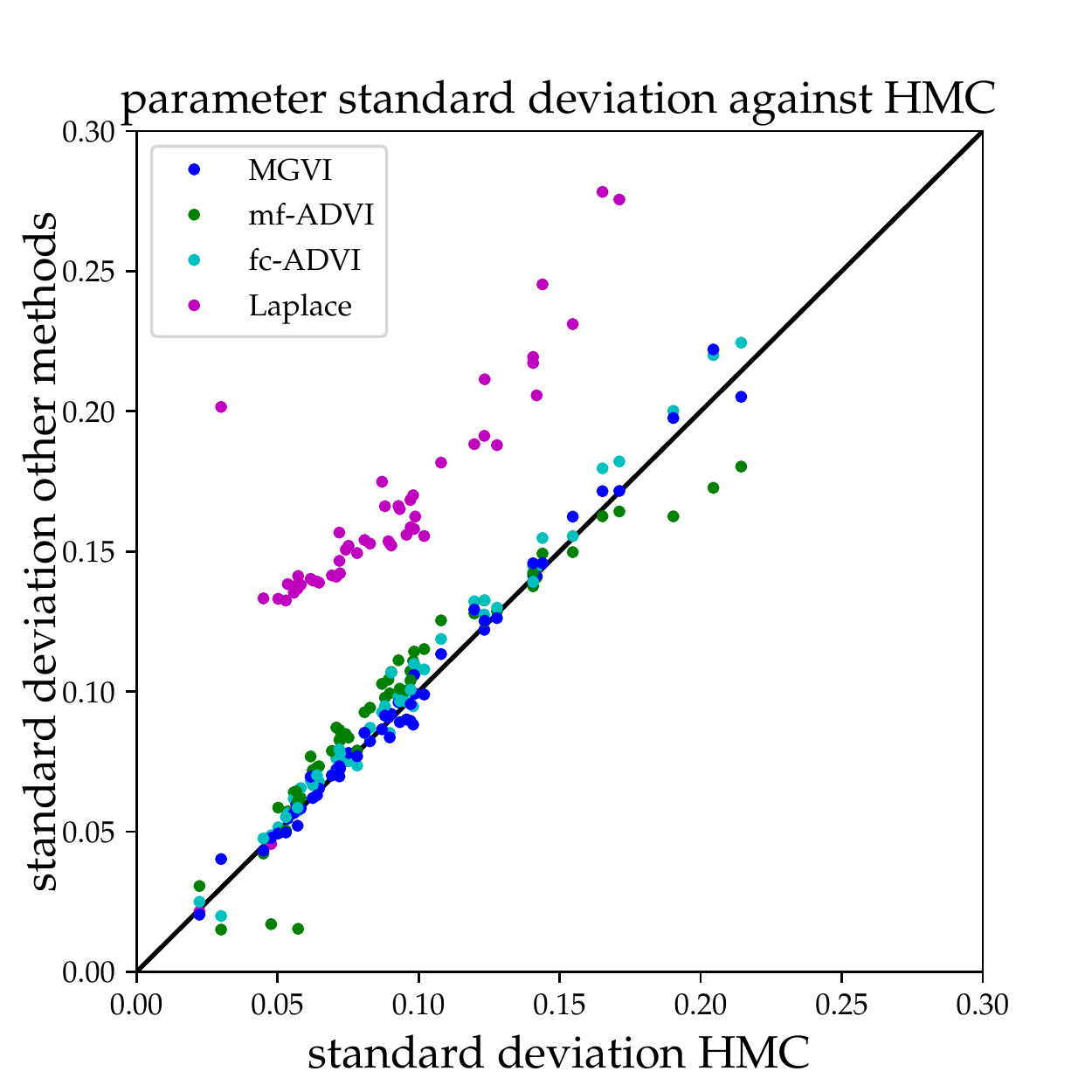}
\end{subfigure}
\caption{Mean (left) and standard deviation estimates (right) for all parameters and methods plotted against the HMC estimates.}
	\label{fig:against_hmc}
\end{figure}

For our analysis we compare MGVI, fc- and mf-ADVI, a Laplace approximation, as well as HMC. Our initial sampling accuracy are only $25$ conjugate gradient steps due to the relatively low number of problem parameters, and we increase it to $100$ towards the end and otherwise we use the setup from the first two examples. Regarding convergence, we ran MGVI  and the ADVI methods for a total of $1000$ seconds each, although all, except fc-ADVI, converged within seconds, as did the MAP estimate. After a burn-in and parameter tuning phase we sampled with five chains for several hours, ending with mean Gelman-Rubin test statistic $\widehat{R}_\mathrm{mean} = 1.002$ over all parameters and maximum $\widehat{R}_\mathrm{max}=1.009$. The smallest effective sample size was $\mathrm{ESS}_{\mathrm{min}}=500$, which is the number of samples we use for our analysis. Fig.~\ref{fig:small_EL} shows scatter plots of different model parameters against each other, comparing HMC to the other methods. MGVI (blue) performs remarkably well, as it is almost indistinguishable from HMC in all cases, matching in mean, variance, and correlation. As expected, fc-ADVI (cyan) also captures the true posterior distribution quite well, but only at extremely high computational cost. As in the previous examples, mf-ADVI (green) does not capture any correlations, but also tends to under-estimate the uncertainty. The  recovered mean clearly differs form the sampled posterior mean, and for the more nonlinear $\sigma_\mathrm{state}$ parameter, a systematic shift is observed. The Laplace approximation works decently for some parameters, for others only the variance is off, and for other directions straightforwardly fails, as it can be observed to happen most severely in the last panel.

Fig.~\ref{fig:against_hmc} shows the means and standard deviations of all model parameters and methods against the HMC results. Again, in both plots MGVI seems to be superior to mf-ADVI, as well as the Laplace approximation, which is also supported by the RMS errors, as shown in Tab.~\ref{tab:EL_RMS}. Here MGVI has significantly smaller mean errors compared to all other methods. Compared to fc-ADVI, the error is only a third, and to mf-ADVI one seventh. In the standard deviations the difference is not as severe, but also there MGVI is the closest to HMC. In the means, the Laplace approximation is only completely off once, namely for the hierarchical $\sigma_\mathrm{state}$ parameter. For the standard deviations, Laplace is rarely correct, most uncertainties appear far too large. Several points are outside the plot, with deviations up to $0.9$. Overall, MGVI seems to be the best among the tested methods for this problem in terms of accuracy.
 \begin{table}
 	\centering
 	\caption{The RMS error of parameter means and standard deviations relative to  HMC.}
 	\label{tab:EL_RMS}
 	\begin{tabular}{|c|c|c|c|c|} 
 		\hline
 		RMS HMC against&MGVI &fc-ADVI&mf-ADVI&Laplace\\
 		\hline
 		mean & $0.0047$ &$0.015$ &$0.035$ &$ 0.13$ \\
 		 standard deviation & $0.0051$&$0.0067$ & $0.014$ &$ 0.14$ \\
 		\hline
 	\end{tabular}
 \end{table}

\subsubsection{The full model}

\begin{figure}
	\centering
		\includegraphics[scale=0.5]{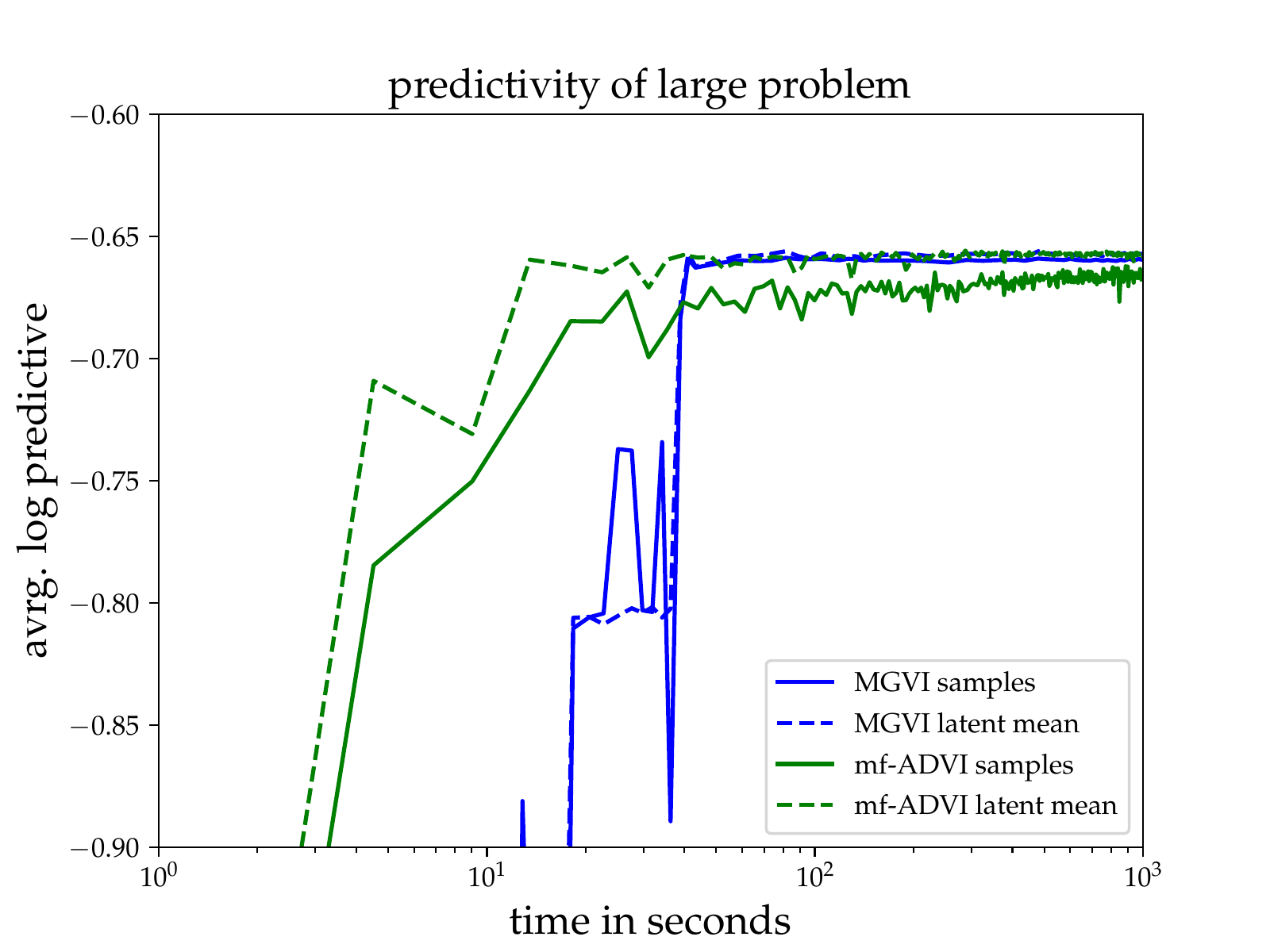}
	\caption{The predictive likelihood of MGVI and mf-ADVI in the large logistic regression example.}
		\label{fig:EL_large}
\end{figure}

The full model, as described in detail in \cite{polldata}, additionally takes further regressors into account, such as the multi-class variables of age, education and region, as well as combinations of categories, and previous election results. In addition, now the coefficients of all categories follow a Gaussian prior with a priori unknown standard deviation. As in the simple problem, a uniform, hierarchical prior with some upper limit is imposed on those. In the original model the interval $[0,100]$ is proposed, corresponding to largely uninformative prior distributions. From a Bayesian inference perspective, the posterior is extremely far away from the prior distribution, containing much more information. This is a problem for every method starting somewhat close to the prior distribution. For HMC it is hard to find the posterior mode to explore, making sampling in such scenarios inefficient and laborious. MGVI also experiences something similar, which can be seen in Fig.~\ref{fig:EL_large}. In this case it is significantly slower in the beginning, compared to mf-ADVI. Everywhere, except close to the posterior mode, the metric demands a large variance of the Gaussian and the stochastic nature of the optimization may result in a new location still far away from the posterior mode with practically unchanged metric. Therefore, only by chance the mode is found, and once it is, MGVI will quickly contract its variance and converges. For MGVI we found that in this case the stochasticity introduced by only a single pair of antithetic samples in combination with deep convergence for this given sample yields best results. This way we can escape local minima and flat energy landscapes.  The initial position of mf-ADVI is close to a delta distribution and will therefore initially mimic the behavior of a MAP estimate, which is much better suited for such a scenario with unconstraining priors and strong likelihood. The mean converges relatively fast, but the sample average of the predictive likelihood is slow, as seen in the other examples as well.

To overcome the limitations of MGVI in problems with weak priors and strong likelihoods, one could come up with heuristic schemes to artificially reduce the sample variance in the beginning, also imitating MAP, or possibly even starting with mf-ADVI and later on switching to MGVI, keeping the mean estimate.

\section{Conclusion}
We proposed Metric Gaussian Variational Inference (MGVI) as a general method to perform approximate Bayesian inference for high-dimensional and complex posterior distributions. MGVI scales linearly in terms of memory and computations with the problem size, making it applicable in scenarios with millions of parameters. 
MGVI iterates between approximating the covariance with the inverse Fisher information metric at the current mean estimate and optimizing the KL-divergence to the true posterior for the current covariance estimate to update the mean. Drawing samples from the approximate distribution via implicit sampling avoids storing the covariance explicitly at any point in time, leading to the linear scaling. The samples are used for an stochastic estimate of the KL-divergence and its gradient. The variance of these estimates can be reduced via antithetic sampling and the optimization is performed via natural gradient descent. The algorithm has converged once the mean estimate is self-consistent with the covariance. The result is a set of samples from the approximate posterior distribution that implicitly represent correlations between all parameters, going beyond a mean-field approximation while circumnavigating the quadratic scaling of an explicit covariance.

In our numerical experiments we demonstrate the accuracy of MGVI by comparing it to HMC samples, outperforming the Laplace approximation, mean-field, and even full-covariance ADVI. In addition to this, in most examples MGVI is significantly faster than the ADVI methods, as well as HMC. Applying MGVI in a diverse set of different contexts illustrates the versatility of the method. In the logistic regression example we have shown that MGVI is intrinsically different to an Laplace approximation, as it finds better solutions in complex models, mimicking the behavior of full-covariance ADVI. MGVI is also suited for large-scale image reconstruction problems with millions of parameters and complex models and it provides accurate uncertainty quantification, which can be used to propagate errors to any derived science result.

For the future it is left to explore the limits of MGVI, both, numerically and theoretically. For the problem dimensionality we do not see any conceptual limitation, except the linear scaling. More problematic is model complexity, and especially degenerate parameter directions. Those lead to numerical stiffness, and therefore slow convergence. Solving certain sub-problems individually, or using tempering methods might result in overall faster convergence. Better justified heuristics for the meta-parameter choices will have to be developed. It is also unclear how to deal with truly large data sets, i.e. too large to work with at once, as the Fisher information metric requires it. How sub-sampling the likelihood and mini-batching the data affects MGVI is to be explored. On the theoretical side, the properties of the proposed covariance approximation and the convergence behavior of the method have to be explored more rigorously. 

Finally, we hope that MGVI will open the door to even larger and more complex Bayesian inference problems in the future.
\section{Acknowledgments}
We acknowledge Philipp Arras, Philipp Frank, Maksim Greiner, Sebastian Hutschenreuter, Reimar Leike, Daniel Pumpe, Martin Reinecke and Theo Steininger for fruitful discussions.

\bibliography{citations}

\begin{thebibliography}{41}
\providecommand{\natexlab}[1]{#1}
\providecommand{\url}[1]{\texttt{#1}}
\expandafter\ifx\csname urlstyle\endcsname\relax
  \providecommand{\doi}[1]{doi: #1}\else
  \providecommand{\doi}{doi: \begingroup \urlstyle{rm}\Url}\fi

\bibitem[Amari(1997)]{amari1997neural}
Shun-ichi Amari.
\newblock Neural learning in structured parameter spaces-natural riemannian
  gradient.
\newblock In \emph{Advances in neural information processing systems}, pages
  127--133, 1997.

\bibitem[Amari(2016)]{InformationGeometry}
Shun-ichi Amari.
\newblock \emph{Information geometry and its applications}.
\newblock Springer, 2016.

\bibitem[Arras et~al.(2019{\natexlab{a}})Arras, Baltac, Ensslin, Frank,
  Hutschenreuter, Knollmueller, Leike, Newrzella, Platz, Reinecke,
  et~al.]{Nifty5}
Philipp Arras, Mihai Baltac, Torsten~A Ensslin, Philipp Frank, Sebastian
  Hutschenreuter, Jakob Knollmueller, Reimar Leike, Max-Niklas Newrzella, Lukas
  Platz, Martin Reinecke, et~al.
\newblock Nifty5: Numerical information field theory v5.
\newblock \emph{Astrophysics Source Code Library}, 2019{\natexlab{a}}.

\bibitem[Arras et~al.(2019{\natexlab{b}})Arras, Frank, Leike, Westermann, and
  En{\ss}lin]{Radiocalibration}
Philipp Arras, Philipp Frank, Reimar Leike, R{\"u}diger Westermann, and Torsten
  En{\ss}lin.
\newblock Unified radio interferometric calibration and imaging with joint
  uncertainty quantification.
\newblock \emph{arXiv preprint arXiv:1903.11169}, 2019{\natexlab{b}}.

\bibitem[Barrau and Bonnabel(2013)]{barrau2013note}
Axel Barrau and Silvere Bonnabel.
\newblock A note on the intrinsic cramer-rao bound.
\newblock In \emph{International Conference on Geometric Science of
  Information}, pages 377--386. Springer, 2013.

\bibitem[Betancourt and Girolami(2015)]{HMCHierarchy}
Michael Betancourt and Mark Girolami.
\newblock Hamiltonian monte carlo for hierarchical models.
\newblock \emph{Current trends in Bayesian methodology with applications},
  79:\penalty0 30, 2015.

\bibitem[Bishop(2006)]{Bishop}
Christopher~M. Bishop.
\newblock \emph{Pattern Recognition and Machine Learning (Information Science
  and Statistics)}.
\newblock Springer-Verlag, Berlin, Heidelberg, 2006.
\newblock ISBN 0387310738.

\bibitem[Blei et~al.(2017)Blei, Kucukelbir, and McAuliffe]{VariationalReview}
David~M Blei, Alp Kucukelbir, and Jon~D McAuliffe.
\newblock Variational inference: A review for statisticians.
\newblock \emph{Journal of the American Statistical Association}, 112\penalty0
  (518):\penalty0 859--877, 2017.

\bibitem[Braun and McAuliffe(2010)]{braun2010variational}
Michael Braun and Jon McAuliffe.
\newblock Variational inference for large-scale models of discrete choice.
\newblock \emph{Journal of the American Statistical Association}, 105\penalty0
  (489):\penalty0 324--335, 2010.

\bibitem[Cram{\'e}r(1946)]{cramer}
Harald Cram{\'e}r.
\newblock \emph{Mathematical methods of statistics}, volume~9.
\newblock Princeton university press, 1946.

\bibitem[Devroye(1986)]{InverseTransform}
Luc Devroye.
\newblock Sample-based non-uniform random variate generation.
\newblock In \emph{Proceedings of the 18th conference on Winter simulation},
  pages 260--265. ACM, 1986.

\bibitem[Duane et~al.(1987)Duane, Kennedy, Pendleton, and Roweth]{HMC}
Simon Duane, Anthony~D Kennedy, Brian~J Pendleton, and Duncan Roweth.
\newblock Hybrid monte carlo.
\newblock \emph{Physics letters B}, 195\penalty0 (2):\penalty0 216--222, 1987.

\bibitem[Frank et~al.(2019)Frank, Leike, and En{\ss}lin]{CausalFields}
Philipp Frank, Reimar Leike, and Torsten~A En{\ss}lin.
\newblock Field dynamics inference for local and causal interactions.
\newblock \emph{arXiv preprint arXiv:1902.02624}, 2019.

\bibitem[Gelman and Hill(2006)]{polldata}
Andrew Gelman and Jennifer Hill.
\newblock \emph{Data analysis using regression and multilevel/hierarchical
  models}.
\newblock Cambridge university press, 2006.

\bibitem[Ghosh and Ramamoorthi(2011)]{BCLT}
J.~Ghosh and R.~Ramamoorthi.
\newblock Bayesian nonparametrics.
\newblock \emph{Springer Series in Statistics}, 16, 01 2011.

\bibitem[Giordano et~al.(2018)Giordano, Broderick, and Jordan]{LR-ADVI}
Ryan Giordano, Tamara Broderick, and Michael~I Jordan.
\newblock Covariances, robustness and variational bayes.
\newblock \emph{The Journal of Machine Learning Research}, 19\penalty0
  (1):\penalty0 1981--2029, 2018.

\bibitem[Hartmann and Vanhatalo(2018)]{hartmann2017laplace}
Marcelo Hartmann and Jarno Vanhatalo.
\newblock Laplace approximation and natural gradient for gaussian process
  regression with heteroscedastic student-t model.
\newblock \emph{Statistics and Computing}, pages 1--21, 2018.

\bibitem[Hutschenreuter and En{\ss}lin(2019)]{NewFaraday}
Sebastian Hutschenreuter and Torsten~A En{\ss}lin.
\newblock The galactic faraday depth sky revisited.
\newblock \emph{arXiv preprint arXiv:1903.06735}, 2019.

\bibitem[Kass and Raftery(1995)]{kass1995bayes}
Robert~E Kass and Adrian~E Raftery.
\newblock Bayes factors.
\newblock \emph{Journal of the american statistical association}, 90\penalty0
  (430):\penalty0 773--795, 1995.

\bibitem[Khan et~al.(2013)Khan, Aravkin, Friedlander, and Seeger]{khan2013fast}
Mohammad~Emtiyaz Khan, Aleksandr Aravkin, Michael Friedlander, and Matthias
  Seeger.
\newblock Fast dual variational inference for non-conjugate latent gaussian
  models.
\newblock In \emph{International Conference on Machine Learning}, pages
  951--959, 2013.

\bibitem[Kingma and Welling(2013)]{AEVB}
Diederik~P Kingma and Max Welling.
\newblock Auto-encoding variational bayes.
\newblock \emph{arXiv preprint arXiv:1312.6114}, 2013.

\bibitem[Knollm{\"u}ller and En{\ss}lin(2018)]{WhitePriors}
Jakob Knollm{\"u}ller and Torsten~A En{\ss}lin.
\newblock Encoding prior knowledge in the structure of the likelihood.
\newblock \emph{arXiv preprint arXiv:1812.04403}, 2018.

\bibitem[Knowles and Minka(2011)]{knowles2011non}
David~A Knowles and Tom Minka.
\newblock Non-conjugate variational message passing for multinomial and binary
  regression.
\newblock In \emph{Advances in Neural Information Processing Systems}, pages
  1701--1709, 2011.

\bibitem[Kroese et~al.(2013)Kroese, Taimre, and Botev]{kroese2013handbook}
Dirk~P Kroese, Thomas Taimre, and Zdravko~I Botev.
\newblock \emph{Handbook of monte carlo methods}, volume 706.
\newblock John Wiley \& Sons, 2013.

\bibitem[Kucukelbir et~al.(2017)Kucukelbir, Tran, Ranganath, Gelman, and
  Blei]{ADVI}
Alp Kucukelbir, Dustin Tran, Rajesh Ranganath, Andrew Gelman, and David~M Blei.
\newblock Automatic differentiation variational inference.
\newblock \emph{The Journal of Machine Learning Research}, 18\penalty0
  (1):\penalty0 430--474, 2017.

\bibitem[Kullback and Leibler(1951)]{KLdivergence}
Solomon Kullback and Richard~A Leibler.
\newblock On information and sufficiency.
\newblock \emph{The annals of mathematical statistics}, 22\penalty0
  (1):\penalty0 79--86, 1951.

\bibitem[Kuss and Rasmussen(2005)]{kuss2005assessing}
Malte Kuss and Carl~Edward Rasmussen.
\newblock Assessing approximate inference for binary gaussian process
  classification.
\newblock \emph{Journal of machine learning research}, 6\penalty0
  (Oct):\penalty0 1679--1704, 2005.

\bibitem[L{\'a}zaro-Gredilla and Titsias(2011)]{lazaro2011variational}
Miguel L{\'a}zaro-Gredilla and Michalis~K Titsias.
\newblock Variational heteroscedastic gaussian process regression.
\newblock In \emph{ICML}, pages 841--848, 2011.

\bibitem[Leike and En{\ss}lin(2019)]{leike2019charting}
RH~Leike and TA~En{\ss}lin.
\newblock Charting nearby dust clouds using gaia data only.
\newblock \emph{arXiv preprint arXiv:1901.05971}, 2019.

\bibitem[Martens(2014)]{NaturalGradientReview}
James Martens.
\newblock New insights and perspectives on the natural gradient method.
\newblock \emph{arXiv preprint arXiv:1412.1193}, 2014.

\bibitem[Nickisch and Rasmussen(2008)]{nickisch2008approximations}
Hannes Nickisch and Carl~Edward Rasmussen.
\newblock Approximations for binary gaussian process classification.
\newblock \emph{Journal of Machine Learning Research}, 9\penalty0
  (Oct):\penalty0 2035--2078, 2008.

\bibitem[Opper and Archambeau(2009)]{GaussianRevisited}
Manfred Opper and C{\'e}dric Archambeau.
\newblock The variational gaussian approximation revisited.
\newblock \emph{Neural computation}, 21\penalty0 (3):\penalty0 786--792, 2009.

\bibitem[Papandreou and Yuille(2010)]{papandreou2010gaussian}
George Papandreou and Alan~L Yuille.
\newblock Gaussian sampling by local perturbations.
\newblock In \emph{Advances in Neural Information Processing Systems}, pages
  1858--1866, 2010.

\bibitem[Rao(1992)]{rao}
C~Radhakrishna Rao.
\newblock Information and the accuracy attainable in the estimation of
  statistical parameters.
\newblock In \emph{Breakthroughs in statistics}, pages 235--247. Springer,
  1992.

\bibitem[R{\"u}schendorf(2009)]{DistributionalTransform}
Ludger R{\"u}schendorf.
\newblock On the distributional transform, sklar's theorem, and the empirical
  copula process.
\newblock \emph{Journal of Statistical Planning and Inference}, 139\penalty0
  (11):\penalty0 3921--3927, 2009.

\bibitem[Sch{\"u}tzenberger(1957)]{schutzenberger1957generalization}
MP~Sch{\"u}tzenberger.
\newblock A generalization of the fr{\'e}chet-cram{\'e}r inequality to the case
  of bayes estimation.
\newblock \emph{Bull. Amer. Math. Soc}, 63\penalty0 (142), 1957.

\bibitem[Selig et~al.(2013)Selig, Bell, Junklewitz, Oppermann, Reinecke,
  Greiner, Pachajoa, and En{\ss}lin]{Nifty}
Marco Selig, Michael~R Bell, Henrik Junklewitz, Niels Oppermann, Martin
  Reinecke, Maksim Greiner, Carlos Pachajoa, and Torsten~A En{\ss}lin.
\newblock {NIFTY--Numerical Information Field Theory-A versatile PYTHON library
  for signal inference}.
\newblock \emph{Astronomy \& Astrophysics}, 554:\penalty0 A26, 2013.

\bibitem[Shewchuk et~al.(1994)]{AgonizingPain}
Jonathan~Richard Shewchuk et~al.
\newblock An introduction to the conjugate gradient method without the
  agonizing pain, 1994.

\bibitem[Steininger et~al.(2017)Steininger, Dixit, Frank, Greiner,
  Hutschenreuter, Knollm{\"u}ller, Leike, Porqueres, Pumpe, Reinecke,
  et~al.]{Nifty3}
Theo Steininger, Jait Dixit, Philipp Frank, Maksim Greiner, Sebastian
  Hutschenreuter, Jakob Knollm{\"u}ller, Reimar Leike, Natalia Porqueres,
  Daniel Pumpe, Martin Reinecke, et~al.
\newblock Nifty 3-numerical information field theory-a python framework for
  multicomponent signal inference on hpc clusters.
\newblock \emph{arXiv preprint arXiv:1708.01073}, 2017.

\bibitem[Van~der Vaart(2000)]{van2000asymptotic}
A.~W. Van~der Vaart.
\newblock \emph{Asymptotic statistics}, volume~3.
\newblock Cambridge university press, 2000.

\bibitem[Wijsman et~al.(1973)]{wijsman1973attainment}
RA~Wijsman et~al.
\newblock On the attainment of the cram{\'e}r-rao lower bound.
\newblock \emph{The Annals of Statistics}, 1\penalty0 (3):\penalty0 538--542,
  1973.

\end{thebibliography}
\end{document}